\documentclass[lettersize,journal]{IEEEtran}
\usepackage{amsmath,amsfonts}
\usepackage{algorithmic}
\usepackage{array}
\usepackage{textcomp}
\usepackage{stfloats}
\usepackage{url}
\usepackage{verbatim}
\usepackage{graphicx}
\def\BibTeX{{\rm B\kern-.05em{\sc i\kern-.025em b}\kern-.08em
    T\kern-.1667em\lower.7ex\hbox{E}\kern-.125emX}}
\usepackage{balance}
\usepackage{multirow}
\usepackage{rotating}
\usepackage{booktabs}
\usepackage{cleveref}
\usepackage{enumitem}
\usepackage{amssymb}
\newcommand{\circnum}[1]{\textcircled{\raisebox{0.15pt}{\scriptsize\strut #1}}}
\def\myrotcell#1{\rotatebox{90}{#1}}
\usepackage[ruled,vlined]{algorithm2e}
\usepackage{caption}
\usepackage{subcaption}
\usepackage{tabularx}
\usepackage{makecell}
\usepackage{bbm}
\usepackage{color}
\usepackage{mdframed}

\newcommand{\xq}[1]{{\color{black}#1}}

\begin{document}
\title{Modeling Temporal Dependencies within the Target for Long-Term Time Series Forecasting}
\author{
Qi Xiong, 
Kai Tang, 
Minbo Ma, 
Ji Zhang, 
Jie Xu,
and Tianrui Li$^*$~\IEEEmembership{Senior Member,~IEEE}
\thanks{
Qi Xiong, Kai Tang, Minbo Ma, Ji Zhang, and Tianrui Li (the corresponding author) are with the School of Computing and Artificial Intelligence,
Engineering Research Center of Sustainable Urban Intelligent Transportation, Ministry of Education, National Engineering Laboratory of Integrated
Transportation Big Data Application Technology, Manufacturing Industry
Chains Collaboration and Information Support Technology Key Laboratory
of Sichuan Province, Southwest Jiaotong University, Chengdu 611756, China
(e-mails: xiongqi@my.swjtu.edu.cn; tangkailh@outlook.com; minboma@my.swjtu.edu.cn; jizhang.jim@gmail.com; trli@swjtu.edu.cn).}
\thanks{Jie Xu is with the School of Computing, The University of Leeds, LS2 9JT
Leeds, U.K. (e-mail: j.xu@leeds.ac.uk).}
}

\markboth{Journal of \LaTeX\ Class Files,~Vol.~18, No.~9, September~2020}%
{How to Use the IEEEtran \LaTeX \ Templates}

\maketitle

\begin{abstract}
Long-term time series forecasting (LTSF) is a critical task across diverse domains. Despite significant advancements in LTSF research, we identify a performance bottleneck in existing LTSF methods caused by the inadequate modeling of Temporal Dependencies within the Target (TDT). To address this issue, we propose a novel and generic temporal modeling framework, Temporal Dependency Alignment (TDAlign), that equips existing LTSF methods with TDT learning capabilities. TDAlign introduces two key innovations: 1) a loss function that aligns the change values between adjacent time steps in the predictions with those in the target, ensuring consistency with variation patterns, and 2) an adaptive loss balancing strategy that seamlessly integrates the new loss function with existing LTSF methods without introducing additional learnable parameters. As a plug-and-play framework, TDAlign enhances existing methods with minimal computational overhead, featuring only linear time complexity and constant space complexity relative to the prediction length. Extensive experiments on six strong LTSF baselines across seven real-world datasets demonstrate the effectiveness and flexibility of TDAlign. On average, TDAlign reduces baseline prediction errors by \textbf{1.47\%} to \textbf{9.19\%} and change value errors by \textbf{4.57\%} to \textbf{15.78\%}, highlighting its substantial performance improvements.
We make our codes publicly available at \url{https://github.com/XQ-edu/TDAlign}.

\end{abstract}


\section{Introduction}
Time series forecasting refers to predicting the target future time series based on the input historical time series. It has garnered significant attention due to its crucial applications in various real-world domains, such as stock price prediction~\cite{feng2024multi,sezer2020financial}, traffic planning~\cite{10643332,ZHANG2024102413}, energy demand analysis~\cite{9935292,deb2017review}, and extreme weather warning~\cite{salman2015weather,gong2024spatio}. 
In the early stage, numerous studies \cite{al2018short} focused on short-term time series forecasting, typically predicting 48 time steps or fewer (e.g., hours or days)~\cite{lippi2013short, li2017diffusion, qin2017dual}. 
However, these methods often fail to meet the predictive capacity required for long-term time series forecasting (LTSF) in many practical scenarios, where anticipating future trends over extended periods, such as 720 time steps, is essential for significantly ahead in resource allocation, risk management, and decision-making. 
In response, recent studies present various neural networks to capture long-range temporal dependencies from input historical time series, including Multi-Layer Perceptrons (MLPs) ~\cite{li2023revisiting,zeng2023transformers,das2023long}, Convolutional Neural Networks (CNNs) ~\cite{liu2022scinet,wang2022micn,wu2022timesnet}, Transformer ~\cite{li2019enhancing,nie2022time,zhang2023crossformer}, and Recurrent Neural Networks (RNNs) \cite{lin2023segrnn}.

\begin{figure}[t]
    \centering
    \includegraphics[width=\columnwidth]{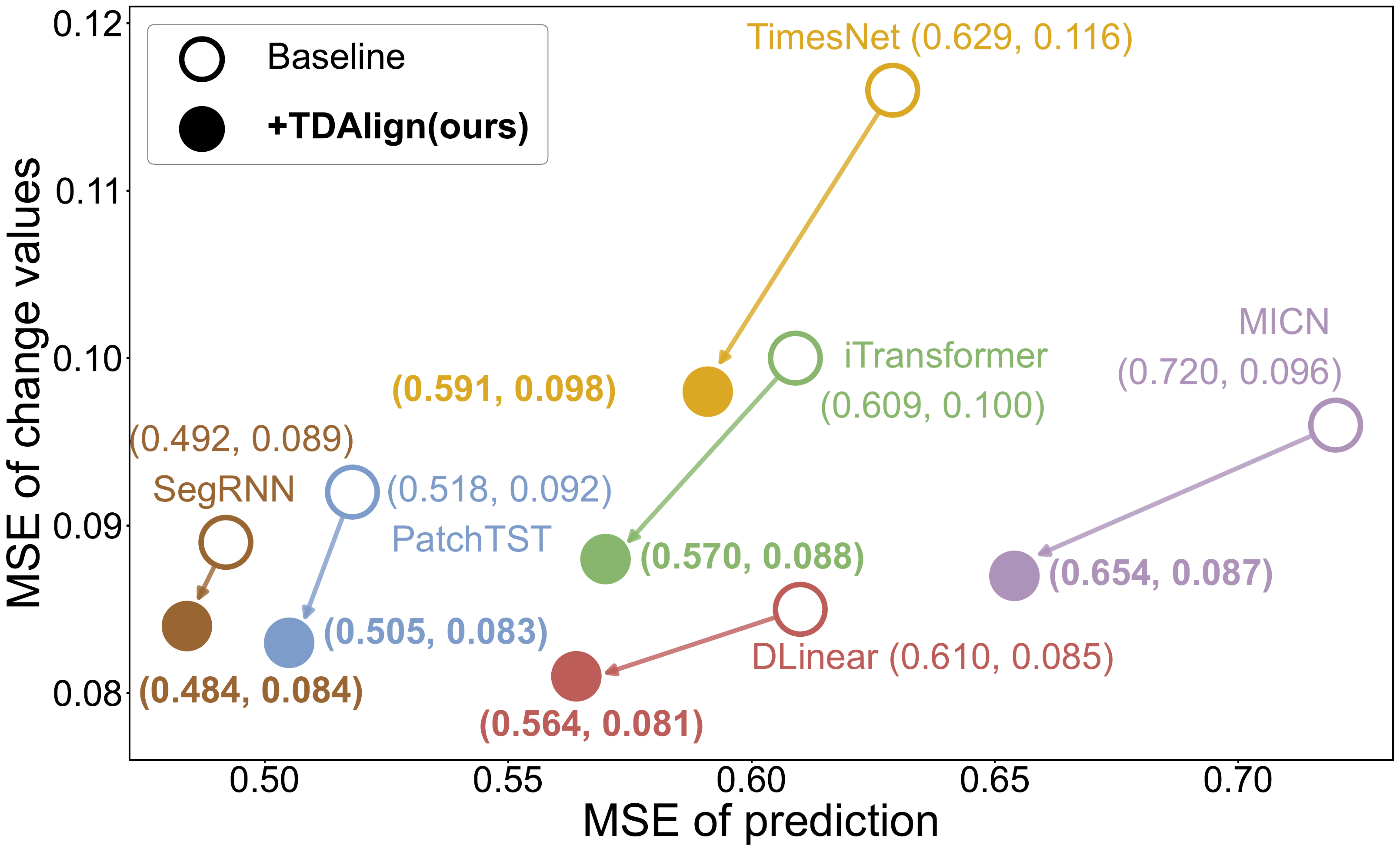}
    \caption{Performance comparisons between methods with and without our proposed TDAlign across six baseline methods. The x-axis presents the Mean Squared Error (MSE) scores, while the y-axis displays the MSE of change values between adjacent time steps within the prediction. The results are averaged over seven datasets. For MSE, a smaller value indicates better performance.}
    \label{fig:statistic_imp}
\end{figure}
Despite substantial progress, our in-depth analysis of temporal relations between adjacent time steps reveals that existing LTSF methods lack adequate modeling of \emph{temporal dependencies within the target} (TDT), which significantly limits the method's capacity to generate accurate and realistic predictions (see the analysis in Section~\ref{TDOS_effect} for details). 
Prior efforts such as \cite{das2023long, liu2021pyraformer, lin2023segrnn} introduce position/timestamp embeddings (e.g., minute, hour, week, month, and year) to preserve the temporal ordering information within the target. While modeling temporal feature discrepancy, such implicit learning cannot capture dynamic temporal patterns that evolve over time~\cite{zeng2023transformers,zhang2024intriguing}.
The autoregressive decoding strategy is another relevant solution that feeds the former prediction into the next step operation, naturally learning sequential relationships during the step-by-step inference process~\cite{hochreiter1997long, cho2014learning}. However, it suffers two major drawbacks: error accumulation and computational overhead, where both forecast bias and inference time rapidly increase as the forecast horizon expands~\cite{zhou2021informer, shen2023non}.

In this work, we propose Temporal Dependency Alignment (TDAlign), a generic framework that enables existing methods to capture temporal dependencies within the target time series via an additional learning objective. In Section~\ref{OTDM}, we first formulate the TDT with the change values between adjacent time steps to facilitate capturing intrinsic temporal patterns of the target. Building upon this, we introduce a novel loss function to align the change values within the prediction with those within the target time series, thereby guiding the model to effectively learn intricate temporal relations. 
Finally, we design an adaptive loss balancing strategy that dynamically adjusts the relative importance between the conventional forecasting objective and the TDT learning objective, enabling TDAlign to flexibly integrate with existing LTSF methods without introducing additional learnable parameters (Section \ref{DDR}).
\xq{We provide further theoretical analysis of TDAlign in Section \ref{sec:theory}.}
Computational complexity analysis in Section \ref{complexity} demonstrates that TDAlign incurs negligible computational overhead, with time complexity of $\mathcal{O}(H)$ and space complexity of $\mathcal{O}(1)$, where $H$ represents the length of the prediction.

\textbf{Flexibility and Effectiveness.}
Remarkably, our proposed TDAlign is orthogonal to existing LTSF methods without introducing additional learnable parameters. 
We evaluate TDAlign using a broad spectrum of strong baseline methods, including the MLP-based method DLinear~\cite{zeng2023transformers}, CNN-based methods MICN~\cite{wang2022micn} and TimesNet~\cite{wu2022timesnet}, Transformer-based methods PatchTST~\cite{nie2022time} and iTransformer~\cite{liu2023itransformer}, and the RNN-based method SegRNN~\cite{lin2023segrnn}. 
Experimental results on seven widely used LTSF benchmark datasets demonstrate that TDAlign consistently improves the forecasting performance of these six baseline methods, showcasing its strong flexibility and effectiveness (Section \ref{results}). 
As shown in Figure~\ref{fig:statistic_imp}, TDAlign achieves significant error reductions across seven datasets: 
(1) For change values between adjacent time steps within the prediction, absolute error reductions range from \textbf{4.57\%} to \textbf{15.78\%}.
(2) For the prediction, absolute error reductions range from \textbf{1.47\%} to \textbf{9.19\%}.

In summary, our contributions are as follows:
\begin{itemize}
    \item We provide a new perspective from the temporal relations between adjacent time steps within the target to analyze the performance bottleneck of existing LTSF methods, revealing their inadequate temporal dependency modeling.
    \item We propose TDAlign, a novel and generic framework for learning TDT. TDAlign is orthogonal to existing methods and introduces no additional learnable parameters, incurring minimal computational overhead.
    \item We conduct extensive experiments on six state-of-the-art methods across seven real-world datasets. The results demonstrate that TDAlign consistently improves the forecasting performance of these methods by a significant margin, showcasing its flexibility and effectiveness.
\end{itemize}

\section{Related Work}\label{related-work}
\subsection{Long-Term Time Series Forecasting}

Long-term time series forecasting (LTSF) has been extensively studied in recent years. In contrast to traditional short-term forecasting, which typically predicts 48 steps or fewer~\cite{ lippi2013short, li2017diffusion, qin2017dual}, LTSF extends forecast horizons, such as 720 steps~\cite{zhou2021informer}. The task poses substantial challenges for existing neural time series architectures, particularly in capturing long-term dependencies and managing model complexity.
Consequently, despite their proficiency in handling sequential data \cite{hochreiter1997long,salinas2020deepar}, the Recurrent Neural Network (RNN) has gradually lost prominence in the LTSF domain due to the rapid increase in error and inference time as the forecast horizon expands \cite{lai2018modeling,tan2023neural}. Recently, SegRNN \cite{lin2023segrnn} has revitalized the RNN architecture in the LTSF field by replacing segment-wise iterations instead of traditional point-wise processing, while leveraging non-autoregressive strategies to enable efficient parallel multi-step forecasting.

The Transformer architecture \cite{vaswani2017attention}, benefiting from the attention mechanism's capacity to capture long-term dependencies in sequential data, has garnered significant attention in the LTSF domain \cite{wu2021autoformer, zhou2022fedformer, liu2022non}.
Early efforts to adapt the vanilla Transformer for LTSF tasks focused on designing novel mechanisms to reduce the computational complexity of the original attention mechanism. 
For example, LogSparse \cite{li2019enhancing} proposes convolutional self-attention by employing causal convolutions to produce queries and keys in the self-attention layer. 
Informer \cite{zhou2021informer} introduces a ProbSparse self-attention with distilling techniques to extract dominant queries efficiently and a generative-style decoder for parallel multi-step forecasting. 
Pyraformer \cite{liu2021pyraformer} designs a hierarchical pyramidal attention module to capture temporal dependencies of different ranges with linear time and memory complexity.
These works use point-wise representations, which are proven insufficient to capture local semantic information in time series. To further enhance efficiency and forecasting performance, recent advancements have been devoted to exploring patch-wise and series-wise dependencies.
For instance, Crossformer \cite{zhang2023crossformer} and PatchTST \cite{nie2022time} leverage patch-based techniques, commonly employed in Natural Language Processing \cite{devlin2018bert} and Computer Vision \cite{he2022masked}, to divide time series data into subseries-level patches, capturing patch-wise dependency.
iTransformer \cite{liu2023itransformer} embeds independent time series as tokens and repurposes the Transformer architecture by applying attention mechanisms to these variate tokens, capturing series-wise dependency.

Besides the Transformer architecture, the Convolutional Neural Network (CNN) and the Multi-layer Perceptron (MLP) have also emerged as prominent architectures, demonstrating impressive results in the LTSF field.
CNN-based methods, such as SCINet \cite{liu2022scinet}, MICN \cite{wang2022micn}, and TimesNet \cite{wu2022timesnet}, excel at capturing local and multi-scale temporal features.
Following the success of DLinear \cite{zeng2023transformers}, which highlighted the efficacy of parsimonious models in addressing complex time series data, MLP-based methods have gained increasing attention, as exemplified by RLinear \cite{li2023revisiting}, TiDE \cite{das2023long} and TSMixer \cite{ekambaram2023tsmixer}.

\subsection{Modeling TDT in Time Series Forecasting}
Modeling the temporal dependencies within the target refers to the method's ability to effectively capture the internal correlations across multiple steps in the target time series, which fundamentally influence and reflect the overall forecasting performance.
Existing research predominantly explores these complex temporal relations through two primary approaches.

Early studies adopted the autoregressive decoding strategy to forecast multiple steps, such as LSTM \cite{hochreiter1997long}, GRU \cite{cho2014learning}, and DeepAR \cite{salinas2020deepar}. These methods utilize the most recently forecasted value as part of the input for predicting the next step. During the recursive process, models naturally learn the dependencies between consecutive steps.
However, despite their impressive success in short-term time series forecasting \cite{wu2020adversarial}, autoregressive methods are unsuitable for long-term time series forecasting tasks due to error accumulation and slow inference speeds inherent in step-by-step prediction \cite{zhou2021informer, shen2023non}.

On the other hand, numerous methods attempt to preserve the temporal order information within the target time series by incorporating auxiliary features such as position or timestamp embeddings. For instance, Pyraformer \cite{liu2021pyraformer} combines zero placeholders, positional embeddings, and timestamp embeddings to represent prediction tokens. SegRNN \cite{lin2023segrnn} leverages relative positional embeddings to indicate the sequential order of multiple segments. Methods can implicitly learn some internal correlations with temporal order information, where the model assigns distinct weights to different positions or timestamps \cite{shao2022spatial, wang2024rethinking}. However, such implicit learning is insufficient for effectively capturing intricate and dynamic temporal patterns \cite{zeng2023transformers, zhang2024intriguing}. 

The modeling of TDT remains inadequate in existing long-term time series forecasting methods. In this work, we aim to address this limitation and substantially enhance the forecasting performance. We highlight that our proposed framework is orthogonal to existing methods, enabling end-to-end training for non-autoregressive forecasting.

\section{Method}
In this section, we start with the definition of the LTSF tasks and the introduction of frequently used notions. Subsequently, we provide an in-depth analysis of existing LTSF methods and reveal their inadequate modeling of TDT. Following this, we present a detailed exposition of our proposed Temporal Dependency Alignment (TDAlign) framework, designed to address this limitation. Finally, we conduct a complexity analysis of TDAlign to demonstrate its efficiency.

\subsection{Preliminaries}\label{LTSF}
\begin{table}[ht]
\centering
\caption{Mathematical Notation.}
\label{tab:notions}
\resizebox{\columnwidth}{!}{%
\begin{tabularx}{\columnwidth}{>{\arraybackslash}m{2cm} X}
\toprule
Notions & \makecell{Descriptions} \\
\midrule
$N$ & The number of variables. \\
$x_t \in \mathbb{R}^N$ & The observation at the $t$-th time step of $N$ variables. \\
$L$ & The length of the historical input. \\
$H$ & The length of the future target and prediction. \\
$X \in \mathbb{R}^{L\times N}$ & The input of \( N \) variables over \( L \) time steps. \\
$Y, \widehat{Y} \in \mathbb{R}^{H\times N}$ & The target and prediction of \( N \) variables over \( H \) time steps. \\
$d_t, \widehat{d_t} \in \mathbb{R}^N$ & The temporal dependency at the \( t \)-th time step within the target and prediction of \( N \) variables. \\
$D, \widehat{D} \in \mathbb{R}^{H \times N}$ & The temporal dependencies within the target and prediction of \( N \) variables over \( H \) time steps. \\
$\operatorname{sgn}(\cdot)$ & The sign function. \\
$\mathbbm{1}(\cdot)$ & The indicator function. \\
$\mathcal{L}_Y$ & The loss between \( Y \) and \( \widehat{Y} \). \\
$\mathcal{L}_D$ & The loss between \( D \) and \( \widehat{D} \). \\
$\rho$ & The sign inconsistency ratio between \( D \) and \( \widehat{D} \). \\
\bottomrule
\end{tabularx}%
}
\end{table}

In the following, \emph{input} refers to the historical time series, while \emph{target} and \emph{prediction} represent the ground truth and predicted values of the future time series, respectively. Let \( x_t \in \mathbb{R}^N \) denote the observation at time step \( t \) of \( N \) variables. In LTSF tasks, given the input \( X = \left\{ x_{t-L+1}, x_{t-L+2}, \cdots, x_{t} \right\} \in \mathbb{R}^{L \times N} \), the model \( \mathcal{F} \) with learnable parameters \( \theta \) aims to predict the target \( Y = \{x_{t+1}, x_{t+2}, \dots, x_{t+H}\} \), where the corresponding prediction is \( \widehat{Y} = \left\{ \hat{x}_{t+1}, \hat{x}_{t+2}, \cdots, \hat{x}_{t+H} \right\} \in \mathbb{R}^{H \times N} \). Here, \( L \) is the length of the input, while \( H \) is the length of the target and the prediction. Typically, \( H \) is much larger in LTSF tasks than in traditional short-term time series forecasting tasks, such as 720 time steps. This forecast behavior can be formally summarized as \( \mathcal{F}_{\theta}: X \rightarrow \widehat{Y} \).

Our ultimate goal is to train a model that minimizes the discrepancy between the prediction \( \widehat{Y} \) and the target \( Y \).
Frequently used notations are listed in Table \ref{tab:notions}.

\subsection{Motivation}\label{TDOS_effect}
Minimizing the discrepancy between the prediction and the target requires not only achieving point-wise accuracy but also capturing the temporal dynamics inherent in the target~\cite{montgomery2015introduction}.
\begin{figure}[ht]
    \centering
    \begin{subfigure}{0.85\columnwidth}
    \captionsetup{skip=1pt}
        \includegraphics[width=\columnwidth]{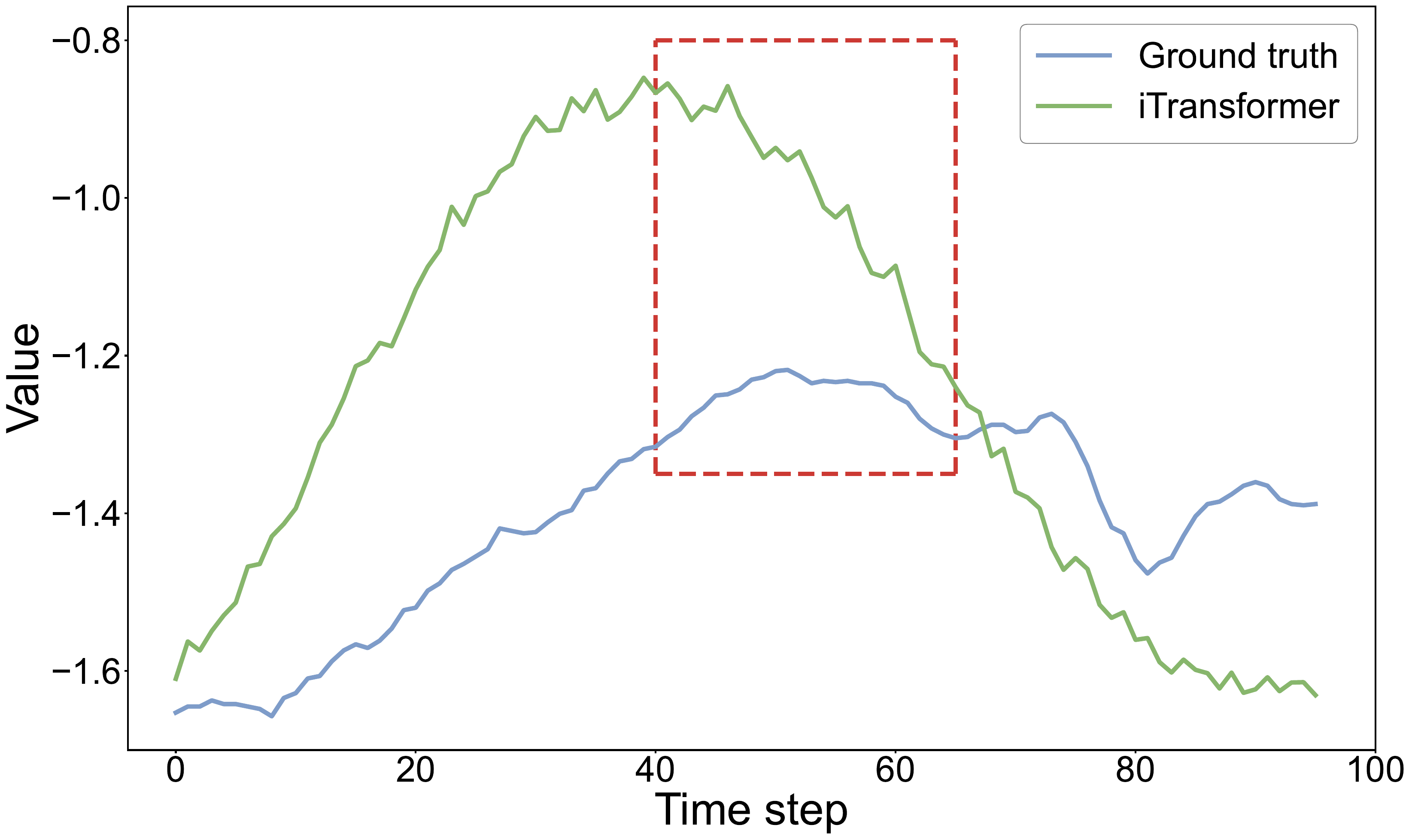}
        \caption{Comparison of the target and prediction.}
    \end{subfigure}
    \vspace{0.6mm}
    \\
    \begin{subfigure}{0.85\columnwidth}
    \captionsetup{skip=1pt}
        \includegraphics[width=\columnwidth]{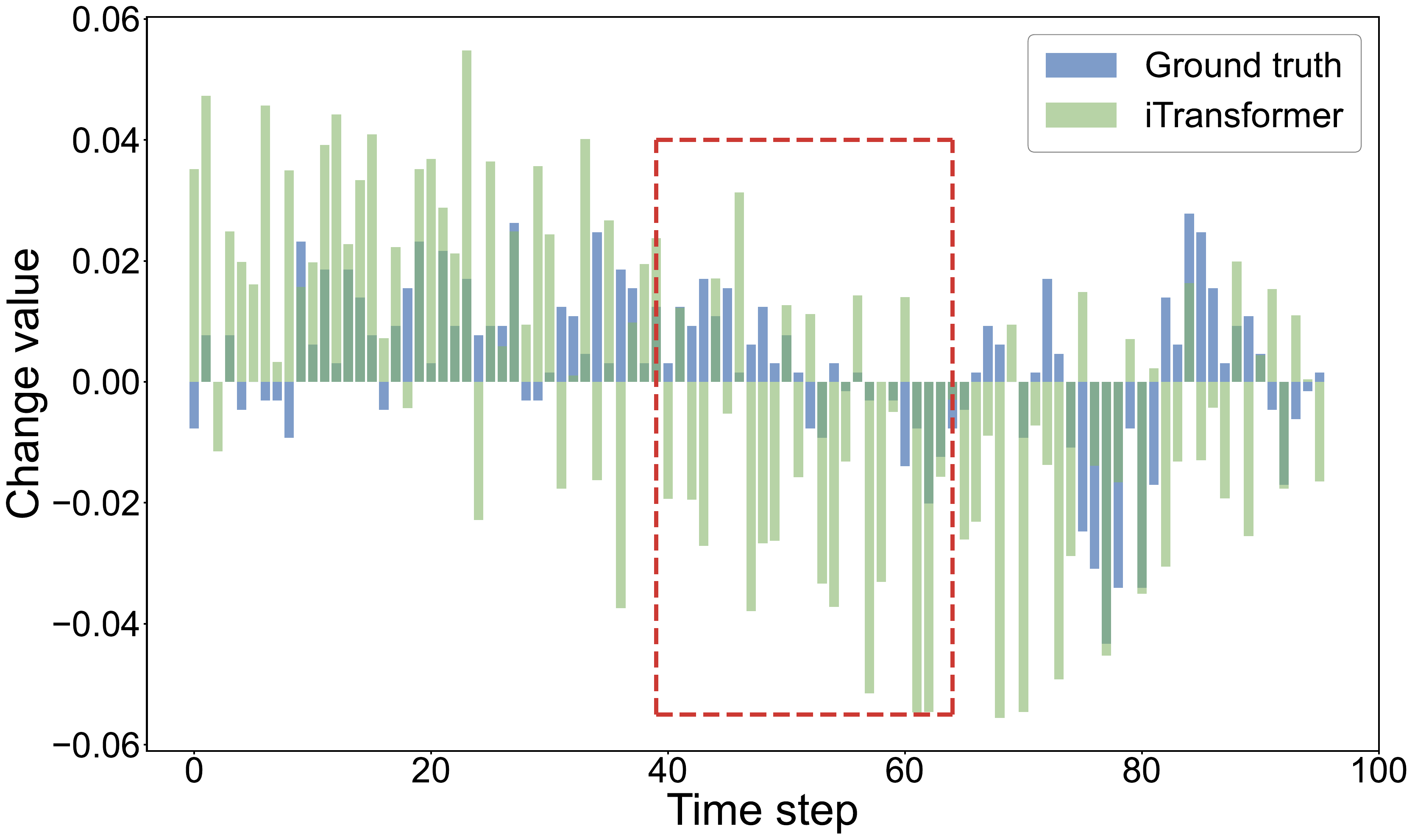}
        \caption{Comparison of change values between adjacent time steps within the target and prediction.}
    \end{subfigure}
    \caption{An illustration of a forecasting instance produced by the iTransformer \cite{liu2023itransformer} on the Weather dataset. $L = 96$, $H = 96$.}
    \label{fig:analysis}
\end{figure}

However, through an investigation of existing LTSF methods, we find that they predominantly focus on extracting temporal patterns within the input but fail to adequately model Temporal Dependencies within the Target (TDT), ultimately leading to suboptimal forecasting performance.
We present a forecasting instance from the state-of-the-art method iTransformer ~\cite{liu2023itransformer} on the Weather dataset in Figure \ref{fig:analysis}. 
As shown in Figure \ref{fig:analysis}(a), there are significant discrepancies between predicted and ground truth values at most time steps. 
Delving deeper into the underlying causes of these discrepancies, we analyze the relations between consecutive time steps. Figure \ref{fig:analysis}(b) illustrates that the method not only exhibits substantial deviations in change values between adjacent time steps but even fails to accurately predict change directions, particularly in the regions marked by red dashed boxes.
These observations demonstrate that the method lacks adequate modeling of temporal relations across multiple steps in the target, resulting in unrealistic and inaccurate predictions.

The inadequate modeling of TDT arises from the reliance on the non-autoregressive decoding strategy and a single loss function. Specifically, the autoregressive decoding strategy is unsuitable for LTSF tasks due to error accumulation and slow inference speeds inherent in step-by-step forecasting \cite{shen2023non}. To avoid these issues, methods that adopt the non-autoregressive decoding strategy have become dominant in LTSF tasks \cite{zeng2023transformers}, simultaneously generating all prediction steps \cite{nie2022time, lin2023segrnn}. However, these methods drop the dependency information in the target \cite{huang2022learning, xiao2023survey}.
Besides, regardless of the architectures and parameterizations, the vast majority of methods~\cite{wu2022timesnet, liu2023itransformer} optimize a single conventional forecasting objective $\mathcal{L}_Y$, whose general formulation can be expressed as:  
\begin{equation}\label{eq:loss_y}  
\mathcal{L}_Y = \frac{1}{H} \sum^H_{i=1} \ell(x_{t+i}, \hat{x}_{t+i}),
\end{equation}  
where \( \ell(\cdot, \cdot) \) denotes the error evaluated at a single time step. Popular choices for \( \mathcal{L}_Y \) include MSE, MAE and other variants~\cite{jadon2024comprehensive}. However, methods merely optimize the model towards the minimum of the average of point-wise prediction error with $\mathcal{L}_Y$, overlooking the temporal relations when generating prediction steps~\cite{xue2013perceptual, le2019shape}.

\subsection{Temporal Dependency Alignment}\label{OTDM}
\begin{figure*}[!ht]
    \centering
    \includegraphics[width=0.95\textwidth]{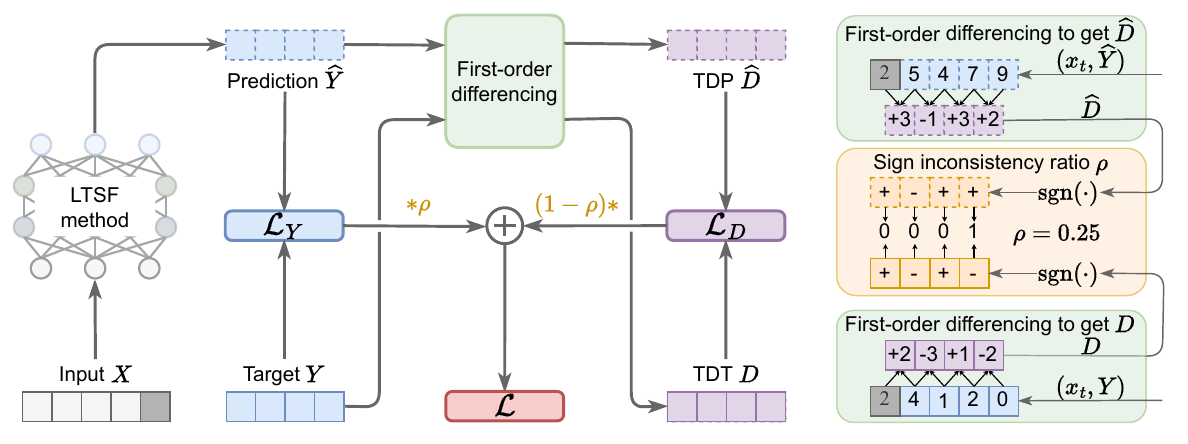}
    \caption{Overview of our proposed Temporal Dependency Alignment (TDAlign) framework. First, TDAlign formulates temporal dependencies within the target (TDT)—denoted as $D$, and temporal dependencies within the prediction (TDP)—denoted as $\widehat{D}$, using the first-order differencing values of the target $Y$ and the prediction $\widehat{Y}$, respectively (in green boxes). Then, TDAlign devises a loss $\mathcal{L}_D$ that guides the model to align $\widehat{D}$ with $D$ for capturing temporal patterns. 
    Finally, TDAlign calculates a balance weight $\rho$ (in the orange box) to adaptively control the relative contributions of the conventional forecasting loss $\mathcal{L}_Y$ and the devised $\mathcal{L}_D$, thereby improving existing methods without introducing additional learnable parameters. 
    }
    \label{fig:framework}
\end{figure*}

Motivated by the analysis of existing LTSF methods' inadequate capture of change patterns in Section \ref{TDOS_effect}, we propose a generic and novel Temporal Dependency Alignment (TDAlign) framework that can plug into existing methods to model TDT. An overview of the proposed TDAlign framework is shown in Figure~\ref{fig:framework}.

\subsubsection{Temporal Dependencies within the Target}\label{TDOS_definition}
A fair formulation of temporal dependencies should benefit both learnability and generalizability in capturing these dependencies. 
\xq{
To achieve this, we consider two key factors. First, in time series, temporally proximate time steps typically exhibit stronger correlations compared to those with far temporal distance~\cite{box2015time, hyndman2018forecasting, montgomery2015introduction}. Consequently, change patterns between adjacent time steps contain more direct and reliable information than those across longer intervals. Second, while some formulations could capture specific dependencies, such as using higher-order differences for complex relations, long-range intervals for periodic patterns~\cite{wu2022timesnet}, or frequency-domain features for global structures~\cite{wang2025fredf}, they face inherent limitations. Specifically, they often incorporate strong assumptions tailored to specific data characteristics, limiting their transferability across different domains. Moreover, their increased complexity makes them more susceptible to overfitting spurious patterns or amplifying noise, which can be detrimental to the baseline model's performance.
}

Building upon these considerations, we formulate Temporal Dependencies within the Target (TDT) using first-order differencing values \cite{dickey1987determining} of the target. This formulation captures essential local temporal dynamics while maintaining simplicity and broad applicability without imposing restrictive data-specific assumptions.
Specifically, we formulate TDT as $\textstyle D = \left\{d_{t+1},d_{t+2},\cdots,d_{t+H}\right\}\in \mathbb{R}^{H\times N}$, where each term $d_{t+i}\in \mathbb{R}^N$ equals to the change value between adjacent time steps, calculated as:
\begin{equation}\label{eq:d}
d_{t+i}=x_{t+i}-x_{t+i-1}, 1\leq i \leq H.
\end{equation}

We formulate Temporal Dependencies within the Prediction (TDP) as $\widehat{D} = \left\{\hat{d}_{t+1},\hat{d}_{t+2},\cdots,\hat{d}_{t+H}\right\}\in \mathbb{R}^{H\times N}$, where $\hat{d}_{t+i}$ is computed as:
\begin{equation}\label{eq:d_hat}
\hat{d}_{t+i} = 
\begin{cases}
\hat{x}_{t+i}-x_{t} & \text{if } i=1, \\
\hat{x}_{t+i}-\hat{x}_{t+i-1} & \text{if } 2\leq i \leq H .
\end{cases}
\end{equation}

Despite its simplicity, this formulation using first-order differencing values (i.e., change values between adjacent time steps) demonstrates greater generality and effectiveness compared to alternative formulations of temporal dependencies, such as higher-order differencing values, change values between non-adjacent time steps, or frequency domain features (see the experiments in Section~\ref{ablation} for details).

\subsubsection{A Novel Loss Function for TDAlign}\label{DL}
In order to effectively capture intricate temporal patterns of the target, a direct approach is to guide the model to align TDP with TDT. To achieve this, we propose a novel loss function, \( \mathcal{L}_D \), which measures the average error between TDP and TDT. Consistent with the conventional forecasting objective $\mathcal{L}_Y$ (Equation \ref{eq:loss_y}), the TDT learning objective \( \mathcal{L}_D \) is computed using either \text{MSE} or \text{MAE} and its general formulation is expressed as:
\begin{equation}\label{eq:loss_d}
\mathcal{L}_D = \frac{1}{H} \sum^H_{i=1}{\ell(d_{t+i},\hat{d}_{t+i})},
\end{equation}
where $\ell(\cdot,\cdot)$ denotes the error evaluated at a single time step. Notably, a key distinction is that each term in $\mathcal{L}_Y$ involves individual time steps, whereas each term in $\mathcal{L}_D$ involves two consecutive time steps.

\subsubsection{Adaptive Loss Balancing}\label{DDR}
To integrate TDAlign with forecasting models, a prevalent strategy is to introduce \xq{a weight $\rho$ to balance the relative importance between the conventional forecasting objective $\mathcal{L}_Y$ and the TDT learning objective $\mathcal{L}_D$.} Thus, the overall loss function of TDAlign can be expressed as:
\begin{equation}\label{eq:loss}
\mathcal{L} = \rho \mathcal{L}_Y + (1-\rho) \mathcal{L}_D.
\end{equation}

While manually tuning $\rho$ is a straightforward option, this process lacks flexibility and is labor-intensive. Alternatively, treating $\rho$ as a trainable parameter offers a more automated solution. However, it cannot adaptively modulate the relative contributions in response to evolving temporal dynamics within the data or the model's varying performance across different training phases. To address these limitations, we propose an adaptive loss balancing strategy using a dynamic weight $\rho$, which is updated based on the model's instantaneous performance on the TDT learning task and calculated as follows:
\begin{equation}\label{eq:loss_sgn}
    \rho = \frac{1}{H} \sum_{i=1}^{H} \mathbbm{1} (\operatorname{sgn}(d_{t+i}) \neq \operatorname{sgn}(\hat{d}_{t+i})),
\end{equation}
where $\mathbbm{1}(\cdot)$ is the indicator function that returns 1 if the condition $\operatorname{sgn}(d_{t+i}) \neq \operatorname{sgn}(\hat{d}_{t+i})$ is true and 0 otherwise, and $\operatorname{sgn}(\cdot)$ denotes the sign function. Consequently, $\rho$ represents the sign inconsistency ratio between TDP and TDT, with values ranging from 0 to 1.

The dynamic weight $\rho$ offers several advantages: (1) Simplicity and informativeness. $\rho$ is directly computed from TDP and TDT without requiring additional data or complex preprocessing. Moreover, the sign indicates the change direction between adjacent time steps. Therefore, $\rho$ can also serve as a metric to quantify the method's ability to capture TDT, as detailed in Section~\ref{metric}.
(2) Effectiveness. $\rho$ dynamically adjusts the relative importance between the conventional forecasting objective $\mathcal{L}_Y$ and the TDT learning objective $\mathcal{L}_D$ based on the model's current learning state. The effectiveness of $\rho$ is further theoretically analyzed in Section~\ref{sec:theory} and validated by experiments detailed in Section~\ref{ablation}.
(3) Flexibility. The formulation of $\rho$ introduces no additional learnable parameters, enabling TDAlign to serve as a model-agnostic framework that seamlessly integrates with various LTSF methods.

The complete training procedure of TDAlign is described in \textbf{Algorithm} \ref{alg:OTDM}.
\begin{algorithm}

\caption{Temporal Dependency Alignment} 
\label{alg:OTDM}
\KwIn{The input $X\in \mathbb{R}^{L \times N}$, the target $Y\in \mathbb{R}^{H \times N}$, the forecasting model $\mathcal{F}$ with parameters $\theta$.
}
\KwOut{Updated model parameters $\theta^\ast$}
\SetAlgoLined
\While{not converged}{
    Predict $\widehat{Y} \leftarrow \mathcal{F}_\theta(X)$\;
    Construct $D$ from $Y$ using Equation \ref{eq:d} and compute $\widehat{D}$ from $\widehat{Y}$ using Equation \ref{eq:d_hat}\;
    Compute $\mathcal{L}_Y$ and $\mathcal{L}_D$ according to Equation \ref{eq:loss_y} and Equation \ref{eq:loss_d}\ respectively\;
    Compute the weight $\rho$ using Equation \ref{eq:loss_sgn}\;
    Compute the overall loss $\mathcal{L} \leftarrow \rho \mathcal{L}_Y + (1-\rho) \mathcal{L}_D$\;
    Update parameters $\theta \leftarrow \theta - \eta \nabla_\theta \mathcal{L}$;
}
\end{algorithm}

\subsection{Computational Complexity Analysis}\label{complexity}
We analyze the additional computational complexity introduced by TDAlign in terms of both time and space. 

For time complexity, TDAlign requires several additional computations. The calculation of $D$ (Equation~\ref{eq:d}) and $\widehat{D}$ (Equation~\ref{eq:d_hat}) each incurs a time complexity of $O(H)$. The computation of $\mathcal{L}_D$ (Equation~\ref{eq:loss_d}) and $\rho$ (Equation~\ref{eq:loss_sgn}) each contributes $O(H)$ to the time complexity. The final weighted sum of $\mathcal{L}_Y$ and $\mathcal{L}_D$ (Equation~\ref{eq:loss}) adds a constant time $O(1)$. In total, the additional time complexity of TDAlign is $O(4H + 1)$, simplifying to $O(H)$ in asymptotic terms, where $H$ denotes the length of the prediction.

Regarding space complexity, TDAlign introduces only a constant overhead. It requires storing a few scalar values such as $\mathcal{L}_D$, $\rho$, and some intermediate results. Importantly, $D$ and $\widehat{D}$ are computed on-the-fly and discarded after their use in calculating $\mathcal{L}_D$ and $\rho$, ensuring no contribution to the persistent space complexity. As a result, the additional space complexity of TDAlign is $O(1)$.

In conclusion, TDAlign is an efficient plug-and-play framework, introducing minimal computational overhead that scales linearly with prediction length in time complexity, while maintaining constant space complexity. \xq{We empirically validate these efficiency claims in Section~\ref{cost}.}

\xq{
\section{Theoretical Analysis}
\label{sec:theory}

\label{sec:deficiency_correction}
For computational efficiency, conventional LTSF methods adopt non-autoregressive decoding, generating all prediction steps in a single forward pass. They assume temporal independence across prediction steps and optimize solely with a point-wise loss function \(\mathcal{L}_Y\). However, such independence assumptions fundamentally neglect the temporal correlations inherent in time series, leading to a critical paradox: while these methods may achieve low numerical errors like MSE, they often generate predictions with unrealistic high-frequency fluctuations. This behavior contradicts the temporal coherence of real-world time series, which typically evolve through smooth, locally consistent transitions rather than abrupt, erratic variations.

To formalize this limitation, we quantify temporal dependencies that \(\mathcal{L}_Y\) fails to capture. Following the maximum likelihood principle, we analyze the discrepancy \(\Psi\) between \(\mathcal{L}_Y\) and the Negative Log-Likelihood (NLL) objective under a first-order Markov assumption. This assumption is well-justified as it directly models the underlying generative process of many time series, where the current state predominantly determines the  next~\cite{box2015time}. This fundamental principle not only manifests as strong dependencies between adjacent time steps but also forms the foundational basis of autoregressive modeling.

\begin{mdframed}[linecolor=white]
    \label{thm:discrepancy}
    \textbf{\textit{Theorem 1.}}
    \textit{Under a first-order Markov assumption, where \( p(x_{t+i}|x_{<t+i}, X) \approx p(x_{t+i}|x_{t+i-1}, X) \), the discrepancy \(\Psi\) between the point-wise MSE loss, i.e., \(\mathcal{L}_Y\), and the NLL objective is given by:}
    \begin{equation}
        \label{eq:discrepancy_markov}
        \Psi = \sum_{i=2}^{H} \frac{1}{1-\phi_i^2} \left[ \phi_i^2 (\epsilon_i^2 + \epsilon_{i-1}^2) - 2\phi_i\epsilon_i\epsilon_{i-1} \right],
    \end{equation}
    \textit{where \(\epsilon_i = x_{t+i} - \hat{x}_{t+i}\) is the prediction error at step \(i\), and \(\phi_i\) is the partial autocorrelation coefficient between \(x_{t+i}\) and \(x_{t+i-1}\).}
\end{mdframed}

A detailed proof is provided in Appendix. Theorem 1 reveals a key limitation of the conventional objective $\mathcal{L}_Y$: failing to account for the temporal dependencies among time steps, mathematically quantified by the discrepancy term $\Psi$. This omission explains why models trained solely on $\mathcal{L}_Y$ tend to produce predictions with insufficient temporal coherence. However, directly optimizing $\Psi$ is practically infeasible. The primary barriers are the prohibitive computational cost and numerical instability associated with estimating the partial autocorrelation coefficients, $\phi_i$, especially for non-stationary time series~\cite{box2015time,hyndman2018forecasting}. 

Our proposed loss, $\mathcal{L}_D$, is a tractable surrogate objective that directly addresses the above limitation. Specifically, when implemented using MSE, $\mathcal{L}_D = \sum(d_{t+i} - \hat{d}_{t+i})^2 =\sum((x_{t+i} - x_{t+i-1}) - (\hat{x}_{t+i} - \hat{x}_{t+i-1}))^2= \sum(\epsilon_i - \epsilon_{i-1})^2$. By penalizing the difference between consecutive errors, $\mathcal{L}_D$ mirrors the core principle of $\Psi$ without its computational complexity. Therefore, $\mathcal{L}_D$ explicitly constrains the model to capture step-wise temporal dynamics.

To effectively integrate $\mathcal{L}_Y$ and $\mathcal{L}_D$, we propose an adaptive loss balancing strategy using $\rho$, which modulates the contribution of each loss term according to the model’s performance on the TDT learning task during training.

\begin{mdframed}[linecolor=white]
    \textbf{\textit{Proposition 1.}}
    \textit{Let the prediction error \(e_i = d_{t+i} - \hat{d}_{i+1}\) as a zero-mean Gaussian variable with variance \(\sigma_e^2\), the expected value of the adaptive weight \(\rho\) is given by:}
    \begin{equation}
        \mathbb{E}[\rho | \sigma_e^2] = \frac{1}{H}\sum_{i=1}^{H} \Phi\left(-\frac{|d_{t+i}|}{\sigma_e}\right),
    \end{equation}
    \textit{where \(\Phi(\cdot)\) is the cumulative distribution function of the standard normal distribution, \(|d_{t+i}|\) is the true change magnitude between \(x_{t+i}\) and \(x_{t+i-1}\), and \(\sigma_e\) is the standard deviation of the model's prediction error.}
\end{mdframed}

A detailed proof is provided in Appendix.
Proposition 1 reveals a direct relationship between the expected dynamic weight, $\mathbb{E}[\rho]$, and the error variance, $\sigma_e^2$, which serves as a proxy for the model's performance on the TDT learning task. This relationship dictates a dynamic balancing of the loss terms throughout training. Initially, when the model's predictions are inaccurate, the large error variance $\sigma_e^2$ leads to a higher $\mathbb{E}[\rho]$. This prioritizes $\mathcal{L}_Y$, thereby guiding the model to capture the global pattern of the target sequence. As training progresses, the model's predictive accuracy improves, causing $\sigma_e^2$ and consequently $\mathbb{E}[\rho]$ to decrease. This reduction gradually shifts the optimization focus towards $\mathcal{L}_D$, encouraging the refinement of predictions by capturing fine-grained local dynamics.
In essence, $\rho$ dynamically steers the training focus from learning coarse, global patterns toward capturing fine-grained, local details. This adaptive loss balancing facilitates the model to build richer sequence representations, thereby enhancing its generalization capabilities.
}

\section{Experiments}
In this section, we conduct extensive experiments to demonstrate the effectiveness and generalizability of our proposed TDAlign. We begin with a detailed description of the experimental setup. Subsequently, we present the main experimental results and statistics of improvements achieved by integrating TDAlign into baseline methods. Furthermore, we conduct ablation studies to evaluate the effectiveness of various components of TDAlign and assess the impact of the formulation of temporal dependencies. Additionally, we perform additional investigations into the generalizability and robustness of TDAlign. Finally, we showcase several long-term forecasting instances to demonstrate that TDAlign produces more accurate and realistic predictions compared to baseline methods.

\subsection{Experimental Setup}\label{exp_set}
\subsubsection{Datasets} 
We conduct experiments on seven widely used LTSF benchmark datasets. These multivariate datasets cover various real-world application scenarios, including:
\begin{itemize}
\item ETT (Electricity Transformer Temperature)\footnote{\url{https://github.com/zhouhaoyi/ETDataset}}: This dataset contains the oil temperature and six power load features from electricity transformers from July 2016 to July 2018. It can be further divided into four sub-datasets (ETTh1, ETTh2, ETTm1, and ETTm2) based on data collection granularity and regions.
\item Electricity\footnote{\url{https://archive.ics.uci.edu/ml/datasets/ElectricityLoadDiagrams20112014}}: This dataset collects the electricity consumption of 321 clients from 2012 to 2014.
\item Weather\footnote{\url{https://www.bgc-jena.mpg.de/wetter/}}: This dataset records various climate indicators such as temperature, humidity, and wind speed for 2020 in Germany.
\item ILI (Influenza-Like Illness)\footnote{\url{https://gis.cdc.gov/grasp/fluview/fluportaldashboard.html}}: This dataset describes the ratio of influenza-like illness patients versus the total patients from the Centers for Disease Control and Prevention of the United States from 2002 to 2021.
\end{itemize}
The statistics of these datasets is summarized in Table \ref{tab:datasets}.
\begin{table}[ht]
    \centering
    \caption{Statistics of benchmark datasets.}
    \label{tab:datasets}
        \begin{tabular}{c|ccccccc} 
            \toprule
            Dataset & Variables & Timesteps & Granularity \\ 
            \midrule
            ETTh1       & 7         & 17420       & 1 hour      \\
            ETTh2       & 7         & 17420       & 1 hour      \\
            ETTm1       & 7         & 69680       & 15 minutes  \\
            ETTm2       & 7         & 69680       & 15 minutes  \\
            Weather     & 21        & 52696       & 10 minutes  \\
            Electricity & 321       & 26304       & 1 hour      \\
            ILI         & 7         & 966         & 1 week      \\
            \bottomrule
        \end{tabular}
\end{table}
\begin{table*}[!ht]
\centering
\caption{Input and prediction lengths ($L$ and $H$) of various baseline methods on different datasets.}
\label{tab:configurations}
\begin{tabular}{c|cc} 
\toprule
\multirow{2}{*}{Method} & ILI dataset                       & Other datasets                           \\ 
                             & $L \rightarrow H$                    & $L \rightarrow H$                        \\ 
\midrule
SegRNN & 60 $\rightarrow$ \{24, 36, 48, 60\} & 720 $\rightarrow$ \{96, 192, 336, 720\} \\
DLinear, PatchTST   & 104 $\rightarrow$ \{24, 36, 48, 60\} & 336 $\rightarrow$ \{96, 192, 336, 720\}  \\
iTransformer, MICN, TimesNet & 36 $\rightarrow$ \{24, 36, 48, 60\}  & 96 $\rightarrow$ \{96, 192, 336, 720\}   \\
\bottomrule
\end{tabular}
\end{table*}

\subsubsection{Baselines}
We consider six open-source state-of-the-art LTSF methods as baseline methods, comprising a wide variety of architectures:
\begin{itemize}
\item DLinear\cite{zeng2023transformers}\footnote{\url{https://github.com/cure-lab/LTSF-Linear}}: An MLP-based method that employs two linear layers to extract trend and seasonal features.
\item MICN\cite{wang2022micn}\footnote{\url{https://github.com/wanghq21/MICN}}: A CNN-based method that adopts down-sampled convolution and isometric convolution to capture local and global temporal features. 
\item TimesNet\cite{wu2022timesnet}\footnote{\url{https://github.com/thuml/TimesNet}}: A CNN-based method that converts 1D time series into 2D tensors and captures temporal intraperiod- and interperiod- variations using deep convolutional neural networks.

\item PatchTST\cite{nie2022time}\footnote{\url{https://github.com/yuqinie98/PatchTST}}: A Transformer-based method that leverages the patching structure to capture local semantic information, advancing both the performance and efficiency of Transformers for LTSF.

\item iTransformer\cite{liu2023itransformer}\footnote{\url{https://github.com/thuml/iTransformer}}: A Transformer-based method that embeds time series as variate tokens, captures multivariate correlations through self-attention and extracts temporal dependencies using Transformer's feed-forward network.

\item SegRNN\cite{lin2023segrnn}\footnote{\url{https://github.com/lss-1138/SegRNN}}: An RNN-based method that reintroduces RNN architectures to LTSF through segment-wise iterations and parallel multi-step forecasting. During the decoding phase, it incorporates position encodings to implicitly learn correlations between segments.
\end{itemize}

\begingroup
\setlength{\tabcolsep}{1mm} 
\begin{table*}[!htb]
    \centering
    \fontsize{9pt}{10pt}\selectfont
    \caption{Comparison of forecasting errors between baseline methods and TDAlign. Results are averaged over 5 runs, with better performance highlighted in bold. \text{“Avg error”} represents the average error of each baseline method over 7 datasets.}
    \label{tab:main_results}
    \resizebox{\textwidth}{!}{

\begin{tabular}{cc|cccc|cccc|cccc}
\toprule
\multicolumn{2}{c|}{Architecture}&\multicolumn{8}{c|}{CNN-based}	&\multicolumn{4}{c}{MLP-based} \\  
\midrule
\multicolumn{2}{c|}{Method}&\multicolumn{2}{c}{MICN}	&\multicolumn{2}{c}{+TDAlign}	&\multicolumn{2}{c}{TimesNet}	&\multicolumn{2}{c|}{+TDAlign}	&\multicolumn{2}{c}{DLinear}	&\multicolumn{2}{c}{+TDAlign}	\\
\multicolumn{2}{c|}{Metric}&MSE	&MAE	&MSE	&\multicolumn{1}{c}{MAE}	&MSE	&MAE	&MSE	&\multicolumn{1}{c|}{MAE}	&MSE	&MAE	&MSE	&MAE	\\
\midrule
\multirow{4}{*}{\myrotcell{ETTh1}} & 96  &0.417±0.014&0.442±0.011&\textbf{0.375±0.001}&\textbf{0.398±0.001}&0.406±0.011&0.424±0.007&\textbf{0.384±0.003}&\textbf{0.396±0.001}&0.376±0.005&0.399±0.006&\textbf{0.362±0.000}&\textbf{0.384±0.000}\\
        & 192 &0.483±0.010&0.485±0.005&\textbf{0.416±0.006}&\textbf{0.426±0.002}&0.458±0.012&0.453±0.007&\textbf{0.438±0.012}&\textbf{0.429±0.007}&0.407±0.003&0.416±0.003&\textbf{0.400±0.000}&\textbf{0.407±0.000}\\
        & 336 &0.581±0.037&0.559±0.021&\textbf{0.445±0.003}&\textbf{0.448±0.002}&0.508±0.017&0.479±0.010&\textbf{0.474±0.011}&\textbf{0.449±0.007}&0.490±0.041&0.485±0.033&\textbf{0.431±0.000}&\textbf{0.427±0.000}\\
        & 720 &0.689±0.092&0.633±0.049&\textbf{0.533±0.008}&\textbf{0.524±0.005}&0.516±0.010&0.493±0.006&\textbf{0.499±0.012}&\textbf{0.481±0.009}&0.498±0.021&0.508±0.015&\textbf{0.452±0.000}&\textbf{0.473±0.000}\\
\midrule
\multirow{4}{*}{\myrotcell{ETTh2}} & 96  &0.313±0.012&0.369±0.013&\textbf{0.294±0.002}&\textbf{0.347±0.002}&0.321±0.006&0.363±0.003&\textbf{0.305±0.007}&\textbf{0.348±0.005}&0.294±0.006&0.357±0.005&\textbf{0.278±0.000}&\textbf{0.334±0.000}\\
        & 192 &0.422±0.015&0.445±0.011&\textbf{0.384±0.005}&\textbf{0.406±0.003}&0.411±0.011&0.417±0.007&\textbf{0.390±0.006}&\textbf{0.397±0.004}&0.376±0.005&0.412±0.005&\textbf{0.354±0.000}&\textbf{0.386±0.000}\\
        & 336 &0.693±0.090&0.584±0.046&\textbf{0.446±0.008}&\textbf{0.451±0.003}&0.467±0.016&0.460±0.011&\textbf{0.429±0.012}&\textbf{0.433±0.009}&0.458±0.010&0.466±0.005&\textbf{0.411±0.000}&\textbf{0.430±0.000}\\
        & 720 &0.853±0.046&0.667±0.023&\textbf{0.635±0.025}&\textbf{0.563±0.009}&0.466±0.025&0.469±0.015&\textbf{0.422±0.012}&\textbf{0.437±0.007}&0.714±0.019&0.601±0.006&\textbf{0.582±0.000}&\textbf{0.531±0.000}\\
\midrule
\multirow{4}{*}{\myrotcell{ETTm1}} & 96  &0.312±0.003&0.363±0.004&\textbf{0.301±0.002}&\textbf{0.339±0.001}&0.335±0.004&0.375±0.002&\textbf{0.331±0.004}&\textbf{0.361±0.003}&0.301±0.001&0.345±0.002&\textbf{0.292±0.000}&\textbf{0.333±0.000}\\
        & 192 &0.358±0.004&0.393±0.006&\textbf{0.348±0.003}&\textbf{0.369±0.004}&0.395±0.011&0.405±0.007&\textbf{0.376±0.006}&\textbf{0.385±0.005}&0.336±0.001&0.367±0.001&\textbf{0.331±0.000}&\textbf{0.356±0.000}\\
        & 336 &0.401±0.006&0.424±0.007&\textbf{0.382±0.005}&\textbf{0.396±0.006}&0.420±0.009&0.423±0.003&\textbf{0.402±0.002}&\textbf{0.401±0.002}&0.371±0.003&0.389±0.004&\textbf{0.365±0.000}&\textbf{0.376±0.000}\\
        & 720 &0.470±0.013&0.473±0.012&\textbf{0.451±0.008}&\textbf{0.445±0.009}&0.486±0.010&0.457±0.004&\textbf{0.463±0.002}&\textbf{0.434±0.002}&0.426±0.002&0.422±0.003&\textbf{0.420±0.000}&\textbf{0.411±0.000}\\
\midrule
\multirow{4}{*}{\myrotcell{ETTm2}} & 96  &0.176±0.000&0.273±0.001&\textbf{0.171±0.001}&\textbf{0.259±0.001}&0.187±0.002&0.266±0.001&\textbf{0.181±0.002}&\textbf{0.257±0.001}&0.170±0.002&0.264±0.002&\textbf{0.164±0.000}&\textbf{0.247±0.000}\\
        & 192 &0.266±0.011&0.342±0.011&\textbf{0.233±0.003}&\textbf{0.303±0.003}&0.254±0.002&0.309±0.001&\textbf{0.246±0.002}&\textbf{0.299±0.001}&0.229±0.005&0.306±0.007&\textbf{0.221±0.000}&\textbf{0.289±0.000}\\
        & 336 &0.321±0.025&0.377±0.023&\textbf{0.294±0.006}&\textbf{0.345±0.005}&0.316±0.005&0.346±0.003&\textbf{0.309±0.002}&\textbf{0.338±0.001}&0.288±0.005&0.349±0.007&\textbf{0.278±0.000}&\textbf{0.332±0.000}\\
        & 720 &0.466±0.033&0.468±0.021&\textbf{0.391±0.013}&\textbf{0.408±0.010}&0.425±0.005&0.408±0.002&\textbf{0.417±0.006}&\textbf{0.401±0.003}&0.438±0.072&0.442±0.042&\textbf{0.377±0.000}&\textbf{0.399±0.000}\\
\midrule
\multirow{4}{*}{\myrotcell{Electricity}} & 96  &0.161±0.002&0.268±0.002&\textbf{0.156±0.001}&\textbf{0.258±0.001}&\textbf{0.168±0.003}&0.272±0.003&0.171±0.003&\textbf{0.267±0.003}&\textbf{0.140±0.000}&0.237±0.000&0.141±0.000&\textbf{0.235±0.000}\\
        & 192 &0.176±0.002&0.283±0.002&\textbf{0.173±0.003}&\textbf{0.272±0.002}&0.186±0.003&0.287±0.003&\textbf{0.182±0.003}&\textbf{0.278±0.003}&0.154±0.000&0.250±0.000&\textbf{0.154±0.000}&\textbf{0.247±0.000}\\
        & 336 &0.196±0.004&0.306±0.005&\textbf{0.184±0.004}&\textbf{0.284±0.004}&\textbf{0.200±0.005}&\textbf{0.301±0.005}&0.236±0.024&0.313±0.016&\textbf{0.169±0.000}&0.268±0.000&0.170±0.000&\textbf{0.263±0.000}\\
        & 720 &0.216±0.004&0.324±0.005&\textbf{0.212±0.006}&\textbf{0.308±0.004}&\textbf{0.231±0.008}&\textbf{0.324±0.006}&0.292±0.008&0.354±0.007&0.204±0.000&0.301±0.000&\textbf{0.204±0.000}&\textbf{0.294±0.000}\\
\midrule
\multirow{4}{*}{\myrotcell{ILI}} & 24  &2.760±0.025&1.151±0.008&\textbf{2.546±0.038}&\textbf{1.059±0.010}&2.336±0.359&0.939±0.044&\textbf{2.112±0.192}&\textbf{0.901±0.035}&2.265±0.064&1.048±0.027&\textbf{2.174±0.003}&\textbf{1.001±0.001}\\
        & 36  &2.761±0.049&1.140±0.013&\textbf{2.615±0.046}&\textbf{1.072±0.011}&2.493±0.074&0.994±0.017&\textbf{2.167±0.116}&\textbf{0.905±0.024}&2.251±0.107&1.064±0.049&\textbf{2.091±0.003}&\textbf{0.995±0.001}\\
        & 48  &2.778±0.056&1.133±0.015&\textbf{2.680±0.043}&\textbf{1.078±0.009}&2.379±0.069&0.948±0.019&\textbf{2.168±0.040}&\textbf{0.889±0.009}&2.302±0.021&1.082±0.007&\textbf{2.064±0.006}&\textbf{1.000±0.001}\\
        & 60  &2.879±0.047&1.150±0.010&\textbf{2.708±0.028}&\textbf{1.085±0.006}&2.196±0.100&0.962±0.024&\textbf{2.113±0.087}&\textbf{0.895±0.020}&2.435±0.024&1.120±0.009&\textbf{2.124±0.009}&\textbf{1.029±0.003}\\
\midrule
\multirow{4}{*}{\myrotcell{Weather}} & 96  &0.170±0.003&0.235±0.006&\textbf{0.158±0.001}&\textbf{0.198±0.001}&0.172±0.002&0.222±0.001&\textbf{0.170±0.001}&\textbf{0.214±0.002}&0.175±0.001&0.238±0.003&\textbf{0.173±0.000}&\textbf{0.217±0.000}\\
        & 192 &0.222±0.006&0.284±0.007&\textbf{0.207±0.001}&\textbf{0.247±0.001}&0.235±0.004&0.273±0.004&\textbf{0.225±0.003}&\textbf{0.261±0.002}&0.216±0.001&0.274±0.001&\textbf{0.213±0.000}&\textbf{0.255±0.000}\\
        & 336 &0.278±0.014&0.339±0.020&\textbf{0.252±0.004}&\textbf{0.282±0.003}&0.285±0.003&0.306±0.002&\textbf{0.281±0.003}&\textbf{0.300±0.003}&0.262±0.002&0.313±0.003&\textbf{0.258±0.000}&\textbf{0.292±0.000}\\
        & 720 &0.336±0.014&0.376±0.017&\textbf{0.311±0.002}&\textbf{0.328±0.003}&0.360±0.002&0.355±0.001&\textbf{0.355±0.002}&\textbf{0.349±0.001}&0.325±0.001&0.364±0.001&\textbf{0.320±0.000}&\textbf{0.342±0.000}\\
\midrule
\multicolumn{2}{c|}{Avg error} & 0.720 & 0.510 & \textbf{0.654} & \textbf{0.461} & 0.629 & 0.455 & \textbf{0.591} &\textbf{0.435} &0.610 &	0.467 &	\textbf{0.564} &	\textbf{0.439} \\
\midrule
\multicolumn{2}{c|}{Architecture} & \multicolumn{8}{c|}{Transformer-based} & \multicolumn{4}{c}{RNN-based} \\
\midrule
\multicolumn{2}{c|}{Method} & \multicolumn{2}{c}{PatchTST} & \multicolumn{2}{c}{+TDAlign} & \multicolumn{2}{c}{iTransformer} & \multicolumn{2}{c|}{+TDAlign} & \multicolumn{2}{c}{SegRNN} & \multicolumn{2}{c}{+TDAlign} \\
\multicolumn{2}{c|}{Metric} & MSE & MAE & MSE & \multicolumn{1}{c}{MAE} & MSE & MAE & MSE & \multicolumn{1}{c|}{MAE} & MSE & MAE & MSE & MAE \\

\midrule
\multirow{4}{*}{\myrotcell{ETTh1}} & 96  &0.380±0.001&0.402±0.001&\textbf{0.373±0.001}&\textbf{0.396±0.001}&0.387±0.001&0.405±0.001&\textbf{0.375±0.001}&\textbf{0.389±0.000}&0.351±0.001&0.382±0.000&\textbf{0.349±0.000}&\textbf{0.380±0.000}\\
        & 192 &0.412±0.000&0.420±0.000&\textbf{0.409±0.001}&\textbf{0.417±0.001}&0.441±0.002&0.436±0.001&\textbf{0.426±0.001}&\textbf{0.420±0.001}&0.391±0.001&0.407±0.001&\textbf{0.384±0.001}&\textbf{0.403±0.000}\\
        & 336 &0.436±0.003&0.437±0.003&\textbf{0.433±0.001}&\textbf{0.431±0.001}&0.491±0.004&0.461±0.003&\textbf{0.469±0.001}&\textbf{0.442±0.001}&0.423±0.006&0.426±0.004&\textbf{0.409±0.002}&\textbf{0.418±0.001}\\
        & 720 &0.453±0.003&0.469±0.002&\textbf{0.433±0.002}&\textbf{0.453±0.002}&0.511±0.010&0.494±0.006&\textbf{0.460±0.001}&\textbf{0.460±0.001}&0.446±0.007&0.456±0.005&\textbf{0.445±0.010}&\textbf{0.452±0.006}\\
\midrule
\multirow{4}{*}{\myrotcell{ETTh2}} & 96  &\textbf{0.275±0.001}&0.338±0.001&0.277±0.001&\textbf{0.332±0.000}&0.300±0.002&0.351±0.001&\textbf{0.290±0.002}&\textbf{0.337±0.001}&0.274±0.001&0.329±0.001&\textbf{0.273±0.003}&\textbf{0.327±0.001}\\
        & 192 &\textbf{0.340±0.001}&0.380±0.001&0.347±0.002&\textbf{0.376±0.001}&0.380±0.002&0.399±0.001&\textbf{0.367±0.001}&\textbf{0.386±0.000}&0.340±0.002&0.373±0.001&\textbf{0.333±0.003}&\textbf{0.368±0.002}\\
        & 336 &\textbf{0.364±0.003}&0.398±0.001&0.367±0.004&\textbf{0.395±0.002}&0.423±0.002&0.432±0.001&\textbf{0.417±0.001}&\textbf{0.424±0.001}&0.386±0.006&0.409±0.004&\textbf{0.368±0.009}&\textbf{0.397±0.004}\\
        & 720 &0.390±0.002&0.427±0.002&\textbf{0.388±0.003}&\textbf{0.420±0.002}&0.431±0.004&0.448±0.002&\textbf{0.419±0.001}&\textbf{0.438±0.000}&0.406±0.004&0.432±0.003&\textbf{0.384±0.003}&\textbf{0.417±0.002}\\
\midrule
\multirow{4}{*}{\myrotcell{ETTm1}} & 96  &0.290±0.001&0.342±0.001&\textbf{0.287±0.004}&\textbf{0.327±0.003}&0.341±0.003&0.376±0.002&\textbf{0.329±0.001}&\textbf{0.355±0.001}&0.284±0.001&0.335±0.001&\textbf{0.278±0.001}&\textbf{0.333±0.001}\\
        & 192 &\textbf{0.332±0.002}&0.369±0.001&0.336±0.006&\textbf{0.358±0.001}&0.381±0.001&0.394±0.001&\textbf{0.381±0.001}&\textbf{0.381±0.001}&0.320±0.001&0.360±0.001&\textbf{0.317±0.001}&\textbf{0.359±0.001}\\
        & 336 &0.366±0.001&0.391±0.001&\textbf{0.360±0.006}&\textbf{0.378±0.003}&0.420±0.001&0.419±0.001&\textbf{0.418±0.002}&\textbf{0.407±0.001}&0.351±0.001&0.383±0.001&\textbf{0.346±0.002}&\textbf{0.381±0.001}\\
        & 720 &0.417±0.003&0.423±0.002&\textbf{0.413±0.005}&\textbf{0.410±0.002}&\textbf{0.489±0.001}&0.457±0.001&0.489±0.002&\textbf{0.448±0.001}&0.410±0.003&0.417±0.001&\textbf{0.399±0.002}&\textbf{0.414±0.001}\\
\midrule
\multirow{4}{*}{\myrotcell{ETTm2}} & 96  &\textbf{0.164±0.001}&0.253±0.001&0.164±0.003&\textbf{0.246±0.002}&0.184±0.000&0.269±0.001&\textbf{0.183±0.000}&\textbf{0.263±0.000}&0.158±0.000&0.241±0.000&\textbf{0.156±0.000}&\textbf{0.240±0.000}\\
        & 192 &0.221±0.001&0.292±0.001&\textbf{0.220±0.002}&\textbf{0.286±0.002}&0.252±0.001&0.313±0.001&\textbf{0.250±0.000}&\textbf{0.306±0.000}&0.215±0.001&0.283±0.000&\textbf{0.211±0.001}&\textbf{0.280±0.001}\\
        & 336 &0.276±0.001&0.329±0.001&\textbf{0.269±0.004}&\textbf{0.321±0.002}&0.315±0.001&0.352±0.001&\textbf{0.311±0.000}&\textbf{0.345±0.000}&0.263±0.001&0.317±0.000&\textbf{0.260±0.002}&\textbf{0.315±0.002}\\
        & 720 &0.367±0.001&0.384±0.001&\textbf{0.361±0.002}&\textbf{0.379±0.003}&0.413±0.001&0.406±0.001&\textbf{0.412±0.001}&\textbf{0.401±0.001}&0.333±0.001&0.369±0.001&\textbf{0.329±0.002}&\textbf{0.367±0.001}\\
\midrule
\multirow{4}{*}{\myrotcell{Electricity}} & 96  &0.130±0.000&0.223±0.000&\textbf{0.130±0.000}&\textbf{0.216±0.000}&0.148±0.000&0.240±0.000&\textbf{0.148±0.000}&\textbf{0.234±0.000}&0.129±0.000&0.220±0.000&\textbf{0.125±0.000}&\textbf{0.214±0.000}\\
        & 192 &0.149±0.001&0.241±0.001&\textbf{0.147±0.000}&\textbf{0.232±0.000}&0.165±0.001&0.256±0.001&\textbf{0.164±0.001}&\textbf{0.248±0.001}&0.150±0.000&0.240±0.000&\textbf{0.148±0.001}&\textbf{0.237±0.001}\\
        & 336 &0.166±0.000&0.260±0.000&\textbf{0.162±0.000}&\textbf{0.247±0.000}&0.178±0.001&0.270±0.001&\textbf{0.177±0.001}&\textbf{0.263±0.001}&0.167±0.000&0.258±0.000&\textbf{0.166±0.000}&\textbf{0.257±0.000}\\
        & 720 &0.204±0.001&0.293±0.001&\textbf{0.197±0.000}&\textbf{0.279±0.000}&0.214±0.006&0.302±0.005&\textbf{0.210±0.001}&\textbf{0.292±0.000}&0.201±0.001&0.290±0.000&\textbf{0.200±0.001}&\textbf{0.289±0.001}\\
\midrule
\multirow{4}{*}{\myrotcell{ILI}} & 24  &2.010±0.112&0.896±0.048&\textbf{1.826±0.083}&\textbf{0.826±0.029}&2.356±0.022&0.952±0.008&\textbf{2.173±0.182}&\textbf{0.901±0.026}&\textbf{1.748±0.049}&\textbf{0.764±0.015}&1.756±0.090&0.790±0.026\\
        & 36  &1.886±0.062&0.888±0.032&\textbf{1.783±0.087}&\textbf{0.846±0.032}&2.238±0.034&0.960±0.006&\textbf{1.982±0.048}&\textbf{0.871±0.010}&1.700±0.063&\textbf{0.772±0.018}&\textbf{1.636±0.061}&0.773±0.009\\
        & 48  &1.780±0.108&0.898±0.027&\textbf{1.767±0.144}&\textbf{0.859±0.039}&2.268±0.057&0.971±0.015&\textbf{2.019±0.011}&\textbf{0.872±0.003}&1.735±0.021&\textbf{0.776±0.009}&\textbf{1.719±0.068}&0.799±0.015\\
        & 60  &1.794±0.056&0.905±0.011&\textbf{1.780±0.068}&\textbf{0.862±0.017}&2.289±0.026&0.987±0.004&\textbf{2.058±0.020}&\textbf{0.889±0.005}&1.709±0.051&\textbf{0.788±0.015}&\textbf{1.698±0.040}&0.795±0.009\\
\midrule
\multirow{4}{*}{\myrotcell{Weather}} & 96  &0.151±0.000&0.199±0.000&\textbf{0.150±0.000}&\textbf{0.185±0.000}&0.175±0.001&0.216±0.001&\textbf{0.168±0.001}&\textbf{0.202±0.001}&0.142±0.000&\textbf{0.182±0.000}&\textbf{0.142±0.000}&0.183±0.001\\
        & 192 &0.195±0.001&0.241±0.001&\textbf{0.194±0.000}&\textbf{0.228±0.000}&0.224±0.001&0.257±0.001&\textbf{0.219±0.001}&\textbf{0.247±0.001}&0.186±0.000&0.227±0.000&\textbf{0.186±0.000}&\textbf{0.226±0.000}\\
        & 336 &0.248±0.001&0.282±0.001&\textbf{0.247±0.001}&\textbf{0.269±0.000}&0.281±0.001&0.299±0.001&\textbf{0.277±0.001}&\textbf{0.291±0.000}&0.236±0.000&0.268±0.000&\textbf{0.235±0.000}&\textbf{0.267±0.000}\\
        & 720 &\textbf{0.321±0.001}&0.335±0.001&0.322±0.000&\textbf{0.324±0.000}&0.358±0.001&0.350±0.001&\textbf{0.356±0.001}&\textbf{0.343±0.000}&0.309±0.000&\textbf{0.319±0.000}&\textbf{0.308±0.001}&0.320±0.001\\
    \midrule
    \multicolumn{2}{c|}{Avg error} & 0.518 &0.418 & \textbf{0.505} &	\textbf{0.404} & 0.609 &	0.445 &	\textbf{0.570} &	\textbf{0.423} & 0.492 &	0.394 &	\textbf{0.484} &	\textbf{0.393} \\
\bottomrule
\end{tabular}

    }

\end{table*}
\endgroup
\subsubsection{Evaluation Metrics}\label{metric}
Following the evaluation metrics employed in baseline methods, we assess the forecasting error of the prediction using \text{MSE} and \text{MAE}. Beyond the conventional evaluation, we extend our assessment to the forecasting error of TDP, also using \text{MSE} and \text{MAE} but denoted as $\text{MSE}_D$ and $\text{MAE}_D$ for distinction. Additionally, we also use $\rho$ (Equation~\ref{eq:loss_sgn}) to evaluate the forecasting error of TDP, as defined in Equation~\ref{eq:loss_sgn}. As described in Section~\ref{DDR}, $\rho$ represents the sign inconsistency ratio between TDP and TDT, and can also indicate the method's error rate in predicting the change direction between adjacent time steps. Therefore, unless otherwise specified, \text{MSE} and \text{MAE} refer to the errors between $Y$ and $\widehat{Y}$ , whereas $\text{MSE}_D$, $\text{MAE}_D$, and $\rho$ specifically represent the errors between $D$ and $\widehat{D}$. All these metrics are non-negative, with lower values indicating better performance.

\subsubsection{Implementation Details}
We conduct fair experiments to evaluate and demonstrate the effectiveness and flexibility of our proposed TDAlign.
First, we strictly adhere to the consistent preprocessing procedures and data loading parameters for each baseline method to ensure result comparability. Based on the statistics of the training set, we apply z-score normalization to standardize the data. For dataset partitioning, we use a 6:2:2 ratio to split the training, validation, and test sets for the four ETT datasets, while the other datasets follow a 7:2:1 splitting ratio. Second, we fully adopt the parameter configurations for baseline methods as described in their official code repositories. Although all baseline methods maintain a consistent prediction length for each dataset, their input lengths vary. In Table \ref{tab:configurations}, we summarize the specific configurations of input and prediction lengths of each baseline method on different datasets. Third, it is worth emphasizing that we do not employ the \textit{Drop Last} operation during testing for all experiments, as suggested by~\cite{qiu2024tfb}. Finally, all experiments are implemented using PyTorch and conducted on the NVIDIA RTX 4090 GPU with 24GB memory. Each experiment is repeated five times with different random seeds, and the mean and standard deviation of the results are reported.

\begin{table*}[!htb]
    \centering
    \caption{
    Averaged performance improvement of TDAlign over 7 datasets, derived from Table~\ref{tab:main_results} and Appendix Tables.
    Values in parentheses indicate the count of cases showing improvement. For each metric, ``Avg improvement" represents the average improvement over 6 baseline methods, and ``Count" indicates the total number of cases showing a strict improvement (excluding ties).
}
    \label{tab:count}
\begin{tabular}{c|ccccc}
\toprule
      Method  & $\Delta\text{MSE}$ & $\Delta\text{MAE}$ & $\Delta\text{MSE}_D$ & $\Delta\text{MAE}_D$ & $\Delta\rho$ \\
\midrule
        MICN &  +9.19\% (28) &  +9.72\% (28) &  +9.60\% (26) &  +7.70\% (28) &  +9.03\% (28) \\
    TimesNet &  +6.10\% (25) &  +4.38\% (26) &  +15.78\% (27) & +11.99\% (28) & +11.32\% (28) \\
    DLinear &  +7.42\% (24) &  +6.13\% (28) &  +4.57\% (18) &  +4.80\% (26) &  +4.56\% (26) \\
    PatchTST &  +2.58\% (21) &  +3.56\% (28) &  +9.05\% (24) &  +8.71\% (28) &  +9.73\% (28) \\
iTransformer &  +6.49\% (25) &  +4.95\% (28) &  +11.68\% (26) & +11.45\% (28) & +12.17\% (28) \\
      SegRNN &  +1.47\% (25) &  +0.20\% (22) &  +5.72\% (24) &  +3.79\% (26) &  +6.11\% (28) \\
         Avg improvement (Count) & +5.54\% (148) & +4.82\% (160) & +9.40\% (145) & +8.07\% (164) & +8.82\% (166) \\
\bottomrule
\end{tabular}
\end{table*}


\begin{table}[ht]
    \centering
    \caption{\xq{Averaged performance improvement of TDAlign over 6 baseline methods, derived from Table~\ref{tab:main_results}. A smaller ADF test statistic indicates a more stationary time series dataset.}}
    \label{tab:stationary}
        \begin{tabular}{c|ccccccc} 
            \toprule
            Dataset & ADF test statistic & $\Delta$MSE & $\Delta$MAE \\ 
            \midrule
            Weather     & -26.68       & +2.28\%       & +5.58\%  \\
            ETTm1       & -14.98       & +2.13\%      & +3.59\%  \\
            Electricity & -8.44        & -0.70\%      & +2.11\%      \\
            ETTh1       & -5.91        & +6.29\%       & +5.60\%      \\
            ETTm2       & -5.66        & +3.65\%       & +3.92\%  \\
            ILI         & -5.33        & +6.39\%       & +5.26\%     \\
            ETTh2       & -4.13        & +6.85\%       & +5.49\%      \\
            \bottomrule
        \end{tabular}
\end{table}

\subsection{Main Results and Analysis}\label{results}
We conducted extensive experiments encompassing seven real-world benchmark datasets, six baseline methods, and four different prediction lengths for each dataset, resulting in 168 experimental configurations.

Table~\ref{tab:main_results} presents a detailed comparison of the forecasting errors ($\text{MSE}$ and $\text{MAE}$) between baseline methods and TDAlign. As shown in this table, first, we can clearly observe that our proposed TDAlign framework improves baseline methods by a significant margin in most cases, showcasing its effectiveness and generalizability. Notably, TDAlign improves MICN's performance across all prediction lengths and datasets. Second, TDAlign contributes to more stable performance, where it shows smaller standard deviations compared to baseline methods in most cases. For instance, DLinear shows non-zero standard deviations across six datasets except for the Electricity dataset, whereas TDAlign demonstrates zero standard deviations on six datasets and exhibits non-zero but significantly smaller standard deviations on the ILI dataset compared to DLinear. 
Third, implicit learning is insufficient to capture TDT, as demonstrated by SegRNN, which maintains sequential order through position embeddings yet still benefits from TDAlign.
\begin{table}[ht]
    \centering
    \caption{Analysis of the impact of $\mathcal{L}_D$ and $\rho$ on the ETTh1 dataset, with the iTransformer as the baseline method. The best performances are highlighted in bold.}
    \label{tab:ablation_other}
    
\begin{tabular}{c|l|cc}
\toprule
$H$ & \multicolumn{1}{c|}{Setting} & MSE & MAE \\
\midrule
\multirow{5}{*}{96} & \circnum{1}~baseline only & 0.387±0.001 & 0.405±0.001 \\
  & \circnum{2}~+ $\mathcal{L}_D$  & 0.377±0.002 & 0.391±0.000 \\
  & \circnum{3}~+ $\rho$ & 0.387±0.001 & 0.399±0.001 \\
  & \circnum{4}~+ $\mathcal{L}_D$ + $\alpha$ & 0.378±0.001 & 0.391±0.001 \\
  & \circnum{5}~+ $\mathcal{L}_D$ + $\rho$ (TDAlign) & \textbf{0.375±0.001} & \textbf{0.389±0.000} \\
\midrule
\multirow{5}{*}{192} & \circnum{1}~baseline only  & 0.441±0.002 & 0.436±0.001 \\
  & \circnum{2}~+ $\mathcal{L}_D$  & 0.430±0.001 & 0.422±0.001 \\
  & \circnum{3}~+ $\rho$  & 0.438±0.001 & 0.428±0.001 \\
  & \circnum{4}~+ $\mathcal{L}_D$ + $\alpha$ & 0.430±0.002 & 0.423±0.001 \\
  & \circnum{5}~+ $\mathcal{L}_D$ + $\rho$ (TDAlign) & \textbf{0.426±0.001} & \textbf{0.420±0.001} \\
\midrule
\multirow{5}{*}{336} & \circnum{1}~baseline only & 0.491±0.004 & 0.461±0.003 \\
  & \circnum{2}~+ $\mathcal{L}_D$ & 0.473±0.001 & 0.445±0.001 \\
  & \circnum{3}~+ $\rho$ & 0.484±0.002 & 0.454±0.002 \\
  & \circnum{4}~+ $\mathcal{L}_D$ + $\alpha$ & 0.473±0.001 & 0.445±0.001 \\
  & \circnum{5}~+ $\mathcal{L}_D$ + $\rho$ (TDAlign) & \textbf{0.469±0.001} & \textbf{0.442±0.001} \\
\midrule
\multirow{5}{*}{720} & \circnum{1}~baseline only & 0.511±0.010 & 0.494±0.006 \\
  & \circnum{2}~+ $\mathcal{L}_D$ & 0.470±0.009 & 0.466±0.007 \\
  & \circnum{3}~+ $\rho$ & 0.482±0.010 & 0.476±0.006 \\
  & \circnum{4}~+ $\mathcal{L}_D$ + $\alpha$ & 0.471±0.009 & 0.466±0.007 \\
  & \circnum{5}~+ $\mathcal{L}_D$ + $\rho$ (TDAlign) & \textbf{0.460±0.001} & \textbf{0.460±0.001} \\
\bottomrule
\end{tabular}
\end{table}

Table~\ref{tab:count} presents statistics of improvements achieved by TDAlign with six baseline methods. For each baseline method and evaluation metric, we report the average percentage improvement over seven datasets and the count of cases showing improvement. 
The statistical results demonstrate that our proposed TDAlign framework consistently boosts the forecasting performance of six baseline methods in terms of both the TDP and prediction by a large margin. For TDP, the improvements range from 4.56\% to 12.17\% in $\rho$, 4.57\% to 15.78\% in $\text{MSE}_D$, and 3.79\% to 11.90\% in $\text{MAE}_D$. For the prediction, the improvements range from 1.47\% to 9.19\% in $\text{MSE}$ and 0.20\% to 9.72\% in $\text{MAE}$. Moreover, out of the 168 total cases, TDAlign demonstrates improvements in 166 cases for $\rho$, 145 for $\text{MSE}_D$, 164 for $\text{MAE}_D$, 148 for $\text{MSE}$, and 160 for $\text{MAE}$.

\xq{Table~\ref{tab:stationary} presents the average performance improvement conferred by TDAlign across six baseline methods for each dataset. As shown in the table, TDAlign demonstrates consistent performance improvements across all datasets, with particularly pronounced gains on non-stationary datasets. These datasets (e.g., ETTh1, ETTh2, ETTm2, and ILI) exhibit non-stationary characteristics, as confirmed by the Augmented Dickey-Fuller (ADF) test~\cite{elliott1992efficient}. These series, characterized by time-varying statistical properties, pose a significant challenge as their underlying dynamics are constantly evolving. This is precisely where TDAlign offers a crucial advantage. By explicitly constraining the model to capture local step-wise changes, it equips the baseline methods with the capability to better adapt to these non-stationarities, leading to substantial performance improvements. Specifically, averaged over all experimental prediction lengths and baseline methods, TDAlign achieves MSE and MAE reductions of 6.85\% and 5.49\% on ETTh2, 6.29\% and 5.60\% on ETTh1, 3.65\% and 3.92\% on ETTm2, and 6.39\% and 5.26\% on ILI, respectively.}

\subsection{Ablation Studies} \label{ablation}
We conduct ablation experiments on the ETTh1 dataset to assess the effectiveness of devised components in TDAlign and its computational efficiency. All experiments use iTransformer as the baseline method and are repeated five times with different random seeds.

\subsubsection{Effectiveness of $\mathcal{L}_D$ and $\rho$}

\begin{figure*}[ht]
    \centering

    \begin{subfigure}[t]{0.64\columnwidth}
        \centering
        \includegraphics[width=\linewidth]{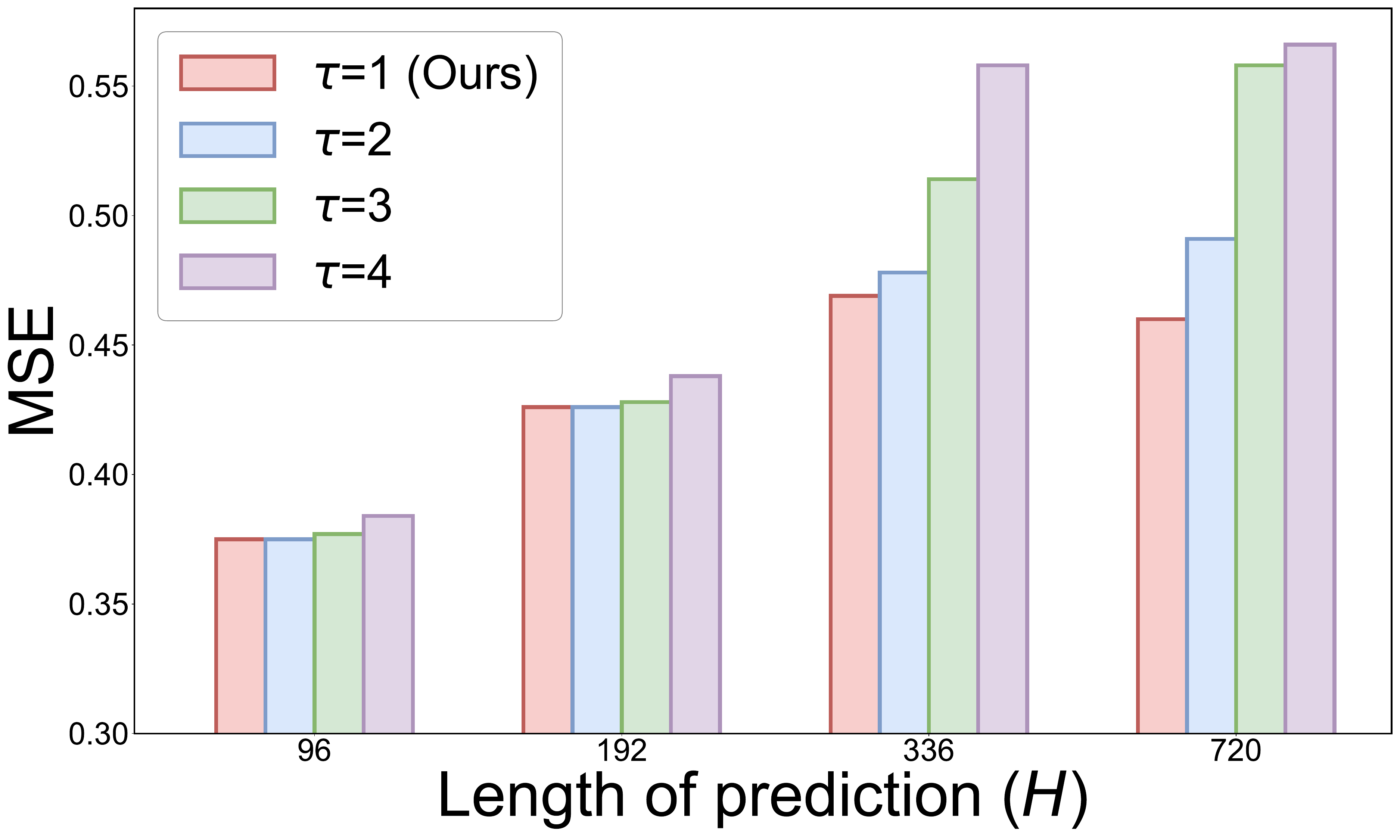}
        \caption{Using $\tau$-order differencing values.}
    \end{subfigure}
    \begin{subfigure}[t]{0.64\columnwidth}
        \centering
        \includegraphics[width=\linewidth]{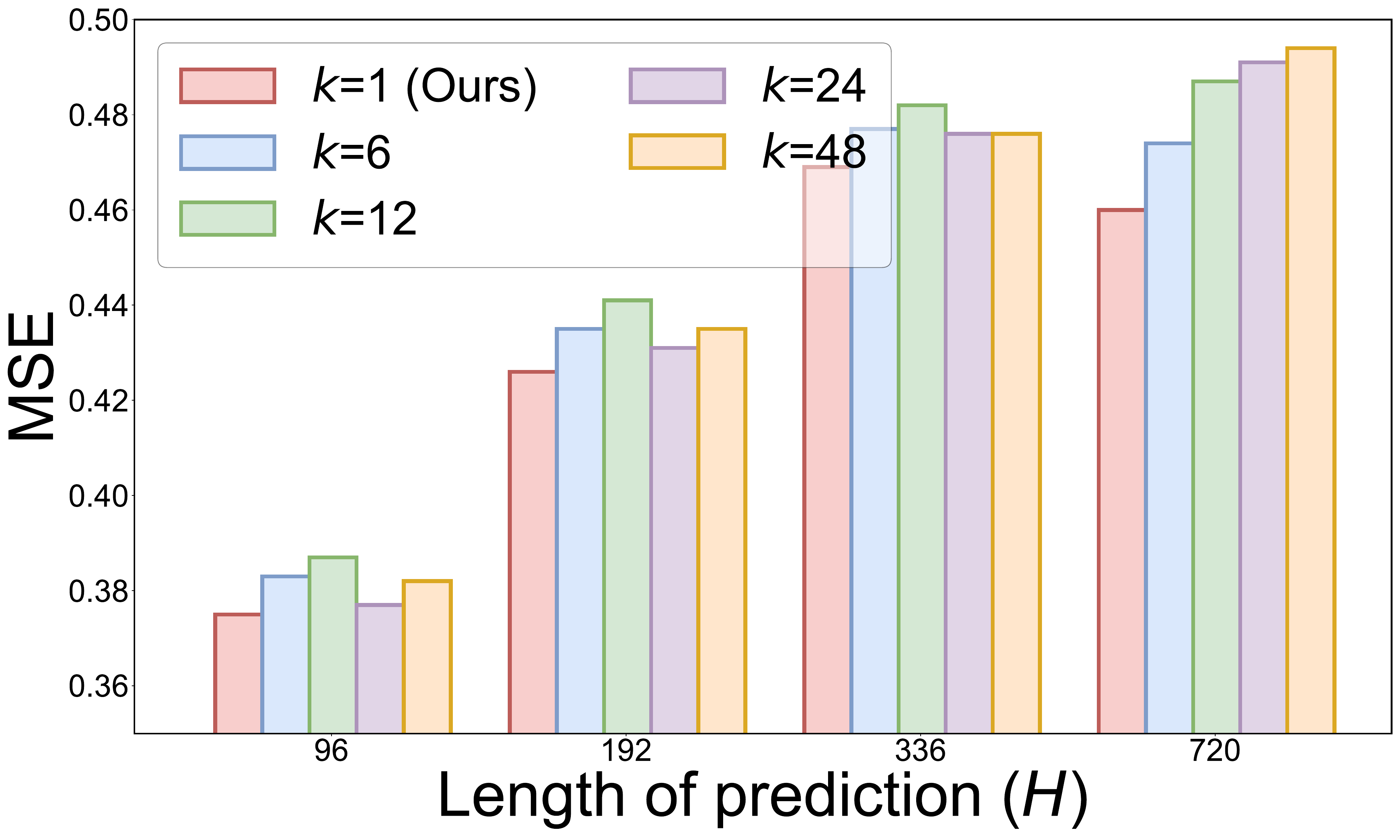}
        \caption{Using change values between time steps separated by the interval $k$.}
    \end{subfigure}
    \begin{subfigure}[t]{0.64\columnwidth}
        \centering
        \includegraphics[width=\linewidth]{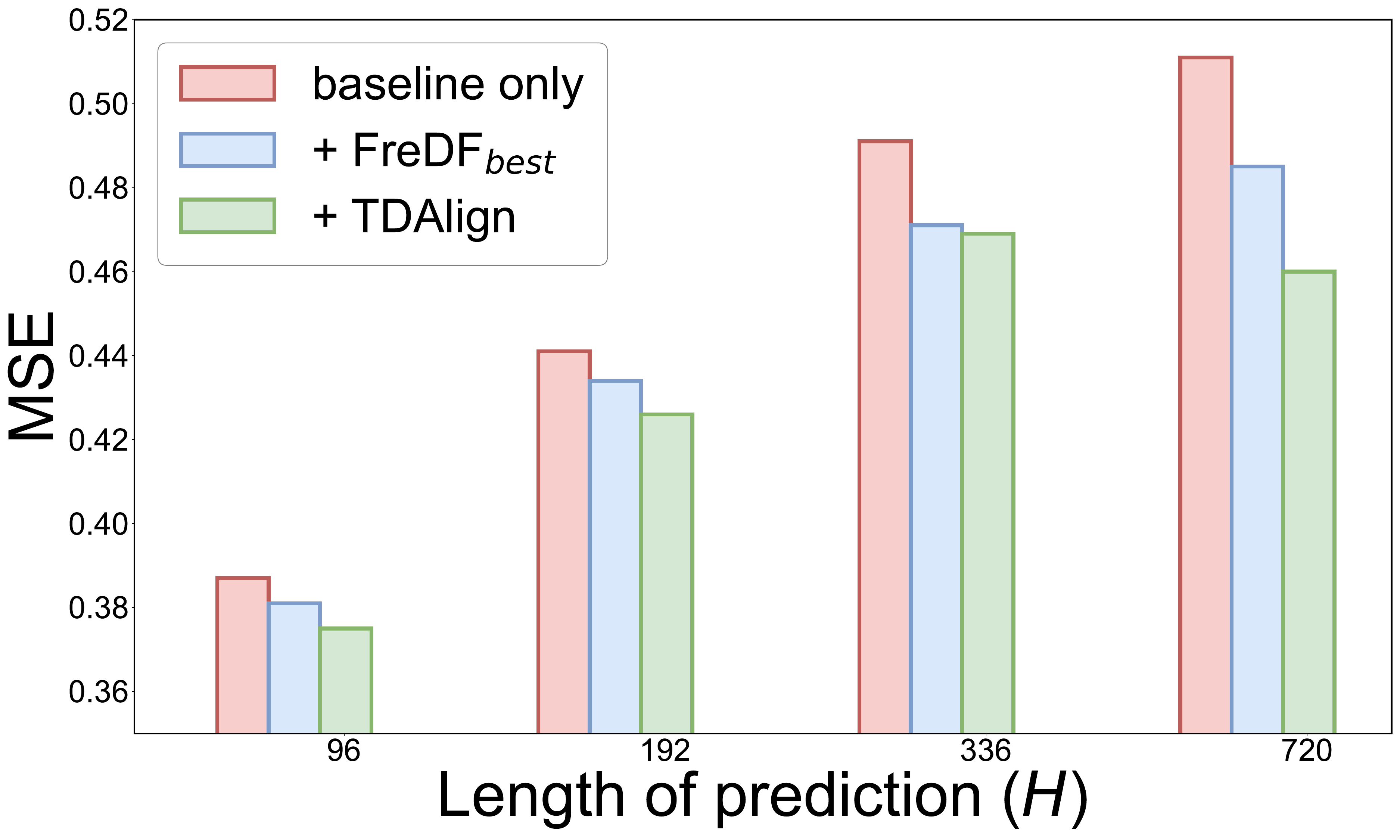}
        \caption{\xq{Using frequency domain features, i.e. FreDF~\cite{wang2025fredf}.}}
    \end{subfigure}

    \caption{Ablation of temporal dependency formulations on the ETTh1 dataset, with the iTransformer as the baseline method.}
    \label{fig:ablation_d}
\end{figure*}

Our TDAlign contains two key components, including a new loss function $\mathcal{L}_D$ for guiding the model to capture TDT, as well as an adaptive loss balancing strategy using a weight $\rho$ for dynamically adjusting the relative importance between the conventional optimization objective and the TDT learning objective. We analyze their impact through experiments with five different settings:
\begin{enumerate}[label=\circnum{\arabic*}, itemsep=4pt]
    \item baseline only, where the overall loss is $\mathcal{L}=\mathcal{L}_Y$.
    \item baseline with $\mathcal{L}_D$, where the overall loss is $\mathcal{L}=\mathcal{L}_Y+\mathcal{L}_D$.
    \item baseline with $\rho$, where the overall loss is $\mathcal{L}=\rho \mathcal{L}_Y$.
    \item baseline with $\mathcal{L}_D$ and $\alpha$ (i.e., replacing $\rho$ by a learning weight $\alpha$), where the overall loss is $\mathcal{L}=\alpha \mathcal{L}_Y+ (1-\alpha) \mathcal{L}_D$.
    \item baseline with TDAlign, where the overall loss is $\mathcal{L}=\rho \mathcal{L}_Y+ (1-\rho) \mathcal{L}_D$.
\end{enumerate}

The results are presented in Table~\ref{tab:ablation_other}, from which we observe that:
First, TDAlign significantly improves the forecasting performance of iTransformer across all four prediction lengths, as shown by the comparison between \circnum{1} and \circnum{5}.
Second, learning TDT is crucial for improving forecasting performance. Specifically, \circnum{2} enables the model to account for the change value between adjacent time steps when forecasting, while \circnum{3} enables the model to consider the change direction between adjacent time steps when forecasting. Both settings have superior forecasting performance compared to the baseline method, which overlooks internal correlations across multiple steps.
Third, our proposed new loss function has a more significant impact on TDAlign’s performance than our adaptive loss balancing strategy, where \circnum{2} outperforms \circnum{3} by a large margin.
Fourth, our proposed adaptive loss balancing strategy effectively controls the relative importance between the conventional optimization objective and the TDT learning objective. Specifically, by incorporating $\rho$, 
\circnum{5} outperforms \circnum{2}, which only incorporates $\mathcal{L}_D$, and \circnum{4}, where $\rho$ is replaced by a learning weight $\alpha$.

\subsubsection{Comparison of the Formulation of Temporal Dependencies}\label{td_repr} 
As shown in Figure~\ref{fig:ablation_d}, we evaluate our choice of using first-order differencing values (i.e., change values between adjacent time steps) to formulate temporal dependencies by comparing it against three alternative formulations:
\begin{itemize}
    \item Using $\tau$-order differencing values to capture more complex relations. We experiment with $\tau \in \{1, 2, 3, 4\}$.
    \item Using change values between time steps separated by an interval $k$ to capture periodic patterns. We experiment with $k \in \{1, 6, 12, 24, 48\}$.
    \item \xq{Using frequency domain features to capture global structures, as exemplified by the recent work FreDF \cite{wang2025fredf}. As noted in the original work, FreDF necessitates manual tuning of the relative weight to balance the loss terms, with optimal values varying across datasets and prediction lengths. To ensure a fair comparison, we experiment with an exhaustive grid search over $\{0.1, 0.2, \ldots, 1.0\}$ and report its best-performing results, denoted as $\text{FreDF}_{best}$.}
\end{itemize}

The results in Figure~\ref{fig:ablation_d}(a) show that using first-order differencing values (i.e., $\tau=1$) achieves the best performance, with performance degrading as $\tau$ increases.
Although higher-order differencing values represent more complex temporal dependencies, they may not provide substantial additional information compared to first-order differencing values. Instead, they tend to overshadow lower-order dependency information and over-constrain the model's learning, thereby compromising its learning and generalization ability. Moreover, since each $d^{(\tau)}_{t+i}$ involves more time steps, higher-order differencing values become more sensitive to noise, leading to degraded model performance. 
The results in Figure~\ref{fig:ablation_d}(b) indicate that using change values between adjacent time steps (i.e., $k=1$) yields the best performance. This phenomenon can be attributed to the fact that as $k$ increases, the correlation between $x_{t+i}$ and $x_{t+i+k}$ typically weakens. Learning unreliable temporal dependency information adversely affects model performance.
\xq{The results in Figure~\ref{fig:ablation_d}(b) further validate the effectiveness of our temporal dependency formulation. Although learning intricate correlations from the frequency domain yields substantial improvements, FreDF requires labor-intensive searching for the optimal relative weight. In contrast, our approach demonstrates superior performance while possessing greater flexibility in diverse forecasting scenarios.}
These results demonstrate that using first-order differencing values (i.e., change values between adjacent time steps) to formulate temporal dependencies is simple yet effective.

\begin{figure*}[ht]
    \centering
    \begin{subfigure}[t]{0.24\textwidth}
        \centering
        \includegraphics[width=\textwidth]{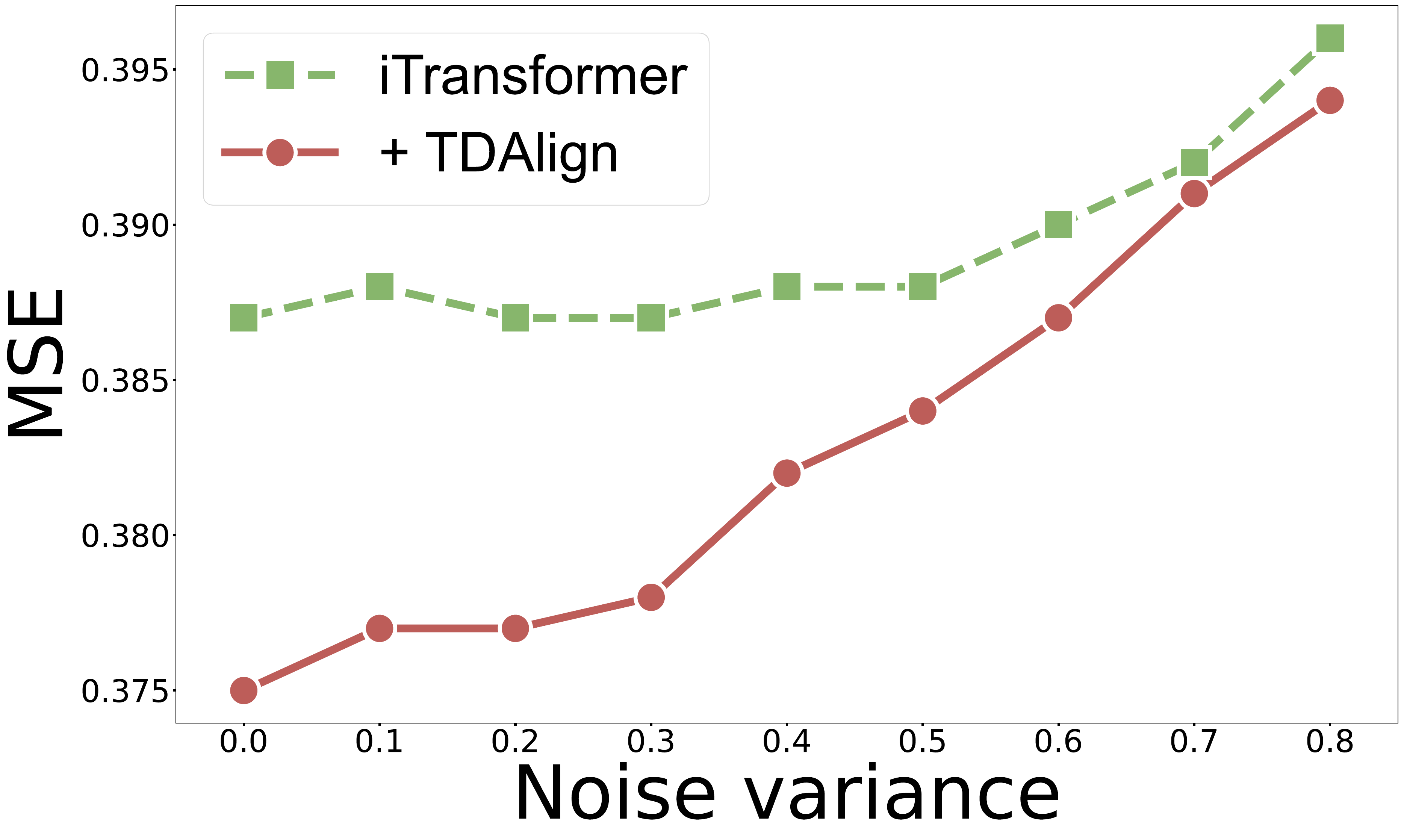}
        \caption{ETTh1, $L=96$, $H=96$}
    \end{subfigure}
    \begin{subfigure}[t]{0.24\textwidth}
        \centering
        \includegraphics[width=\textwidth]{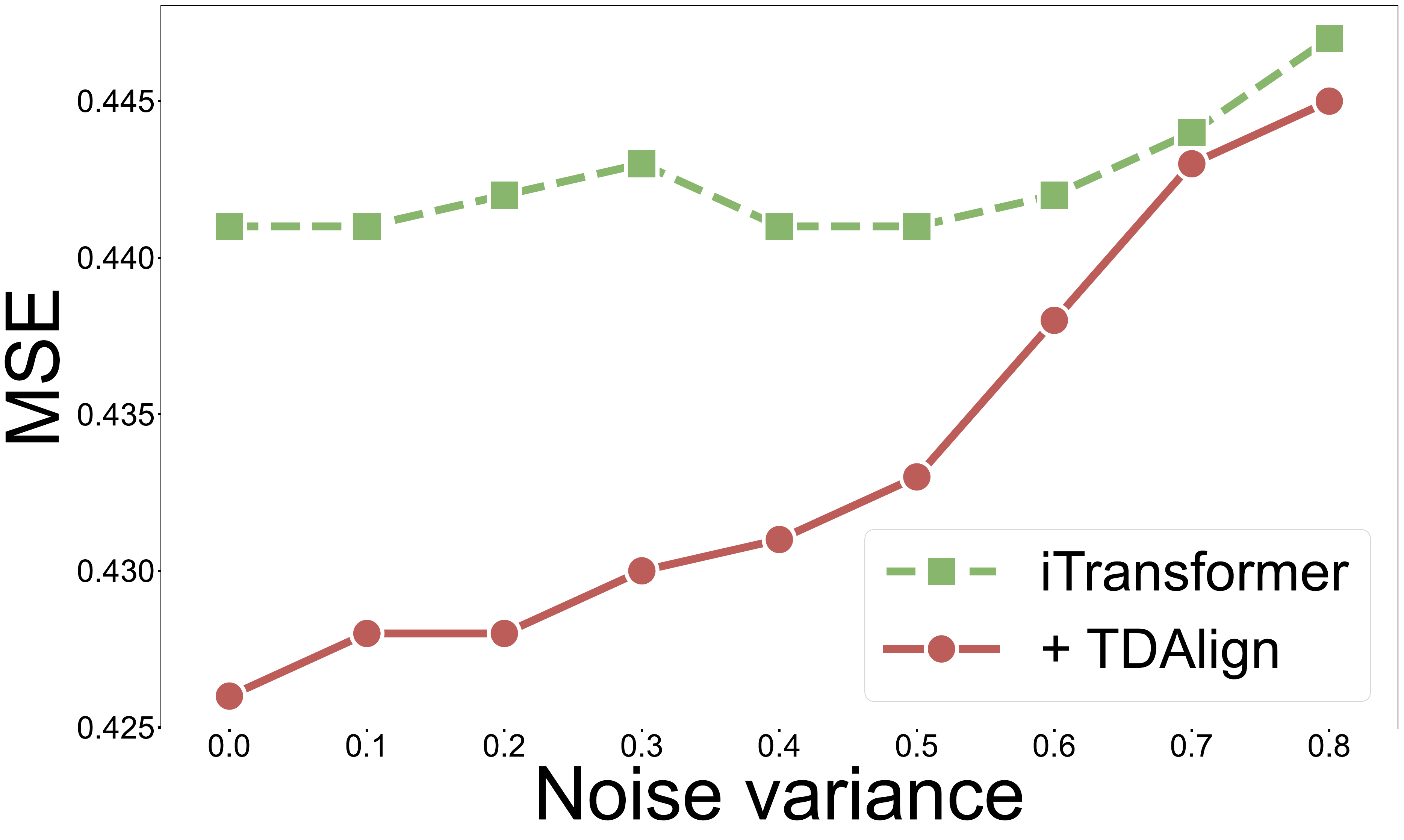}
        \caption{ETTh1, $L=96$, $H=192$}
    \end{subfigure}
    \begin{subfigure}[t]{0.24\textwidth}
        \centering
        \includegraphics[width=\textwidth]{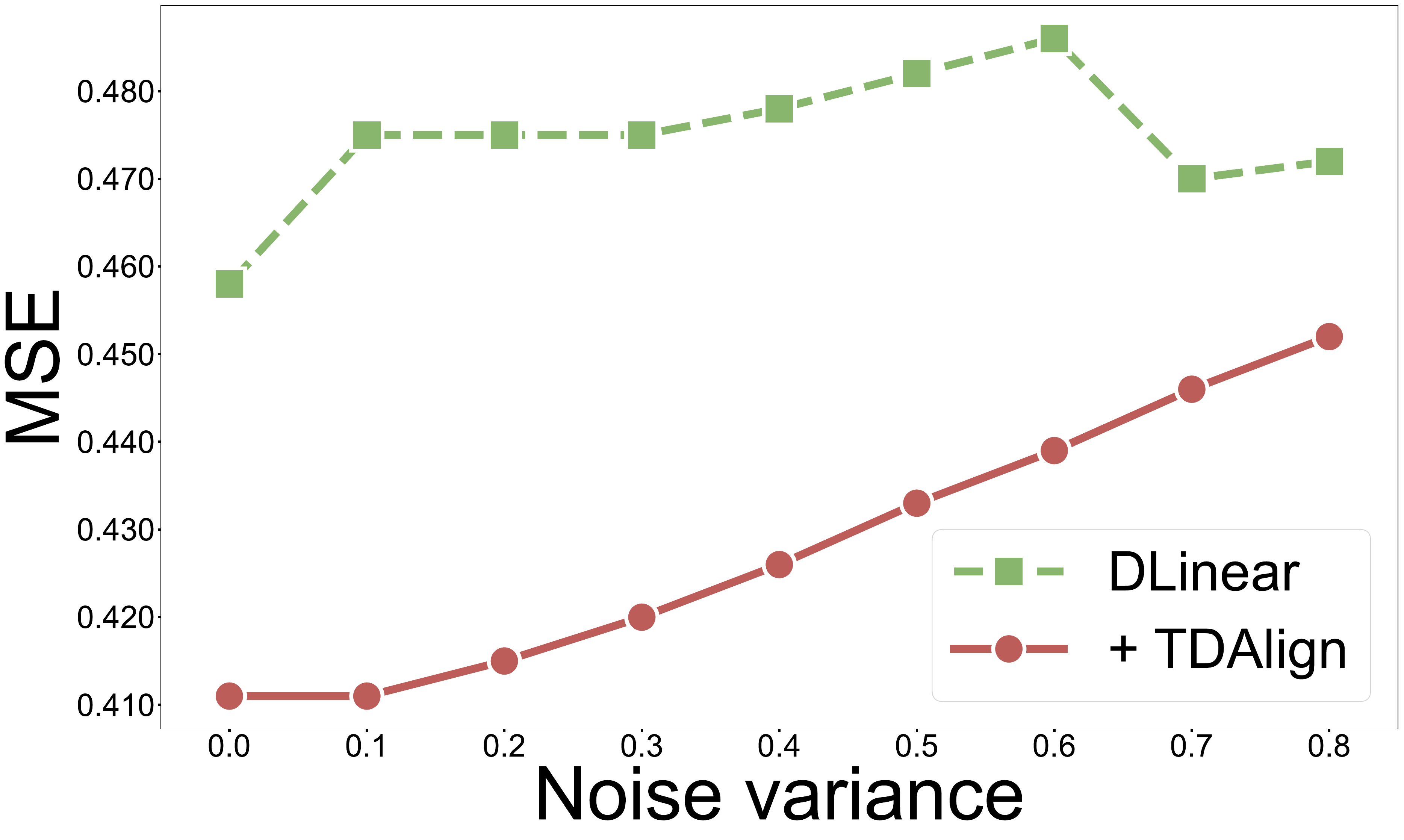}
        \caption{ETTh2, $L=336$, $H=336$}
    \end{subfigure}
    \begin{subfigure}[t]{0.24\textwidth}
        \centering
        \includegraphics[width=\textwidth]{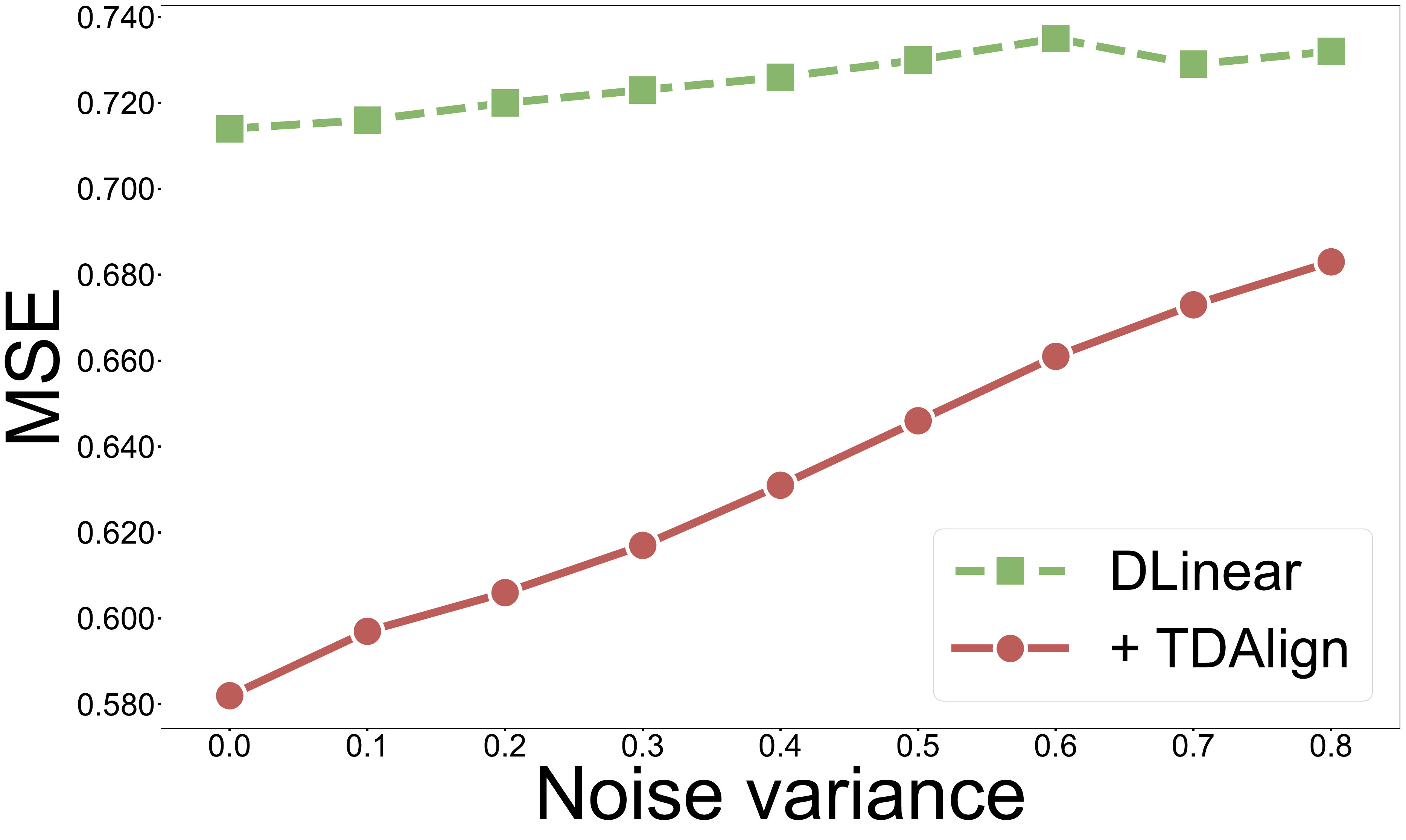}
        \caption{ETTh2, $L=336$, $H=720$}
    \end{subfigure}
    \caption{\xq{Robustness of TDAlign under Gaussian Noise with varying variances. (a)-(b) use iTransformer as the baseline method, (c)-(d) use DLinear as the baseline method.}}
    \label{fig:noise}
\end{figure*}

\begin{figure*}[ht]
    \begin{subfigure}[t]{0.24\textwidth}
        \centering
        \includegraphics[height=2.6cm]{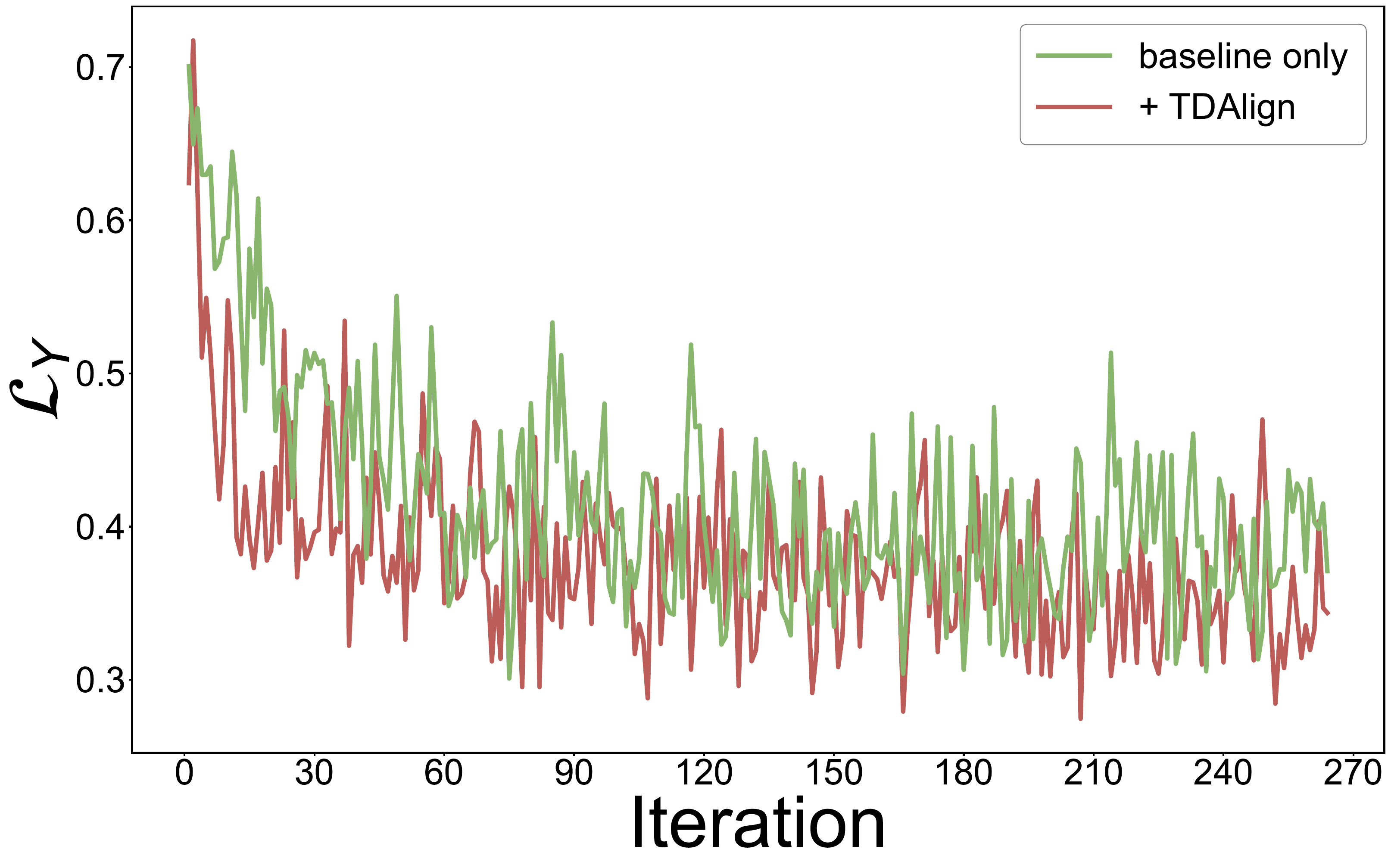}
        \caption{Evolution of $\mathcal{L}_Y$.}
    \end{subfigure}
    \begin{subfigure}[t]{0.24\textwidth}
        \centering
        \includegraphics[height=2.6cm]{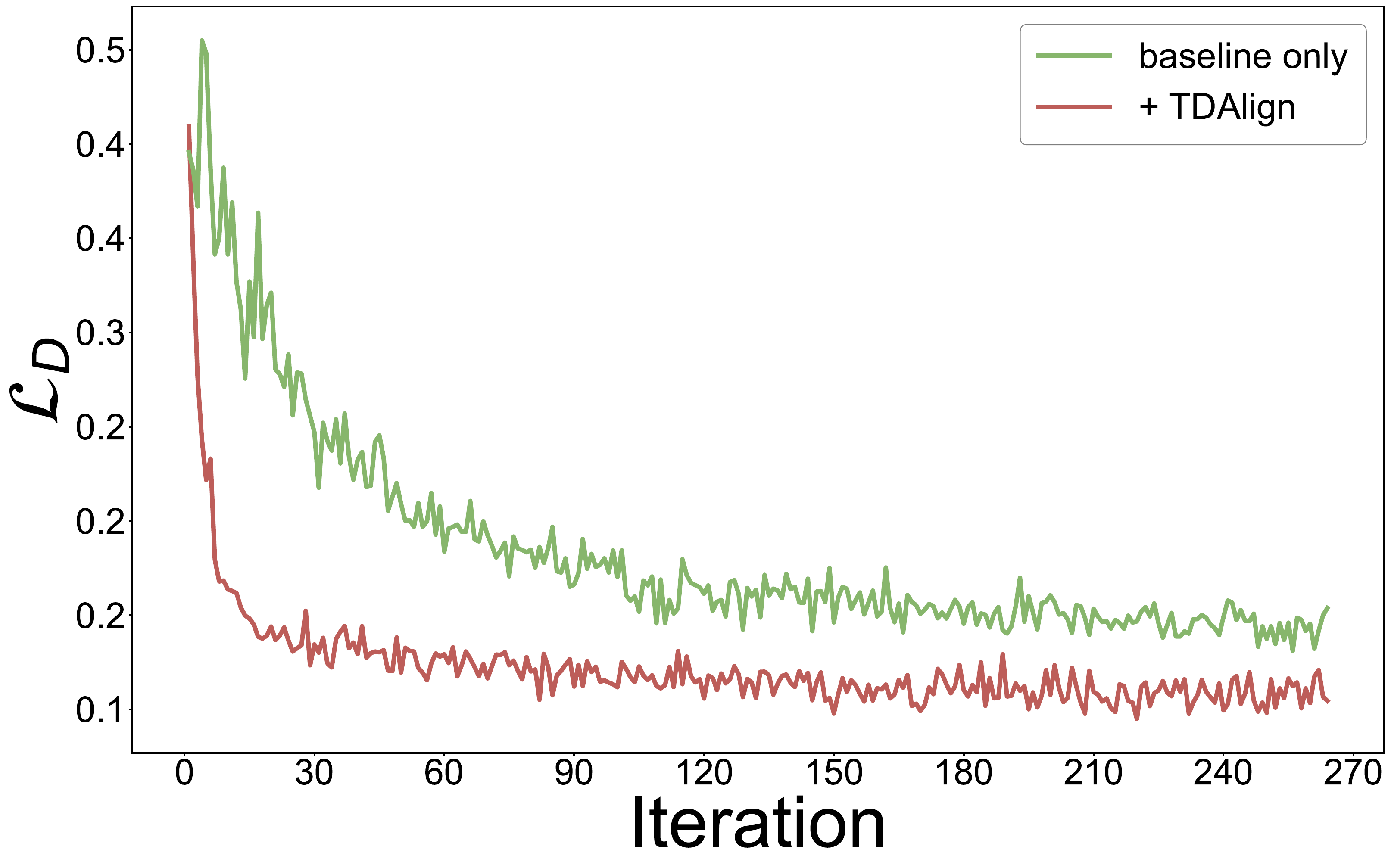}
        \caption{Evolution of $\mathcal{L}_D$.}
    \end{subfigure}
    \begin{subfigure}[t]{0.24\textwidth}
        \centering
        \includegraphics[height=2.6cm]{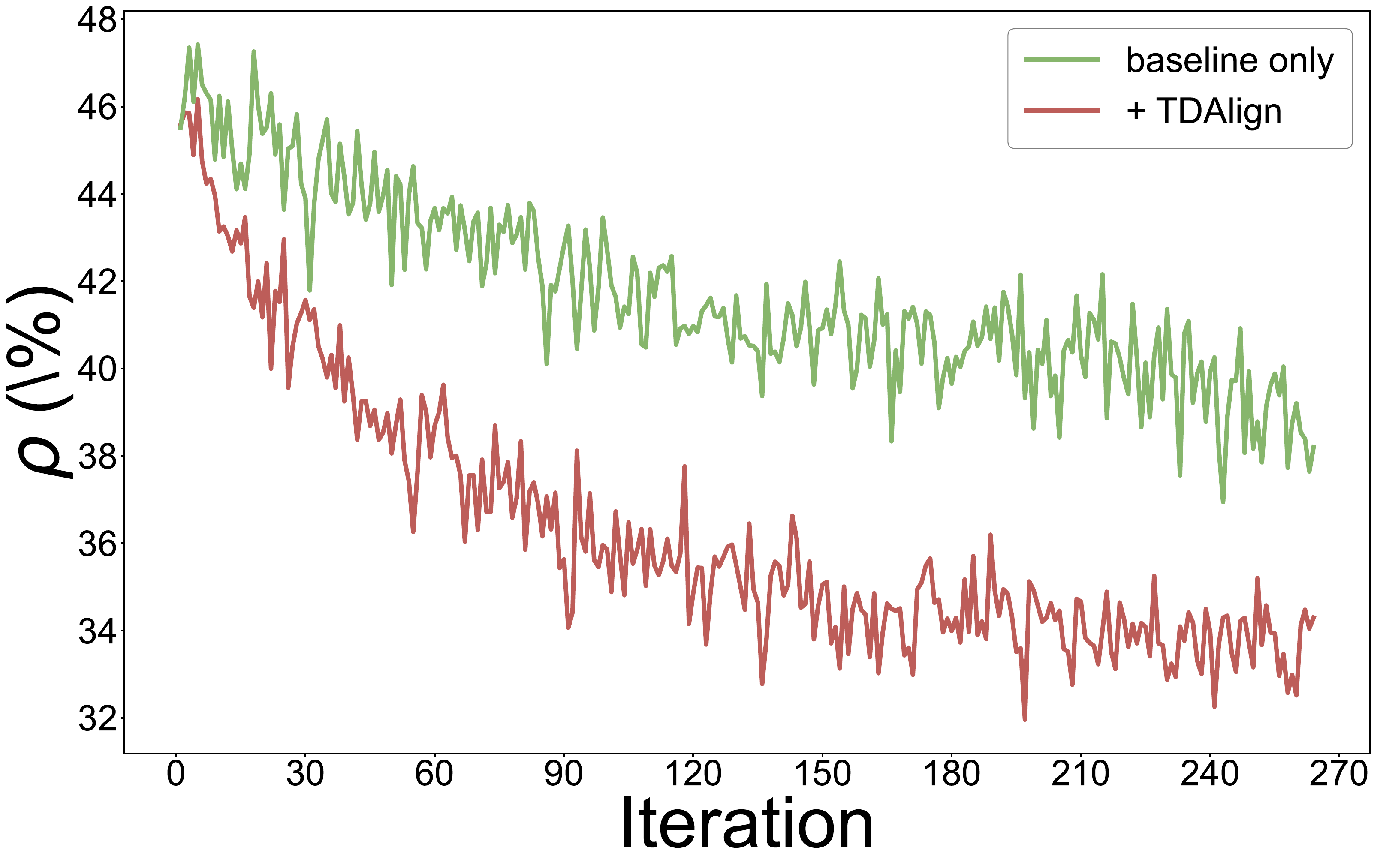}
        \caption{Evolution of $\rho$.}
    \end{subfigure}
    \begin{subfigure}[t]{0.224\textwidth}
        \centering
        \includegraphics[height=2.63cm]{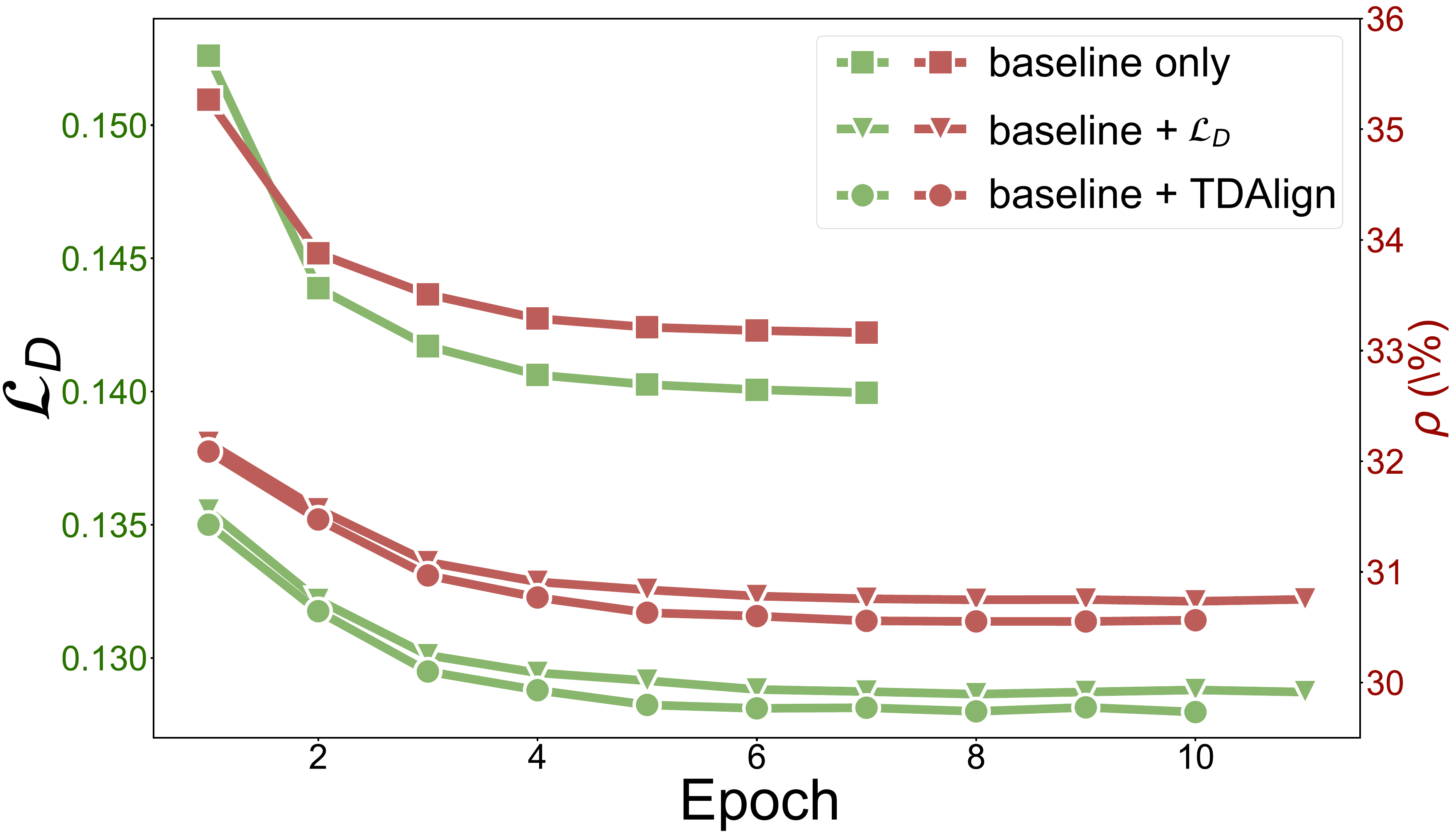}
        \caption{Validation $\text{MSE}_D$ and $\rho$.}
    \end{subfigure}
     \caption{\xq{Learning curves of TDAlign on the ETTh1 dataset, with iTransformer as the baseline method. }}
    \label{fig:training}
\end{figure*}

\begin{table}[ht]
    \centering
    \caption{\xq{Computational cost of TDAlign on the ETTh1 dataset, with iTransformer as the baseline method.}}
    \label{tab:cost}
\begin{tabular}{c|c|ccc}
\toprule
\multirow{2}{*}{$H$}
                   & \multirow{2}{*}{Method} & Learnable params & MACs & Training time \\
                   &  & $(K)$ & $(MB)$ & $(seconds/epoch)$ \\
\midrule
\multirow{2}{*}{96}  & baseline                &   841.568                                 &    295.928          &    4.460                                           \\
                     & +TDAlign                    &   \textbf{+ 0}                               &    \textbf{+ 0}                       &            \textbf{+ 0.189}                                 \\
\midrule   
\multirow{2}{*}{192} & baseline                &    866.240                                 &      304.579         &    4.647                                   \\
                     & +TDAlign                    &   \textbf{+ 0}                                 &     \textbf{+ 0}            &  \textbf{+ 0.053}   \\
\midrule 
\multirow{2}{*}{336} & baseline                &    3379.024                                 &     1188.758               &     4.679                                       \\
                     & +TDAlign                    &     \textbf{+ 0}                                   &     \textbf{+ 0}               &     \textbf{+ 0.197}                               \\
\midrule 
\multirow{2}{*}{720} & baseline                &   3576.016                                  &     1257.964           &    4.524                                           \\
                     & +TDAlign                    &   \textbf{+ 0}                                    &   \textbf{+ 0}               &                           \textbf{+ 0.002}         \\
\bottomrule
\end{tabular}
\end{table}

\begin{figure*}[htb]
    \centering
    \begin{subfigure}{0.31\textwidth}
        \includegraphics[width=\textwidth]{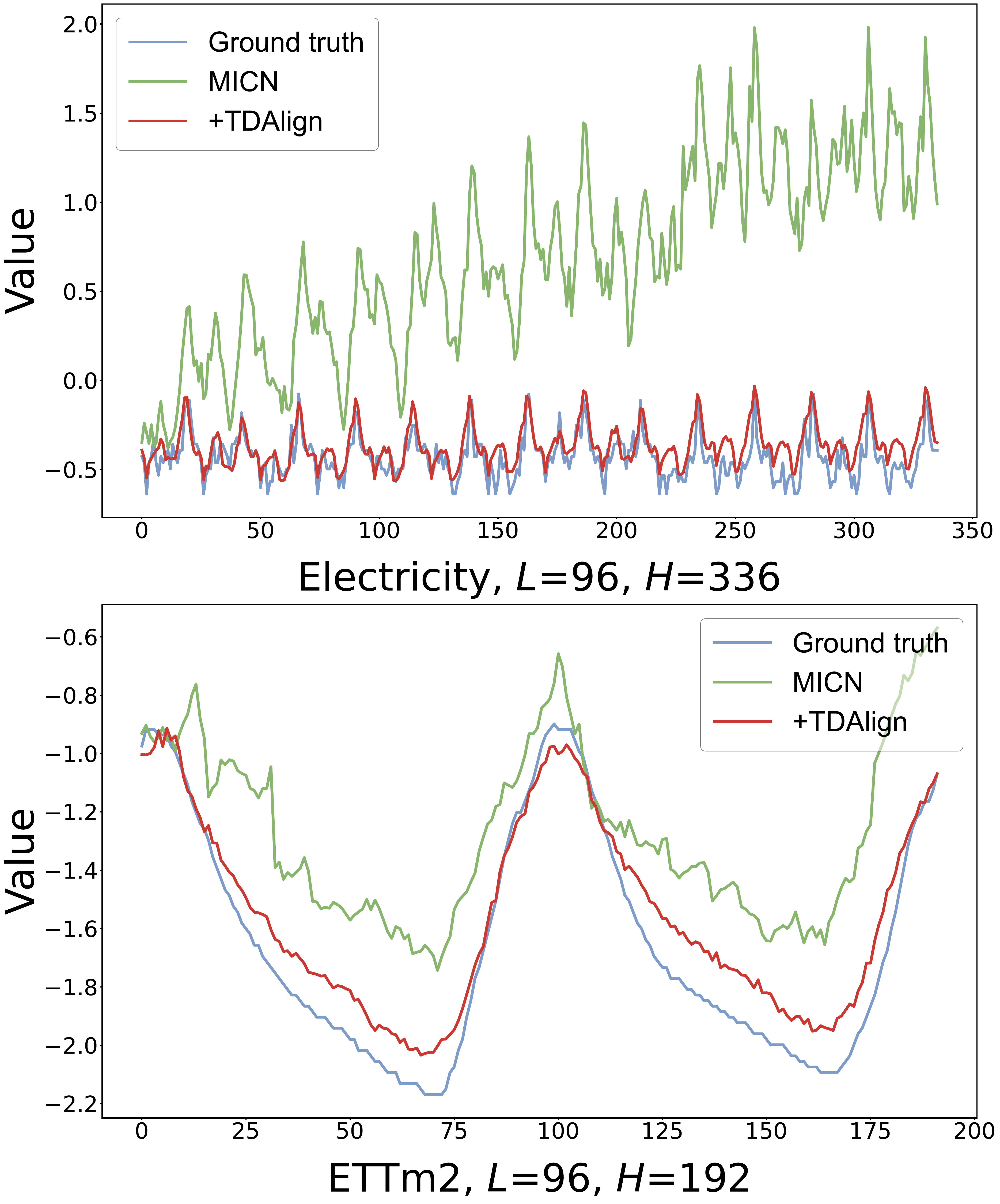}
        \caption{MICN}
    \end{subfigure}
    \begin{subfigure}{0.31\textwidth}
        \includegraphics[width=\textwidth]{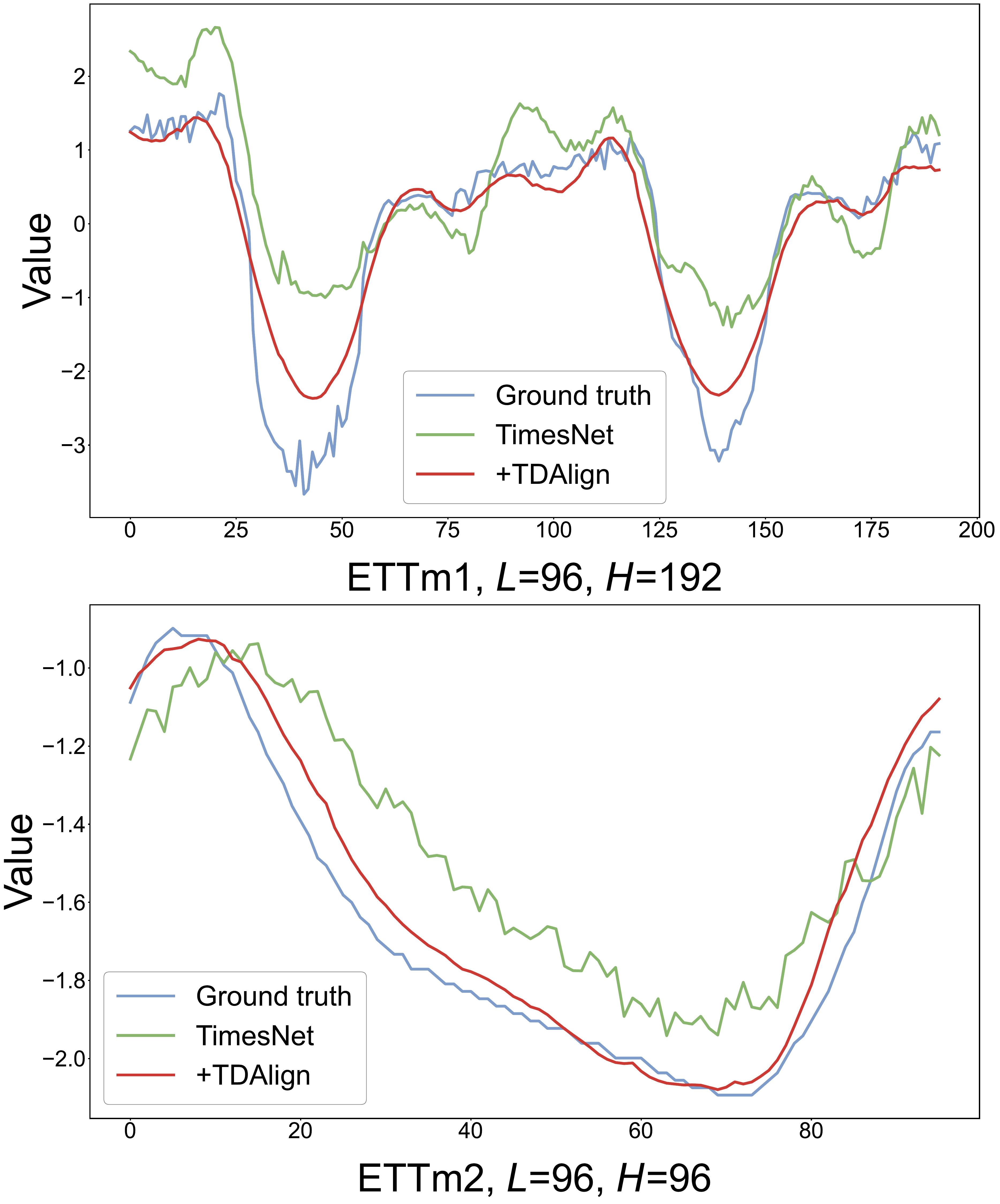}
        \caption{TimesNet}
    \end{subfigure}
    \begin{subfigure}{0.31\textwidth}
        \includegraphics[width=\textwidth]{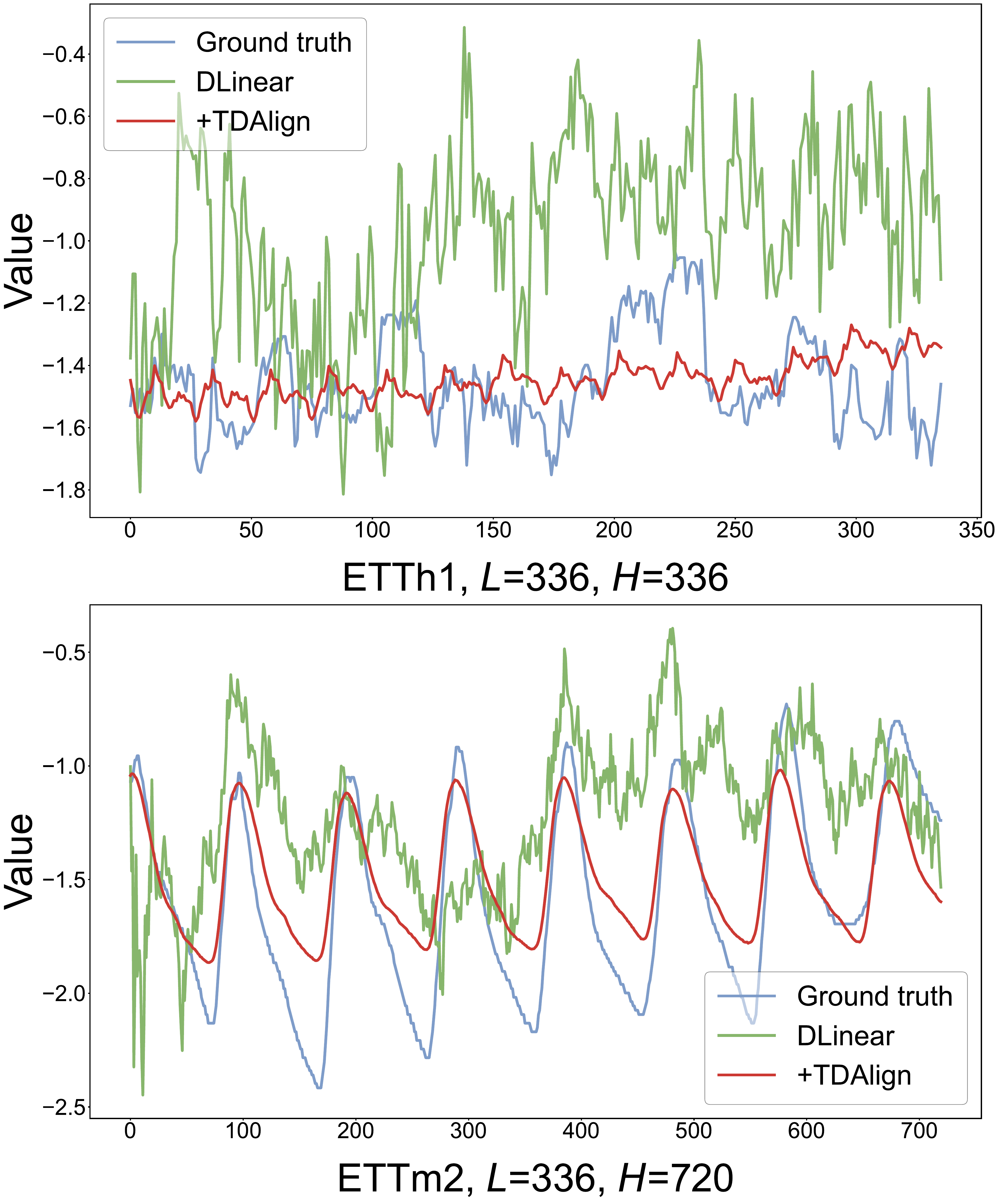}
        \caption{DLinear}
    \end{subfigure}
    
    \begin{subfigure}{0.31\textwidth}
        \includegraphics[width=\textwidth]{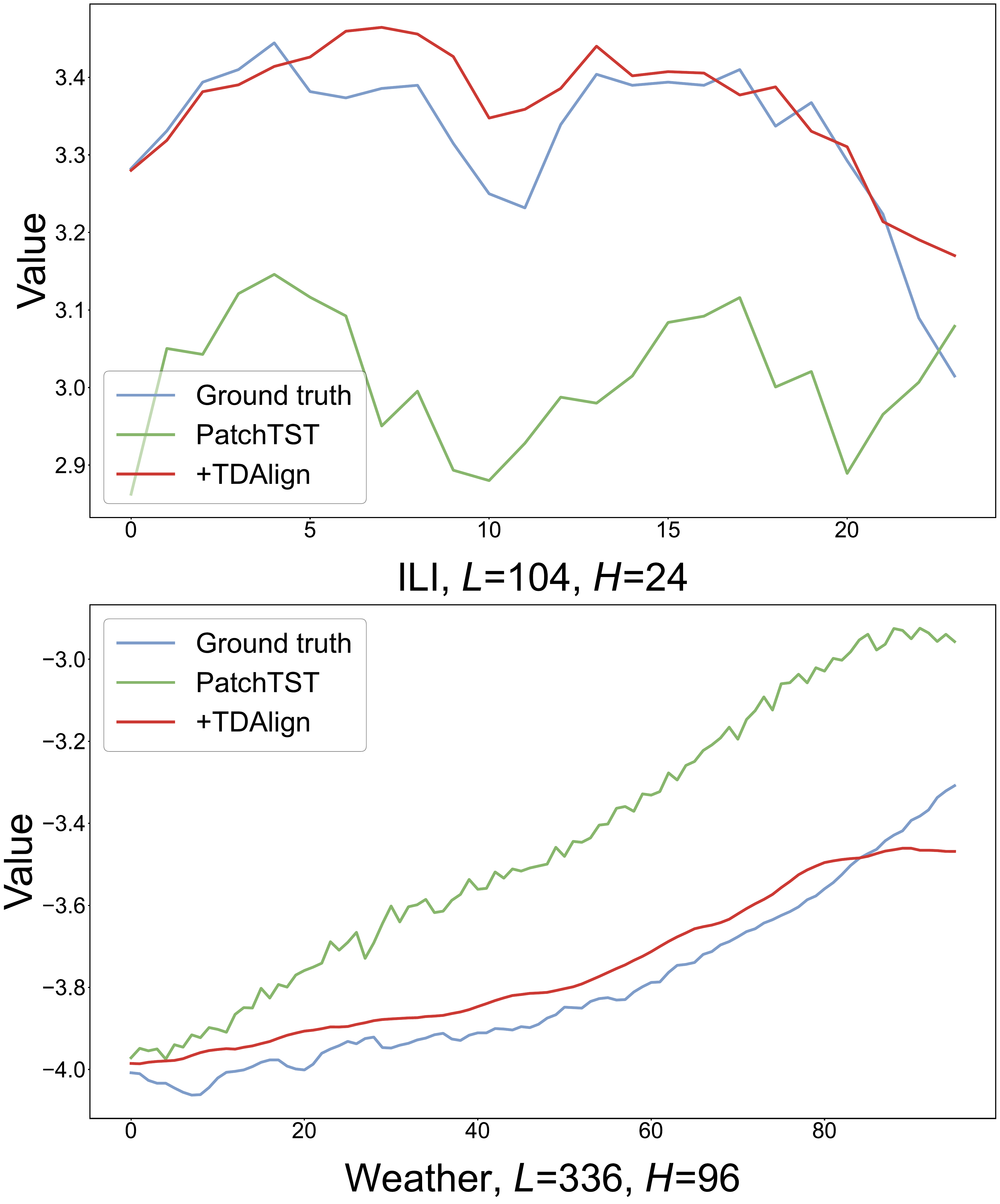}
        \caption{PatchTST}
    \end{subfigure}
    \begin{subfigure}{0.31\textwidth}
        \includegraphics[width=\textwidth]{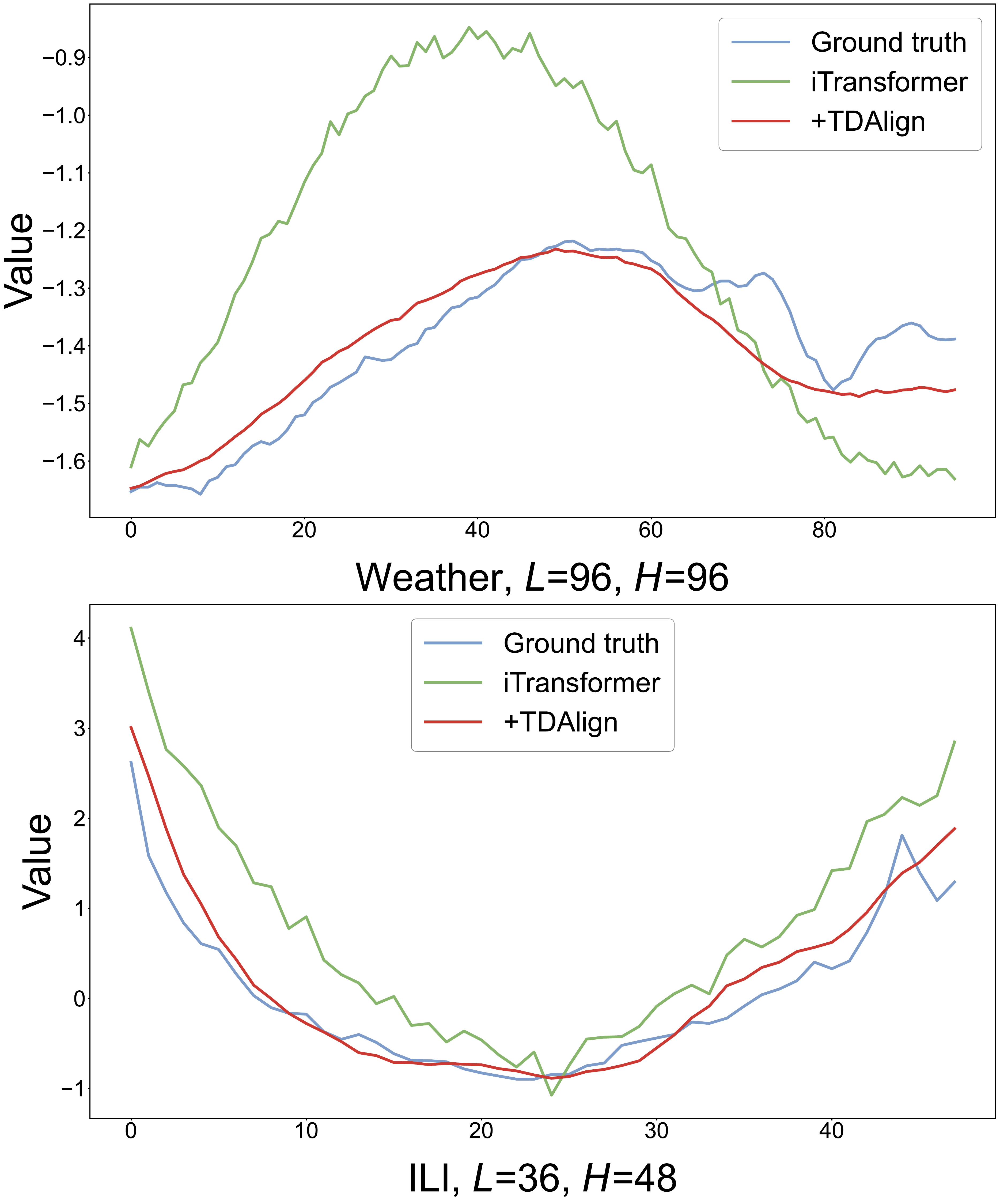}
        \caption{iTransformer}
    \end{subfigure}
    \begin{subfigure}{0.31\textwidth}
        \includegraphics[width=\textwidth]{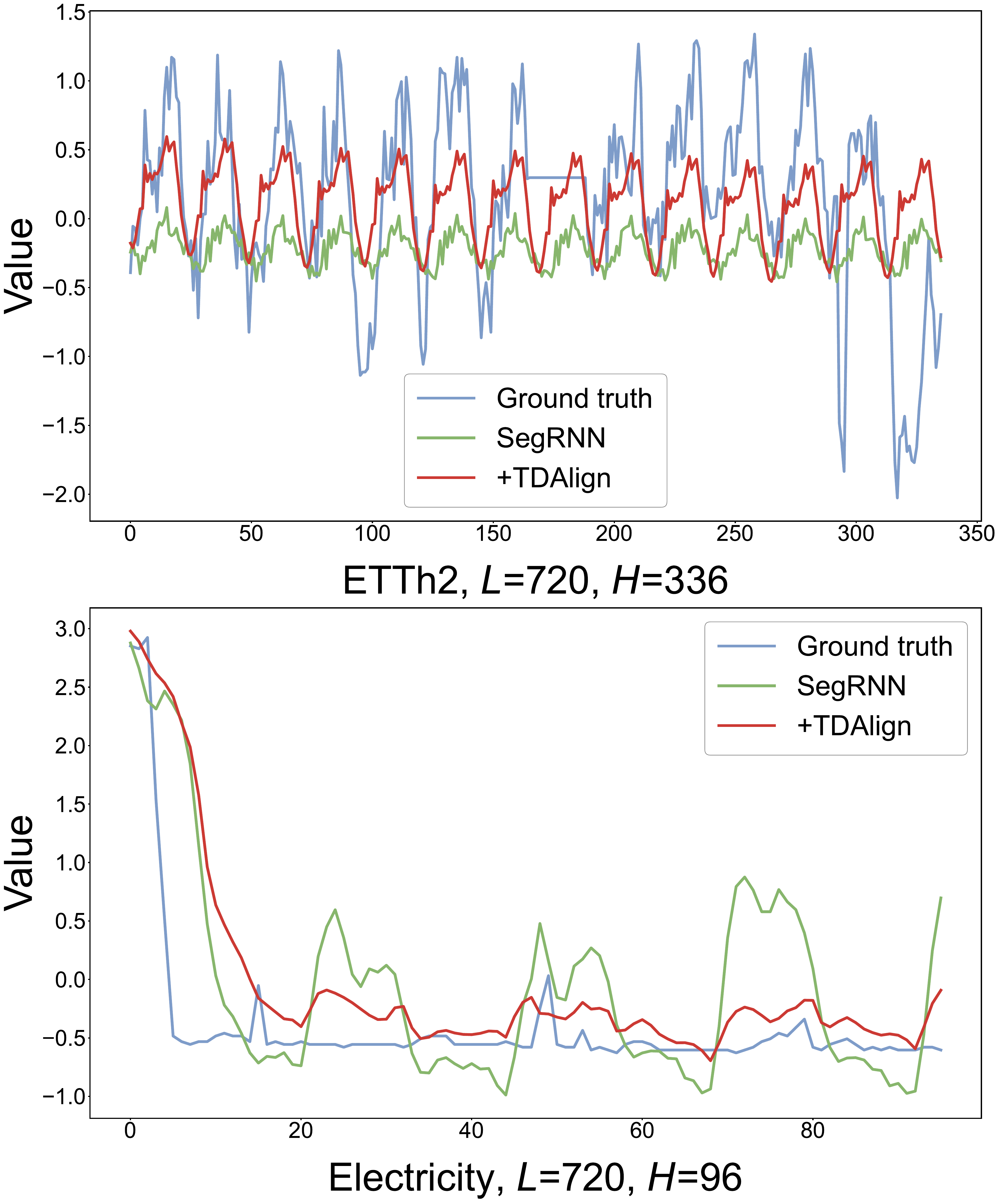}
        \caption{SegRNN}
    \end{subfigure}
    \caption{Comparison of long-term time series forecasting results between our TDAlign and six baseline methods.}
    \label{fig:cases}
\end{figure*}


\xq{
\subsubsection{Computational Cost}\label{cost}
The computational cost of TDAlign is presented in Table~\ref{tab:cost}.
As shown in the table, TDAlign maintains the same number of learnable parameters ($K$) and multiply-accumulate operations (MACs) as the baseline method across various time series lengths. The only overhead introduced by TDAlign is a slight increase in training time per epoch. These statistics demonstrate that TDAlign achieves the aforementioned substantial performance improvements while incurring only a negligible additional computational overhead, making it a highly efficient module that can be integrated with baseline methods for effective TDT modeling.
}

\xq{
\subsection{Additional Experiments}
\subsubsection{Effectiveness on LLM-based Methods}

\begingroup
\begin{table*}[!htb]
    \centering
    \caption{\xq{Effectiveness of our TDAlign over different LLM-based baselines. Results are averaged over 3 runs, with better performance highlighted in bold.}}
    \label{tab:llm}
    {
\begin{tabular}{cc|cccc|cccc}
\toprule
\multicolumn{2}{c|}{Architecture}&\multicolumn{8}{c}{LLM-based} \\  
\midrule
\multicolumn{2}{c|}{Method}&\multicolumn{2}{c}{TimesFM}	&\multicolumn{2}{c}{+TDAlign}	&\multicolumn{2}{c}{Time-LLM}	&\multicolumn{2}{c}{+TDAlign}	\\
\multicolumn{2}{c|}{Metric}&MSE	&MAE	&MSE	&\multicolumn{1}{c}{MAE}	&MSE	&MAE	&MSE	&MAE	\\
\midrule
\multirow{4}{*}{\myrotcell{ETTh1}} 
& 96  
&0.536±0.021&0.496±0.004&\textbf{0.511±0.013}&\textbf{0.466±0.005}
&0.407±0.001&0.426±0.003&\textbf{0.401±0.006}&\textbf{0.414±0.005} \\
& 192 
&0.559±0.021&0.508±0.002&\textbf{0.540±0.011}&\textbf{0.484±0.004}
&0.447±0.001&0.454±0.001&\textbf{0.423±0.011}&\textbf{0.429±0.006}\\
& 336 
&0.563±0.020&0.512±0.006&\textbf{0.544±0.011}&\textbf{0.491±0.004}
&0.462±0.010&0.467±0.007&\textbf{0.436±0.005}&\textbf{0.444±0.003}\\
& 720 
&0.539±0.028&0.516±0.014&\textbf{0.529±0.014}&\textbf{0.502±0.005}
&0.511±0.024&0.506±0.015&\textbf{0.452±0.010}&\textbf{0.478±0.008}\\
\midrule
\multirow{4}{*}{\myrotcell{ETTh2}} 
& 96  
&0.358±0.031  &0.397±0.021  & \textbf{0.323±0.008} & \textbf{0.360±0.003}
&0.304±0.007 &0.363±0.004 &\textbf{0.294±0.014}  & \textbf{0.355±0.010}\\
& 192 
&0.421±0.030  &0.433±0.020  &\textbf{0.380±0.016} & \textbf{0.401±0.001}
&0.363±0.005 &0.393±0.003 &\textbf{0.342±0.003} & \textbf{0.384±0.004}\\
& 336 
&0.429±0.017  &0.448±0.014  &\textbf{0.404±0.007} &\textbf{0.423±0.003}
&0.381±0.006 &0.412±0.003 &\textbf{0.362±0.008} &\textbf{0.399±0.004} \\
& 720 
& 0.476±0.035 & 0.475±0.023 & \textbf{0.443±0.010} & \textbf{0.462±0.009}
&0.414±0.005 &0.447±0.005 &\textbf{0.389±0.005} & \textbf{0.430±0.006} \\
\midrule
\multicolumn{2}{c|}{Avg error} & 0.485 & 0.473 & \textbf{0.459} &	\textbf{0.449} & 0.411 & 0.434 & \textbf{0.387} &	\textbf{0.417} \\
\bottomrule
\end{tabular}
    }
\end{table*}
\endgroup
To evaluate the generalizability of TDAlign to Large Language Model (LLM) based methods, we apply it to two state-of-the-art forecasters: TimesFM (the 500M version)~\cite{das2024decoder} and Time-LLM (the base version)~\cite{jin2024time}. We assess their performance under a full-shot fine-tuning setting for LTSF, repeating each experiment three times to ensure reliable results.

As presented in Table~\ref{tab:llm}, TDAlign yields consistent and substantial performance gains for both models across multiple datasets and prediction horizons. This underscores the broad applicability of TDAlign as a plugin for diverse forecasting architectures, including large-scale pre-trained models.

\subsubsection{Robustness to Noise}\label{sec:noise}

To examine the robustness of TDAlign to noise, we inject Gaussian noise with varying variances into the training set as in \cite{cheng2024robusttsf}. We assess the performance using two baseline methods: iTransformer on the ETTh1 dataset with $H \in \{96, 192\}$, and DLinear on the ETTh2 dataset with $H \in \{336, 720\}$.

Figure~\ref{fig:noise} demonstrates that TDAlign consistently outperforms baseline methods across all experimental noise levels. 
Specifically, by learning local temporal dependencies and adaptively adjusting the relative weight between loss terms, TDAlign delivers a robust performance advantage, even under high-noise conditions. Simple models such as DLinear benefit more from the explicit modeling of TDT than more complex architectures like iTransformer.
Additionally, we observe that TDAlign's performance advantage over baseline methods becomes less pronounced as noise levels increase, reflecting the inherent challenges of learning reliable temporal patterns in noisy data. However, given that our tested noise levels already represent substantial perturbations rarely encountered in practical time series applications, TDAlign maintains its practical robustness and applicability in real-world scenarios.
}

\subsection{Visualization}\label{visualization}
\xq{
\subsubsection{Training Process Analysis}\label{training_process}
To provide an intuitive understanding of the training process of TDAlign, Figure~\ref{fig:training} illustrates its learning curves on the ETTh1 dataset, using iTransformer as the baseline method with input and prediction lengths of 96. Subfigures (a-c) depict the evolution of the conventional forecasting objective $\mathcal{L}_Y$, the TDT learning objective $\mathcal{L}_D$, and the adaptive weight $\rho$ within the first training epoch. Particularly, $\rho$ not only acts as an adaptive weight to balance the two objectives but also serves as a metric representing the sign inconsistency ratio. Subfigure (d) tracks the validation metric MSE$_D$ and the value of $\rho$ across all epochs for three distinct experimental settings:}
\begin{enumerate}[label=\circnum{\arabic*}, itemsep=4pt]
    \item baseline only, where the overall loss is $\mathcal{L}=\mathcal{L}_Y$.
    \item baseline with $\mathcal{L}_D$, where the overall loss is $\mathcal{L}=\mathcal{L}_Y+\mathcal{L}_D$.
    \item baseline with TDAlign, where the overall loss is $\mathcal{L}=\rho \mathcal{L}_Y + (1-\rho) \mathcal{L}_D$.
\end{enumerate}

\xq{Figure~\ref{fig:training} (a–c) reveals two key insights into the benefits of TDAlign.
First, TDAlign significantly enhances TDT learning, as evidenced by its ability to achieve substantially lower $\mathcal{L}_D$ and $\rho$ values compared to the baseline method, while maintaining comparable primary forecasting performance ($\mathcal{L}_Y$).
Second, the proposed dynamic weight $\rho$ effectively balances and steers the training process toward more efficient learning, as reflected in the faster convergence of both $\mathcal{L}_D$ and $\rho$ for TDAlign.}

From validation curves in Figure~\ref{fig:training} (d), we observe the following:
First, the training process of TDAlign is stable, where loss curves of \circnum{3} decrease smoothly.
Second, explicitly modeling TDT significantly improves the prediction accuracy of both change directions and change values between adjacent time steps, as shown by the lower $\text{MSE}_D$ and $\rho$ of \circnum{2} compared to \circnum{1}.
Third, the devised adaptive loss balancing strategy further enhances TDAlign's performance, as indicated by the further reduction in $\text{MSE}_D$ and $\rho$ of \circnum{3} compared to \circnum{2}.

\subsubsection{Case Study}\label{sec:cases}
To illustrate the generalizability and effectiveness of TDAlign across datasets with diverse characteristics, Figure~\ref{fig:cases} visualizes forecasting results from each baseline method on two distinct datasets, covering all baseline methods and collectively spanning all datasets used in our main experiments.
Overall, baseline methods produce inaccurate predictions with significant deviations from the ground truth. A closer examination reveals that these methods struggle to accurately capture both the change values and the change directions. Their predictions exhibit unrealistic fluctuations and jagged patterns, indicating their failure to learn the true temporal dynamics. This behavior is inconsistent with the inherent local smoothness of real-world time series, where consecutive observations transition gradually rather than abruptly.
In contrast, TDAlign consistently generates more accurate predictions that closely align with the ground truth both locally and globally.
Moreover, it mitigates the unrealistic fluctuations observed in baseline methods, resulting in smoother and more realistic predictions that better reflect the inherent local coherence of real-world time series.
These improvements are primarily attributed to the proposed TDT learning objective, which explicitly captures variation patterns inherent in time series. Furthermore, by dynamically adjusting the importance of the TDT learning objective relative to the conventional forecasting objective based on the model's learning state, TDAlign achieves a balanced optimization that enhances both local temporal modeling and global prediction accuracy. Consequently, TDAlign demonstrates significant improvements over baseline methods across diverse forecasting scenarios.
\section{Conclusion}
In this study, we investigated existing LTSF methods, revealing that their forecasting performance is limited by the inadequate modeling of TDT. To alleviate this problem, we proposed the TDAlign framework, which can plug into existing LTSF methods without introducing additional learnable parameters. 
In addition, we conducted extensive experiments on seven widely used benchmark datasets, and the results demonstrate that TDAlign consistently enhances the forecasting performance of six state-of-the-art LTSF methods by a significant margin, showcasing its flexibility and effectiveness. We hope that TDAlign can serve as a fundamental component for time series forecasting and offer a new perspective for improving forecasting performance.

\xq{Despite its promising performance, TDAlign has a primary limitation that its performance improvement would become less pronounced under high-noise scenarios. Future work will explore noise-filtering mechanisms, such as a learnable gating module, to adaptively mitigate noise, thereby enhancing its advantage across varying noise conditions.
}

\section*{Acknowledgments}
This work was supported by the National Natural Science Foundation of China (No. U2468207, 62276215, 6250074528), Sichuan Science and Technology Program (No. 2024ZHCG0166), Postdoctoral Fellowship Program of CPSF (Grant No. GZB20240625) and Fundamental Research Funds for the Central Universities (Grant No. 2682025CX010).
\bibliographystyle{IEEEtran}
\bibliography{Ref}

\begin{thebibliography}{10}
\providecommand{\url}[1]{#1}
\csname url@samestyle\endcsname
\providecommand{\newblock}{\relax}
\providecommand{\bibinfo}[2]{#2}
\providecommand{\BIBentrySTDinterwordspacing}{\spaceskip=0pt\relax}
\providecommand{\BIBentryALTinterwordstretchfactor}{4}
\providecommand{\BIBentryALTinterwordspacing}{\spaceskip=\fontdimen2\font plus
\BIBentryALTinterwordstretchfactor\fontdimen3\font minus \fontdimen4\font\relax}
\providecommand{\BIBforeignlanguage}[2]{{%
\expandafter\ifx\csname l@#1\endcsname\relax
\typeout{** WARNING: IEEEtran.bst: No hyphenation pattern has been}%
\typeout{** loaded for the language `#1'. Using the pattern for}%
\typeout{** the default language instead.}%
\else
\language=\csname l@#1\endcsname
\fi
#2}}
\providecommand{\BIBdecl}{\relax}
\BIBdecl

\bibitem{feng2024multi}
S.~Feng, C.~Miao, K.~Xu, J.~Wu, P.~Wu, Y.~Zhang \emph{et~al.}, ``Multi-scale attention flow for probabilistic time series forecasting,'' \emph{IEEE Transactions on Knowledge and Data Engineering}, vol.~36, no.~5, pp. 2056--2068, 2024.

\bibitem{sezer2020financial}
O.~B. Sezer, M.~U. Gudelek, and A.~M. Ozbayoglu, ``Financial time series forecasting with deep learning : A systematic literature review: 2005–2019,'' \emph{Applied Soft Computing}, vol.~90, p. 106181, 2020.

\bibitem{10643332}
L.~Yang, Z.~Luo, S.~Zhang, F.~Teng, and T.~Li, ``Continual learning for smart city: A survey,'' \emph{IEEE Transactions on Knowledge and Data Engineering}, vol.~36, no.~12, pp. 7805--7824, 2024.

\bibitem{ZHANG2024102413}
S.~Zhang, Z.~Luo, L.~Yang, F.~Teng, and T.~Li, ``A survey of route recommendations: Methods, applications, and opportunities,'' \emph{Information Fusion}, vol. 108, p. 102413, 2024.

\bibitem{9935292}
J.~Deng, X.~Chen, R.~Jiang, X.~Song, and I.~W. Tsang, ``A multi-view multi-task learning framework for multi-variate time series forecasting,'' \emph{IEEE Transactions on Knowledge and Data Engineering}, vol.~35, no.~8, pp. 7665--7680, 2023.

\bibitem{deb2017review}
C.~Deb, F.~Zhang, J.~Yang, S.~E. Lee, and K.~W. Shah, ``A review on time series forecasting techniques for building energy consumption,'' \emph{Renewable and Sustainable Energy Reviews}, vol.~74, pp. 902--924, 2017.

\bibitem{salman2015weather}
A.~G. Salman, B.~Kanigoro, and Y.~Heryadi, ``Weather forecasting using deep learning techniques,'' in \emph{Proceedings of the International Conference on Advanced Computer Science and Information Systems}, 2015, pp. 281--285.

\bibitem{gong2024spatio}
Y.~Gong, T.~He, M.~Chen, B.~Wang, L.~Nie, and Y.~Yin, ``Spatio-temporal enhanced contrastive and contextual learning for weather forecasting,'' \emph{IEEE Transactions on Knowledge and Data Engineering}, 2024.

\bibitem{al2018short}
M.~S. Al-Musaylh, R.~C. Deo, J.~F. Adamowski, and Y.~Li, ``Short-term electricity demand forecasting with mars, svr and arima models using aggregated demand data in queensland, australia,'' \emph{Advanced Engineering Informatics}, vol.~35, pp. 1--16, 2018.

\bibitem{lippi2013short}
M.~Lippi, M.~Bertini, and P.~Frasconi, ``Short-term traffic flow forecasting: An experimental comparison of time-series analysis and supervised learning,'' \emph{IEEE Transactions on Intelligent Transportation Systems}, vol.~14, no.~2, pp. 871--882, 2013.

\bibitem{li2017diffusion}
Y.~Li, R.~Yu, C.~Shahabi, and Y.~Liu, ``Diffusion convolutional recurrent neural network: Data-driven traffic forecasting,'' in \emph{Proceedings of the International Conference on Learning Representations}, 2018.

\bibitem{qin2017dual}
Y.~Qin, D.~Song, H.~Chen, W.~Cheng, G.~Jiang, and G.~Cottrell, ``A dual-stage attention-based recurrent neural network for time series prediction,'' in \emph{Proceedings of the International Joint Conference on Artificial Intelligence}, 2017, pp. 2627--2633.

\bibitem{li2023revisiting}
Z.~Li, S.~Qi, Y.~Li, and Z.~Xu, ``Revisiting long-term time series forecasting: An investigation on linear mapping,'' \emph{arXiv preprint arXiv:2305.10721}, 2023.

\bibitem{zeng2023transformers}
A.~Zeng, M.~Chen, L.~Zhang, and Q.~Xu, ``Are transformers effective for time series forecasting?'' in \emph{Proceedings of the AAAI Conference on Artificial Intelligence}, vol.~37, no.~9, Jun. 2023, pp. 11\,121--11\,128.

\bibitem{das2023long}
A.~Das, W.~Kong, A.~Leach, R.~Sen, and R.~Yu, ``Long-term forecasting with tide: Time-series dense encoder,'' \emph{arXiv preprint arXiv:2304.08424}, 2023.

\bibitem{liu2022scinet}
M.~Liu, A.~Zeng, M.~Chen, Z.~Xu, Q.~Lai, L.~Ma \emph{et~al.}, ``Scinet: Time series modeling and forecasting with sample convolution and interaction,'' in \emph{Proceedings of the Advances in Neural Information Processing Systems}, vol.~35, 2022, pp. 5816--5828.

\bibitem{wang2022micn}
H.~Wang, J.~Peng, F.~Huang, J.~Wang, J.~Chen, and Y.~Xiao, ``Micn: Multi-scale local and global context modeling for long-term series forecasting,'' in \emph{Proceedings of the International Conference on Learning Representations}, 2023.

\bibitem{wu2022timesnet}
H.~Wu, T.~Hu, Y.~Liu, H.~Zhou, J.~Wang, and M.~Long, ``Timesnet: Temporal 2d-variation modeling for general time series analysis,'' in \emph{Proceedings of the International Conference on Learning Representations}, 2023.

\bibitem{li2019enhancing}
S.~Li, X.~Jin, Y.~Xuan, X.~Zhou, W.~Chen, Y.-X. Wang \emph{et~al.}, ``Enhancing the locality and breaking the memory bottleneck of transformer on time series forecasting,'' in \emph{Proceedings of the Advances in Neural Information Processing Systems}, vol.~32.\hskip 1em plus 0.5em minus 0.4em\relax Curran Associates, Inc., 2019.

\bibitem{nie2022time}
Y.~Nie, N.~H.~Nguyen, P.~Sinthong, and J.~Kalagnanam, ``A time series is worth 64 words: Long-term forecasting with transformers,'' in \emph{Proceedings of the International Conference on Learning Representations}, 2023.

\bibitem{zhang2023crossformer}
Y.~Zhang and J.~Yan, ``Crossformer: Transformer utilizing cross-dimension dependency for multivariate time series forecasting,'' in \emph{Proceedings of the International Conference on Learning Representations}, 2023.

\bibitem{lin2023segrnn}
S.~Lin, W.~Lin, W.~Wu, F.~Zhao, R.~Mo, and H.~Zhang, ``Segrnn: Segment recurrent neural network for long-term time series forecasting,'' \emph{arXiv preprint arXiv:2308.11200}, 2023.

\bibitem{liu2021pyraformer}
S.~Liu, H.~Yu, C.~Liao, J.~Li, W.~Lin, A.~X. Liu \emph{et~al.}, ``Pyraformer: Low-complexity pyramidal attention for long-range time series modeling and forecasting,'' in \emph{Proceedings of the International Conference on Learning Representations}, 2021.

\bibitem{zhang2024intriguing}
J.~Zhang, J.~Wang, W.~Qiang, F.~Xu, C.~Zheng, F.~Sun \emph{et~al.}, ``Intriguing properties of positional encoding in time series forecasting,'' \emph{CoRR}, vol. abs/2404.10337, 2024.

\bibitem{hochreiter1997long}
S.~Hochreiter and J.~Schmidhuber, ``Long short-term memory,'' \emph{Neural Computation}, vol.~9, no.~8, pp. 1735--1780, 1997.

\bibitem{cho2014learning}
K.~Cho, B.~van Merri{\"e}nboer, C.~Gulcehre, D.~Bahdanau, F.~Bougares, H.~Schwenk \emph{et~al.}, ``Learning phrase representations using {RNN} encoder{--}decoder for statistical machine translation,'' in \emph{Proceedings of the Conference on Empirical Methods in Natural Language Processing}.\hskip 1em plus 0.5em minus 0.4em\relax Association for Computational Linguistics, 2014, pp. 1724--1734.

\bibitem{zhou2021informer}
H.~Zhou, S.~Zhang, J.~Peng, S.~Zhang, J.~Li, H.~Xiong \emph{et~al.}, ``Informer: Beyond efficient transformer for long sequence time-series forecasting,'' in \emph{Proceedings of the AAAI Conference on Artificial Intelligence}, vol.~35, no.~12, May 2021, pp. 11\,106--11\,115.

\bibitem{shen2023non}
L.~Shen and J.~Kwok, ``Non-autoregressive conditional diffusion models for time series prediction,'' in \emph{Proceedings of the International Conference on Machine Learning}.\hskip 1em plus 0.5em minus 0.4em\relax PMLR, 2023, pp. 31\,016--31\,029.

\bibitem{liu2023itransformer}
Y.~Liu, T.~Hu, H.~Zhang, H.~Wu, S.~Wang, L.~Ma \emph{et~al.}, ``itransformer: Inverted transformers are effective for time series forecasting,'' in \emph{Proceedings of the International Conference on Learning Representations}, 2024.

\bibitem{salinas2020deepar}
D.~Salinas, V.~Flunkert, J.~Gasthaus, and T.~Januschowski, ``Deepar: Probabilistic forecasting with autoregressive recurrent networks,'' \emph{International Journal of Forecasting}, vol.~36, no.~3, pp. 1181--1191, 2020.

\bibitem{lai2018modeling}
G.~Lai, W.-C. Chang, Y.~Yang, and H.~Liu, ``Modeling long-and short-term temporal patterns with deep neural networks,'' in \emph{Proceedings of the International ACM SIGIR Conference on Research \& Development in Information Retrieval}, 2018, pp. 95--104.

\bibitem{tan2023neural}
Y.~Tan, L.~Xie, and X.~Cheng, ``Neural differential recurrent neural network with adaptive time steps,'' \emph{arXiv preprint arXiv:2306.01674}, 2023.

\bibitem{vaswani2017attention}
A.~Vaswani, N.~Shazeer, N.~Parmar, J.~Uszkoreit, L.~Jones, A.~N. Gomez \emph{et~al.}, ``Attention is all you need,'' in \emph{Proceedings of the Advances in Neural Information Processing Systems}, vol.~30, 2017.

\bibitem{wu2021autoformer}
H.~Wu, J.~Xu, J.~Wang, and M.~Long, ``Autoformer: Decomposition transformers with auto-correlation for long-term series forecasting,'' in \emph{Proceedings of the Advances in Neural Information Processing Systems}, vol.~34, 2021, pp. 22\,419--22\,430.

\bibitem{zhou2022fedformer}
T.~Zhou, Z.~Ma, Q.~Wen, X.~Wang, L.~Sun, and R.~Jin, ``Fedformer: Frequency enhanced decomposed transformer for long-term series forecasting,'' in \emph{Proceedings of the International Conference on Machine Learning}.\hskip 1em plus 0.5em minus 0.4em\relax PMLR, 2022, pp. 27\,268--27\,286.

\bibitem{liu2022non}
Y.~Liu, H.~Wu, J.~Wang, and M.~Long, ``Non-stationary transformers: Exploring the stationarity in time series forecasting,'' in \emph{Proceedings of the Advances in Neural Information Processing Systems}, vol.~35, 2022, pp. 9881--9893.

\bibitem{devlin2018bert}
J.~Devlin, ``Bert: Pre-training of deep bidirectional transformers for language understanding,'' \emph{arXiv preprint arXiv:1810.04805}, 2018.

\bibitem{he2022masked}
K.~He, X.~Chen, S.~Xie, Y.~Li, P.~Doll{\'a}r, and R.~Girshick, ``Masked autoencoders are scalable vision learners,'' in \emph{Proceedings of the IEEE/CVF Conference on Computer Vision and Pattern Recognition}, 2022, pp. 16\,000--16\,009.

\bibitem{ekambaram2023tsmixer}
V.~Ekambaram, A.~Jati, N.~Nguyen, P.~Sinthong, and J.~Kalagnanam, ``Tsmixer: Lightweight mlp-mixer model for multivariate time series forecasting,'' in \emph{Proceedings of the ACM SIGKDD Conference on Knowledge Discovery and Data Mining}, 2023, pp. 459--469.

\bibitem{wu2020adversarial}
S.~Wu, X.~Xiao, Q.~Ding, P.~Zhao, Y.~Wei, and J.~Huang, ``Adversarial sparse transformer for time series forecasting,'' in \emph{Proceedings of the Advances in Neural Information Processing Systems}, vol.~33.\hskip 1em plus 0.5em minus 0.4em\relax Curran Associates, Inc., 2020, pp. 17\,105--17\,115.

\bibitem{shao2022spatial}
Z.~Shao, Z.~Zhang, F.~Wang, W.~Wei, and Y.~Xu, ``Spatial-temporal identity: A simple yet effective baseline for multivariate time series forecasting,'' in \emph{Proceedings of the ACM International Conference on Information \& Knowledge Management}, 2022, pp. 4454--4458.

\bibitem{wang2024rethinking}
C.~Wang, Q.~Qi, J.~Wang, H.~Sun, Z.~Zhuang, J.~Wu \emph{et~al.}, ``Rethinking the power of timestamps for robust time series forecasting: A global-local fusion perspective,'' in \emph{Proceedings of the Advances in Neural Information Processing Systems}, 2024.

\bibitem{montgomery2015introduction}
D.~C. Montgomery, C.~L. Jennings, and M.~Kulahci, \emph{Introduction to time series analysis and forecasting}.\hskip 1em plus 0.5em minus 0.4em\relax John Wiley \& Sons, 2015.

\bibitem{huang2022learning}
F.~Huang, T.~Tao, H.~Zhou, L.~Li, and M.~Huang, ``On the learning of non-autoregressive transformers,'' in \emph{Proceedings of the International Conference on Machine Learning}.\hskip 1em plus 0.5em minus 0.4em\relax PMLR, 2022, pp. 9356--9376.

\bibitem{xiao2023survey}
Y.~Xiao, L.~Wu, J.~Guo, J.~Li, M.~Zhang, T.~Qin \emph{et~al.}, ``A survey on non-autoregressive generation for neural machine translation and beyond,'' \emph{IEEE Transactions on Pattern Analysis and Machine Intelligence}, vol.~45, no.~10, pp. 11\,407--11\,427, 2023.

\bibitem{jadon2024comprehensive}
A.~Jadon, A.~Patil, and S.~Jadon, ``A comprehensive survey of regression-based loss functions for time series forecasting,'' in \emph{Proceedings of the International Conference on Data Management, Analytics \& Innovation}.\hskip 1em plus 0.5em minus 0.4em\relax Springer, 2024, pp. 117--147.

\bibitem{xue2013perceptual}
W.~Xue, X.~Mou, L.~Zhang, and X.~Feng, ``Perceptual fidelity aware mean squared error,'' in \emph{Proceedings of the IEEE International Conference on Computer Vision}, 2013, pp. 705--712.

\bibitem{le2019shape}
V.~Le~Guen and N.~Thome, ``Shape and time distortion loss for training deep time series forecasting models,'' in \emph{Proceedings of the Advances in Neural Information Processing Systems}, vol.~32.\hskip 1em plus 0.5em minus 0.4em\relax Curran Associates, Inc., 2019.

\bibitem{box2015time}
G.~E. Box, G.~M. Jenkins, G.~C. Reinsel, and G.~M. Ljung, \emph{Time series analysis: forecasting and control}.\hskip 1em plus 0.5em minus 0.4em\relax John Wiley \& Sons, 2015.

\bibitem{hyndman2018forecasting}
R.~Hyndman and G.~Athanasopoulos, \emph{\BIBforeignlanguage{English}{Forecasting: Principles and Practice}}, 2nd~ed.\hskip 1em plus 0.5em minus 0.4em\relax Australia: OTexts, 2018.

\bibitem{wang2025fredf}
H.~Wang, L.~Pan, Z.~Chen, D.~Yang, S.~Zhang, Y.~Yang \emph{et~al.}, ``Fredf: Learning to forecast in the frequency domain,'' in \emph{Proceedings of the International Conference on Learning Representations}, 2025.

\bibitem{dickey1987determining}
D.~A. Dickey and S.~G. Pantula, ``Determining the order of differencing in autoregressive processes,'' \emph{Journal of Business \& Economic Statistics}, vol.~5, no.~4, pp. 455--461, 1987.

\bibitem{qiu2024tfb}
X.~Qiu, J.~Hu, L.~Zhou, X.~Wu, J.~Du, B.~Zhang \emph{et~al.}, ``Tfb: Towards comprehensive and fair benchmarking of time series forecasting methods,'' vol.~17, no.~9, 2024, pp. 2363--2377.

\bibitem{elliott1992efficient}
G.~Elliott, T.~J. Rothenberg, and J.~H. Stock, ``Efficient tests for an autoregressive unit root,'' \emph{Econometrica}, vol.~64, no.~4, pp. 813--836, 1996.

\bibitem{das2024decoder}
A.~Das, W.~Kong, R.~Sen, and Y.~Zhou, ``A decoder-only foundation model for time-series forecasting,'' in \emph{Proceedings of the International Conference on Machine Learning}, 2024.

\bibitem{jin2024time}
M.~Jin, S.~Wang, L.~Ma, Z.~Chu, J.~Zhang, X.~Shi \emph{et~al.}, ``Time-llm: Time series forecasting by reprogramming large language models,'' in \emph{Proceedings of the International Conference on Learning Representations}, 2024.

\bibitem{cheng2024robusttsf}
H.~Cheng, Q.~Wen, Y.~Liu, and L.~Sun, ``Robusttsf: Towards theory and design of robust time series forecasting with anomalies,'' in \emph{Proceedings of the International Conference on Learning Representations}, 2024.

\end{thebibliography}

\newpage

\appendix
\section{Appendix}
\xq{
\subsection{Proofs} \label{apx:proofs}

    \textbf{\textit{Theorem 1.}}
    \textit{Under a first-order Markov assumption, where \( p(x_{t+i}|x_{<t+i}, X) \approx p(x_{t+i}|x_{t+i-1}, X) \), the discrepancy \(\Psi\) between the point-wise MSE loss, i.e., \(\mathcal{L}_Y\), and the NLL objective is given by:}
    \begin{equation}
        \label{eq:discrepancy_markov}
        \Psi = \sum_{i=2}^{H} \frac{1}{1-\phi_i^2} \left[ \phi_i^2 (\epsilon_i^2 + \epsilon_{i-1}^2) - 2\phi_i\epsilon_i\epsilon_{i-1} \right],
    \end{equation}
    \textit{where \(\epsilon_i = x_{t+i} - \hat{x}_{t+i}\) is the prediction error at step \(i\), and \(\phi_i\) is the partial autocorrelation coefficient between \(x_{t+i}\) and \(x_{t+i-1}\).}

\begin{IEEEproof}[Proof of Theorem 1]
    Following the maximum likelihood principle, we begin by deriving the Negative Log-Likelihood (NLL) objective $\mathcal{L}_{\text{NLL}}$ under a first-order Markov assumption. Given a target time series \( Y = \{x_{t+1}, x_{t+2}, \dots, x_{t+H}\} \) at time step $t$, the likelihood of the target \(Y\) conditioned on the input \(X\) is:

    \[
        \begin{aligned}
            p(Y | X) & = \prod_{i=1}^{H} p(x_{t+i} | X, x_{t+1}, \dots, x_{t+i-1})                                        \\
                     & \approx \prod_{i=1}^{H} p(x_{t+i} | X, x_{t+i-1}) .
        \end{aligned}
    \]

    Let $\epsilon_i = x_{t+i} - \hat{x}_{t+i}$ denote the prediction error at step \(i\), with \(\epsilon_0 = 0\) since \(x_t\) is given and does not require prediction. We model the conditional distribution \( p(x_{t+i} | X, x_{t+i-1}) \) as Gaussian distribution with mean \(\hat{x}_{t+i} + \phi_i \epsilon_{i-1}\) and variance \(\sigma^2 (1 - \phi_i^2)\), where \(\sigma^2\) denotes the marginal variance and \(\phi_i\) is the partial autocorrelation coefficient. Therefore, the NLL is:

    \[
        \begin{aligned}
            \mathcal{L}_{\text{NLL}} & = -\log p(Y | X) = -\sum_{i=1}^{H} \log p(x_{t+i} | X, x_{t+i-1})                                                                    \\
                                     & = \sum_{i=1}^{H} \left[ \frac{(x_{t+i} - (\hat{x}_{t+i} + \phi_i \epsilon_{i-1}))^2}{2\sigma^2 (1 - \phi_i^2)} \right] + \mathcal{C} \\
                                     & = \frac{1}{2\sigma^2} \sum_{i=1}^{H} \left[ \frac{(\epsilon_i - \phi_i \epsilon_{i-1})^2}{1 - \phi_i^2} \right] + \mathcal{C},
        \end{aligned}
    \]
    where \(\mathcal{C}\) is a constant independent of model parameters. Since gradient-based training is unaffected by constants, we can omit \(\mathcal{C}\) and scale factors for clarity. Thus, the gradient relevant part of the $\mathcal{L}_{\text{NLL}}$ is:

    \[
        \mathcal{L}_{\text{NLL}} = \epsilon_1^2+ \sum_{i=2}^{H} \frac{(\epsilon_i - \phi_i \epsilon_{i-1})^2}{1 - \phi_i^2}.
    \]

   However, for computational efficiency, conventional LTSF methods adopt non-autoregressive decoding, generating all prediction steps in a single forward pass. They assume temporal independence across prediction steps and optimize solely with a point-wise loss function \(\mathcal{L}_Y\). When computed using MSE:

    \[
        \mathcal{L}_Y = \sum_{i=1}^{H} \epsilon_i^2.
    \]

    Hence, the discrepancy \(\Psi\) between \(\mathcal{L}_{\text{NLL}}\) and \(\mathcal{L}_Y\) is:

    \[
        \begin{aligned}
            \Psi & = \mathcal{L}_{\text{NLL}} - \mathcal{L}_Y                                                                                                                     \\
                 & = \sum_{i=2}^{H} \frac{(\epsilon_i - \phi_i \epsilon_{i-1})^2}{1 - \phi_i^2} - \sum_{i=2}^{H} \epsilon_i^2                                                     \\
                 & = \sum_{i=2}^{H} \frac{1}{1 - \phi_i^2} \Big[ \epsilon_i^2 - 2\phi_i \epsilon_i \epsilon_{i-1} + \phi_i^2 \epsilon_{i-1}^2 - \epsilon_i^2 (1 - \phi_i^2) \Big] \\
                 & = \sum_{i=2}^{H} \frac{1}{1 - \phi_i^2} \Big[ \phi_i^2 (\epsilon_i^2 + \epsilon_{i-1}^2) - 2\phi_i \epsilon_i \epsilon_{i-1} \Big].
        \end{aligned}
    \]
\end{IEEEproof}

    \textbf{\textit{Proposition 1.}}
    \textit{Let the prediction error \(e_i = d_{t+i} - \hat{d}_{i+1}\) as a zero-mean Gaussian variable with variance \(\sigma_e^2\), the expected value of the adaptive weight \(\rho\) is given by:}
    \begin{equation}
        \mathbb{E}[\rho | \sigma_e^2] = \frac{1}{H}\sum_{i=1}^{H} \Phi\left(-\frac{|d_{t+i}|}{\sigma_e}\right),
    \end{equation}
    \textit{where \(\Phi(\cdot)\) is the cumulative distribution function of the standard normal distribution, \(|d_{t+i}|\) is the true change magnitude between \(x_{t+i}\) and \(x_{t+i-1}\), and \(\sigma_e\) is the standard deviation of the model's prediction error.}

\begin{IEEEproof}[Proof of Proposition 1]
    The sign inconsistency ratio is given by \(\rho = \frac{1}{H} \sum_{i=1}^{H} \mathbbm{1}(\mathcal{E}_i)\), where \(\mathcal{E}_i \equiv \{ \operatorname{sgn}(\hat{d}_{t+i}) \neq \operatorname{sgn}(d_{t+i}) \}\) denotes the event of sign inconsistency. Let the prediction error \(e_i = d_{t+i} - \hat{d}_{t+i}\) as a zero-mean Gaussian random variable, \(e_i \sim \mathcal{N}(0, \sigma_e^2)\), where \(\sigma_e^2\) reflects the prediction error variance.
    The probability of the sign inconsistency event \(\mathcal{E}_i\) is
    \begin{equation}
        P(\mathcal{E}_i) = \Phi\left(-\frac{|d_{t+i}|}{\sigma_e}\right),
    \end{equation}
    where \(\Phi(\cdot)\) is the standard normal cumulative distribution function. 
    This unified expression holds for any nonzero \(d_{t+i}\), as shown below:
    \begin{itemize}
        \item If \(d_{t+i} > 0\), then \(\mathcal{E}_i\) occurs when \(\hat{d}_{t+i} \leq 0\), i.e., \(e_i \geq d_{t+i}\). Hence, \(P(\mathcal{E}_i) = P(e_i \geq d_{t+i}) = \Phi(-|d_{t+i}|/\sigma_e)\).
        \item If \(d_{t+i} < 0\), then \(\mathcal{E}_i\) occurs when \(\hat{d}_{t+i} \geq 0\), i.e., \(e_i \leq d_{t+i}\). Hence, \(P(\mathcal{E}_i) = P(e_i \leq d_{t+i}) = \Phi(-|d_{t+i}|/\sigma_e)\).
    \end{itemize}
    Averaging over the prediction horizon yields the expected value of \(\rho\):
    \begin{equation}
        \mathbb{E}[\rho | \sigma_e^2] = \frac{1}{H} \sum_{i=1}^{H} \Phi\left(-\frac{|d_{t+i}|}{\sigma_e}\right).
    \end{equation}
\end{IEEEproof}
}

\xq{
\subsection{Supplementary Results} \label{apx:proofs}
Tables~\ref{tab:appendix_detailed_results} and \ref{tab:appendix_detailed_results1} present detailed comparisons of the TDP forecasting errors ($\text{MSE}_D$, $\text{MAE}_D$) and the sign inconsistency ratio ($\rho$) between baseline methods and our proposed TDAlign. As demonstrated in these tables, TDAlign shows a superior ability to capture the temporal dependencies between adjacent time steps, predicting both the direction and value of change more accurately across the vast majority of cases.
\begin{table*}[ht]
    \centering
    \caption{\xq{Comparison of TDP forecasting errors between baseline methods and TDAlign. Results are averaged over 5 runs, with better performance highlighted in bold.}}
    \label{tab:appendix_detailed_results}
    \setlength{\tabcolsep}{2pt}
    \resizebox{\textwidth}{!}{
        \begin{tabular}{cc|cccc|cccc|cccc|cccc|cccc|cccc}
            \toprule
            \multicolumn{2}{c|}{Architecture}        & \multicolumn{8}{c|}{CNN-based} & \multicolumn{4}{c|}{MLP-based} & \multicolumn{8}{c|}{Transformer-based} & \multicolumn{4}{c}{RNN-based}                                                                                                                                                                                                                                                                                                                                                                                                                                            \\
            \midrule

            \multicolumn{2}{c|}{Method}              & \multicolumn{2}{c}{MICN}       & \multicolumn{2}{c}{+TDAlign}   & \multicolumn{2}{c}{TimesNet}           & \multicolumn{2}{c|}{+TDAlign} & \multicolumn{2}{c}{DLinear} & \multicolumn{2}{c|}{+TDAlign} & \multicolumn{2}{c}{PatchTST} & \multicolumn{2}{c}{+TDAlign} & \multicolumn{2}{c}{iTransformer} & \multicolumn{2}{c|}{+TDAlign} & \multicolumn{2}{c}{SegRNN} & \multicolumn{2}{c}{+TDAlign}                                                                                                                                                                                 \\
            \multicolumn{2}{c|}{Metric}              & $\text{MSE}_D$                            & \multicolumn{1}{c}{$\text{MAE}_D$ }        & $\text{MSE}_D$                                    & \multicolumn{1}{c}{$\text{MAE}_D$ }       & $\text{MSE}_D$                         & $\text{MAE}_D$                            & $\text{MSE}_D$                          & $\text{MAE}_D$                           & $\text{MSE}_D$                              & $\text{MAE}_D$                            & $\text{MSE}_D$                        & $\text{MAE}_D$                           & $\text{MSE}_D$            & $\text{MAE}_D$    & $\text{MSE}_D$   & \multicolumn{1}{c}{$\text{MAE}_D$ } & $\text{MSE}_D$            & $\text{MAE}_D$    & $\text{MSE}_D$   & $\text{MAE}_D$             & $\text{MSE}_D$            & $\text{MAE}_D$    & $\text{MSE}_D$   & $\text{MAE}_D$                              \\
            \midrule
            \multirow{4}{*}{\myrotcell{ETTh1}}       & 96                             & 0.128                          & 0.233                                  & \textbf{0.117}                & \textbf{0.215}              & 0.141                         & 0.243                        & \textbf{0.126}               & \textbf{0.224}                   & 0.114                         & 0.214                      & \textbf{0.111}               & \textbf{0.209} & 0.131 & 0.232 & \textbf{0.124}          & \textbf{0.223} & 0.130 & 0.234 & \textbf{0.115} & \textbf{0.213} & 0.118 & 0.217 & \textbf{0.109} & \textbf{0.209} \\
                                                     & 192                            & 0.130                          & 0.234                                  & \textbf{0.118}                & \textbf{0.217}              & 0.144                         & 0.246                        & \textbf{0.136}               & \textbf{0.233}                   & 0.116                         & 0.217                      & \textbf{0.115}               & \textbf{0.214} & 0.134 & 0.236 & \textbf{0.128}          & \textbf{0.227} & 0.138 & 0.241 & \textbf{0.120} & \textbf{0.219} & 0.122 & 0.222 & \textbf{0.113} & \textbf{0.214} \\
                                                     & 336                            & 0.133                          & 0.238                                  & \textbf{0.120}                & \textbf{0.220}              & 0.152                         & 0.253                        & \textbf{0.136}               & \textbf{0.233}                   & 0.129                         & 0.237                      & \textbf{0.119}               & \textbf{0.218} & 0.140 & 0.243 & \textbf{0.131}          & \textbf{0.231} & 0.149 & 0.254 & \textbf{0.134} & \textbf{0.232} & 0.130 & 0.230 & \textbf{0.117} & \textbf{0.219} \\
                                                     & 720                            & 0.134                          & 0.240                                  & \textbf{0.125}                & \textbf{0.225}              & 0.155                         & 0.258                        & \textbf{0.140}               & \textbf{0.238}                   & 0.125                         & 0.230                      & \textbf{0.120}               & \textbf{0.220} & 0.157 & 0.263 & \textbf{0.132}          & \textbf{0.233} & 0.155 & 0.260 & \textbf{0.137} & \textbf{0.236} & 0.133 & 0.234 & \textbf{0.122} & \textbf{0.224} \\
            \midrule
            \multirow{4}{*}{\myrotcell{ETTh2}}       & 96                             & 0.083                          & 0.173                                  & \textbf{0.079}                & \textbf{0.168}              & 0.090                         & 0.179                        & \textbf{0.083}               & \textbf{0.162}                   & 0.078                         & 0.170                      & \textbf{0.076}               & \textbf{0.163} & 0.083 & 0.169 & \textbf{0.075}          & \textbf{0.160} & 0.092 & 0.185 & \textbf{0.078} & \textbf{0.164} & 0.080 & 0.161 & \textbf{0.075} & \textbf{0.156} \\
                                                     & 192                            & 0.082                          & 0.173                                  & \textbf{0.079}                & \textbf{0.169}              & 0.093                         & 0.183                        & \textbf{0.084}               & \textbf{0.162}                   & 0.079                         & 0.170                      & \textbf{0.077}               & \textbf{0.164} & 0.087 & 0.177 & \textbf{0.076}          & \textbf{0.161} & 0.092 & 0.184 & \textbf{0.080} & \textbf{0.166} & 0.082 & 0.163 & \textbf{0.076} & \textbf{0.159} \\
                                                     & 336                            & 0.084                          & 0.173                                  & \textbf{0.080}                & \textbf{0.169}              & 0.099                         & 0.195                        & \textbf{0.085}               & \textbf{0.164}                   & 0.086                         & 0.179                      & \textbf{0.078}               & \textbf{0.165} & 0.086 & 0.174 & \textbf{0.078}          & \textbf{0.162} & 0.094 & 0.186 & \textbf{0.082} & \textbf{0.167} & 0.086 & 0.169 & \textbf{0.078} & \textbf{0.161} \\
                                                     & 720                            & 0.086                          & 0.175                                  & \textbf{0.081}                & \textbf{0.169}              & 0.096                         & 0.186                        & \textbf{0.087}               & \textbf{0.165}                   & 0.081                         & 0.170                      & \textbf{0.079}               & \textbf{0.165} & 0.087 & 0.174 & \textbf{0.080}          & \textbf{0.163} & 0.096 & 0.189 & \textbf{0.086} & \textbf{0.169} & 0.089 & 0.175 & \textbf{0.079} & \textbf{0.163} \\
            \midrule
            \multirow{4}{*}{\myrotcell{ETTm1}}       & 96                             & 0.053                          & 0.142                                  & \textbf{0.050}                & \textbf{0.137}              & 0.055                         & 0.145                        & \textbf{0.052}               & \textbf{0.137}                   & 0.049                         & 0.133                      & \textbf{0.049}               & \textbf{0.132} & 0.051 & 0.140 & \textbf{0.050}          & \textbf{0.135} & 0.055 & 0.146 & \textbf{0.051} & \textbf{0.135} & 0.051 & 0.135 & \textbf{0.050} & \textbf{0.135} \\
                                                     & 192                            & 0.053                          & 0.142                                  & \textbf{0.051}                & \textbf{0.137}              & 0.055                         & 0.146                        & \textbf{0.052}               & \textbf{0.137}                   & 0.050                         & 0.134                      & \textbf{0.049}               & \textbf{0.133} & 0.052 & 0.141 & \textbf{0.051}          & \textbf{0.136} & 0.055 & 0.146 & \textbf{0.051} & \textbf{0.136} & 0.051 & 0.136 & \textbf{0.051} & \textbf{0.136} \\
                                                     & 336                            & 0.053                          & 0.143                                  & \textbf{0.051}                & \textbf{0.137}              & 0.056                         & 0.147                        & \textbf{0.052}               & \textbf{0.137}                   & 0.050                         & 0.135                      & \textbf{0.050}               & \textbf{0.134} & 0.053 & 0.143 & \textbf{0.051}          & \textbf{0.136} & 0.056 & 0.147 & \textbf{0.052} & \textbf{0.137} & 0.052 & 0.138 & \textbf{0.051} & \textbf{0.137} \\
                                                     & 720                            & 0.053                          & 0.143                                  & \textbf{0.051}                & \textbf{0.138}              & 0.055                         & 0.146                        & \textbf{0.053}               & \textbf{0.139}                   & 0.051                         & 0.136                      & \textbf{0.051}               & \textbf{0.135} & 0.054 & 0.146 & \textbf{0.052}          & \textbf{0.138} & 0.056 & 0.148 & \textbf{0.053} & \textbf{0.139} & 0.053 & 0.140 & \textbf{0.052} & \textbf{0.139} \\
            \midrule
            \multirow{4}{*}{\myrotcell{ETTm2}}       & 96                             & 0.036                          & 0.105                                  & \textbf{0.035}                & \textbf{0.098}              & 0.037                         & 0.103                        & \textbf{0.035}               & \textbf{0.093}                   & 0.035                         & 0.101                      & \textbf{0.034}               & \textbf{0.093} & 0.037 & 0.106 & \textbf{0.034}          & \textbf{0.093} & 0.039 & 0.109 & \textbf{0.035} & \textbf{0.093} & 0.035 & 0.095 & \textbf{0.034} & \textbf{0.093} \\
                                                     & 192                            & 0.036                          & 0.105                                  & \textbf{0.035}                & \textbf{0.098}              & 0.038                         & 0.106                        & \textbf{0.035}               & \textbf{0.092}                   & 0.035                         & 0.100                      & \textbf{0.034}               & \textbf{0.093} & 0.039 & 0.113 & \textbf{0.034}          & \textbf{0.093} & 0.039 & 0.110 & \textbf{0.035} & \textbf{0.092} & 0.035 & 0.095 & \textbf{0.034} & \textbf{0.093} \\
                                                     & 336                            & 0.035                          & 0.103                                  & \textbf{0.034}                & \textbf{0.098}              & 0.037                         & 0.102                        & \textbf{0.035}               & \textbf{0.092}                   & 0.035                         & 0.100                      & \textbf{0.034}               & \textbf{0.093} & 0.041 & 0.119 & \textbf{0.035}          & \textbf{0.095} & 0.038 & 0.108 & \textbf{0.035} & \textbf{0.092} & 0.035 & 0.096 & \textbf{0.034} & \textbf{0.093} \\
                                                     & 720                            & 0.036                          & 0.106                                  & \textbf{0.034}                & \textbf{0.096}              & 0.037                         & 0.102                        & \textbf{0.035}               & \textbf{0.092}                   & 0.039                         & 0.111                      & \textbf{0.034}               & \textbf{0.093} & 0.041 & 0.119 & \textbf{0.035}          & \textbf{0.094} & 0.038 & 0.108 & \textbf{0.035} & \textbf{0.092} & 0.035 & 0.096 & \textbf{0.034} & \textbf{0.093} \\
            \midrule
            \multirow{4}{*}{\myrotcell{electricity}} & 96                             & 0.078                          & 0.181                                  & \textbf{0.076}                & \textbf{0.174}              & 0.083                         & 0.188                        & \textbf{0.081}               & \textbf{0.178}                   & 0.059                         & 0.150                      & \textbf{0.059}               & \textbf{0.149} & 0.058 & 0.148 & \textbf{0.056}          & \textbf{0.142} & 0.067 & 0.161 & \textbf{0.065} & \textbf{0.155} & 0.061 & 0.152 & \textbf{0.056} & \textbf{0.142} \\
                                                     & 192                            & 0.081                          & 0.185                                  & \textbf{0.077}                & \textbf{0.175}              & 0.085                         & 0.190                        & \textbf{0.081}               & \textbf{0.179}                   & 0.061                         & 0.152                      & \textbf{0.061}               & \textbf{0.151} & 0.060 & 0.151 & \textbf{0.058}          & \textbf{0.144} & 0.067 & 0.161 & \textbf{0.065} & \textbf{0.155} & 0.064 & 0.155 & \textbf{0.059} & \textbf{0.146} \\
                                                     & 336                            & 0.084                          & 0.190                                  & \textbf{0.078}                & \textbf{0.176}              & 0.088                         & 0.193                        & \textbf{0.086}               & \textbf{0.184}                   & 0.063                         & 0.155                      & \textbf{0.063}               & \textbf{0.154} & 0.062 & 0.154 & \textbf{0.060}          & \textbf{0.147} & 0.069 & 0.164 & \textbf{0.067} & \textbf{0.158} & 0.066 & 0.158 & \textbf{0.063} & \textbf{0.151} \\
                                                     & 720                            & 0.086                          & 0.190                                  & \textbf{0.079}                & \textbf{0.177}              & 0.092                         & 0.198                        & \textbf{0.089}               & \textbf{0.187}                   & \textbf{0.068}                & 0.162                      & 0.069                        & \textbf{0.161} & 0.068 & 0.162 & \textbf{0.065}          & \textbf{0.154} & 0.075 & 0.173 & \textbf{0.072} & \textbf{0.165} & 0.071 & 0.165 & \textbf{0.067} & \textbf{0.157} \\
            \midrule
            \multirow{4}{*}{\myrotcell{illness}}     & 24                             & 0.257                          & 0.292                                  & \textbf{0.208}                & \textbf{0.253}              & 0.380                         & 0.391                        & \textbf{0.271}               & \textbf{0.314}                   & 0.207                         & 0.265                      & \textbf{0.188}               & \textbf{0.239} & 0.221 & 0.278 & \textbf{0.195}          & \textbf{0.245} & 0.259 & 0.316 & \textbf{0.218} & \textbf{0.264} & 0.215 & 0.266 & \textbf{0.204} & \textbf{0.255} \\
                                                     & 36                             & 0.248                          & 0.284                                  & \textbf{0.205}                & \textbf{0.245}              & 0.372                         & 0.386                        & \textbf{0.257}               & \textbf{0.300}                   & 0.207                         & 0.262                      & \textbf{0.189}               & \textbf{0.236} & 0.223 & 0.276 & \textbf{0.193}          & \textbf{0.240} & 0.259 & 0.314 & \textbf{0.216} & \textbf{0.257} & 0.215 & 0.262 & \textbf{0.203} & \textbf{0.250} \\
                                                     & 48                             & 0.250                          & 0.283                                  & \textbf{0.210}                & \textbf{0.245}              & 0.323                         & 0.363                        & \textbf{0.243}               & \textbf{0.287}                   & 0.203                         & 0.251                      & \textbf{0.190}               & \textbf{0.235} & 0.222 & 0.275 & \textbf{0.191}          & \textbf{0.236} & 0.263 & 0.317 & \textbf{0.214} & \textbf{0.252} & 0.220 & 0.264 & \textbf{0.209} & \textbf{0.250} \\
                                                     & 60                             & 0.241                          & 0.278                                  & \textbf{0.210}                & \textbf{0.243}              & 0.322                         & 0.364                        & \textbf{0.242}               & \textbf{0.286}                   & 0.210                         & 0.258                      & \textbf{0.193}               & \textbf{0.236} & 0.220 & 0.265 & \textbf{0.191}          & \textbf{0.235} & 0.260 & 0.311 & \textbf{0.220} & \textbf{0.254} & 0.226 & 0.267 & \textbf{0.214} & \textbf{0.255} \\
            \midrule
            \multirow{4}{*}{\myrotcell{weather}}     & 96                             & 0.039                          & 0.051                                  & \textbf{0.039}                & \textbf{0.042}              & 0.040                         & 0.050                        & \textbf{0.039}               & \textbf{0.042}                   & 0.039                         & 0.042                      & \textbf{0.038}               & \textbf{0.041} & 0.039 & 0.044 & \textbf{0.039}          & \textbf{0.040} & 0.039 & 0.046 & \textbf{0.039} & \textbf{0.041} & 0.039 & 0.041 & \textbf{0.038} & \textbf{0.040} \\
                                                     & 192                            & 0.040                          & 0.052                                  & \textbf{0.039}                & \textbf{0.043}              & 0.040                         & 0.048                        & \textbf{0.039}               & \textbf{0.042}                   & 0.039                         & 0.041                      & \textbf{0.039}               & \textbf{0.041} & 0.039 & 0.044 & \textbf{0.039}          & \textbf{0.040} & 0.039 & 0.048 & \textbf{0.039} & \textbf{0.041} & 0.039 & 0.041 & \textbf{0.039} & \textbf{0.040} \\
                                                     & 336                            & 0.040                          & 0.051                                  & \textbf{0.039}                & \textbf{0.043}              & 0.040                         & 0.048                        & \textbf{0.039}               & \textbf{0.042}                   & 0.039                         & 0.042                      & \textbf{0.039}               & \textbf{0.041} & 0.039 & 0.045 & \textbf{0.039}          & \textbf{0.040} & 0.040 & 0.048 & \textbf{0.039} & \textbf{0.041} & 0.039 & 0.041 & \textbf{0.039} & \textbf{0.040} \\
                                                     & 720                            & 0.040                          & 0.048                                  & \textbf{0.040}                & \textbf{0.043}              & 0.040                         & 0.047                        & \textbf{0.040}               & \textbf{0.041}                   & 0.040                         & 0.041                      & \textbf{0.040}               & \textbf{0.041} & 0.040 & 0.045 & \textbf{0.040}          & \textbf{0.040} & 0.041 & 0.049 & \textbf{0.040} & \textbf{0.041} & 0.040 & 0.042 & \textbf{0.040} & \textbf{0.041} \\
            \bottomrule
        \end{tabular}

    }
\end{table*}

\begin{table*}[ht]
    \centering
    \caption{\xq{Comparison of the sign inconsistency ratio ($\rho$) between baseline methods and TDAlign. Results are averaged over 5 runs, with better performance highlighted in bold.}}
    \label{tab:appendix_detailed_results1}
    \resizebox{\textwidth}{!}{
        \begin{tabular}{cc|cccc|cc|cccc|cc}
            \toprule
            \multicolumn{2}{c|}{Architecture}        & \multicolumn{4}{c|}{CNN-based} & \multicolumn{2}{c|}{MLP-based} & \multicolumn{4}{c|}{Transformer-based} & \multicolumn{2}{c}{RNN-based}                                                                                                                                  \\
            \midrule
            \multicolumn{2}{c|}{Method}              & MICN                           & +TDAlign                       & TimesNet                               & +TDAlign                      & DLinear        & +TDAlign & PatchTST       & +TDAlign & iTransformer   & +TDAlign & SegRNN         & +TDAlign                  \\
            \midrule
            \multirow{4}{*}{\myrotcell{ETTh1}}       & 96                             & 0.334                          & \textbf{0.303}                         & 0.372                         & \textbf{0.314} & 0.306    & \textbf{0.296} & 0.344    & \textbf{0.328} & 0.340    & \textbf{0.303} & 0.311    & \textbf{0.296} \\
                                                     & 192                            & 0.332                          & \textbf{0.301}                         & 0.378                         & \textbf{0.328} & 0.309    & \textbf{0.303} & 0.350    & \textbf{0.334} & 0.353    & \textbf{0.311} & 0.319    & \textbf{0.305} \\
                                                     & 336                            & 0.337                          & \textbf{0.306}                         & 0.387                         & \textbf{0.327} & 0.334    & \textbf{0.312} & 0.363    & \textbf{0.342} & 0.371    & \textbf{0.337} & 0.339    & \textbf{0.316} \\
                                                     & 720                            & 0.346                          & \textbf{0.319}                         & 0.399                         & \textbf{0.338} & 0.334    & \textbf{0.321} & 0.387    & \textbf{0.352} & 0.384    & \textbf{0.354} & 0.352    & \textbf{0.328} \\
            \midrule
            \multirow{4}{*}{\myrotcell{ETTh2}}       & 96                             & 0.274                          & \textbf{0.247}                         & 0.318                         & \textbf{0.266} & 0.251    & \textbf{0.237} & 0.283    & \textbf{0.235} & 0.316    & \textbf{0.249} & 0.258    & \textbf{0.234} \\
                                                     & 192                            & 0.273                          & \textbf{0.244}                         & 0.322                         & \textbf{0.268} & 0.249    & \textbf{0.238} & 0.295    & \textbf{0.236} & 0.319    & \textbf{0.255} & 0.265    & \textbf{0.236} \\
                                                     & 336                            & 0.274                          & \textbf{0.240}                         & 0.347                         & \textbf{0.286} & 0.261    & \textbf{0.239} & 0.287    & \textbf{0.237} & 0.323    & \textbf{0.265} & 0.289    & \textbf{0.242} \\
                                                     & 720                            & 0.275                          & \textbf{0.237}                         & 0.342                         & \textbf{0.270} & 0.247    & \textbf{0.240} & 0.287    & \textbf{0.239} & 0.328    & \textbf{0.277} & 0.307    & \textbf{0.243} \\
            \midrule
            \multirow{4}{*}{\myrotcell{ETTm1}}       & 96                             & 0.376                          & \textbf{0.366}                         & 0.394                         & \textbf{0.371} & 0.357    & \textbf{0.349} & 0.375    & \textbf{0.361} & 0.402    & \textbf{0.372} & 0.369    & \textbf{0.366} \\
                                                     & 192                            & 0.377                          & \textbf{0.366}                         & 0.393                         & \textbf{0.370} & 0.359    & \textbf{0.351} & 0.378    & \textbf{0.362} & 0.402    & \textbf{0.376} & 0.370    & \textbf{0.367} \\
                                                     & 336                            & 0.378                          & \textbf{0.366}                         & 0.400                         & \textbf{0.377} & 0.361    & \textbf{0.354} & 0.384    & \textbf{0.366} & 0.405    & \textbf{0.381} & 0.375    & \textbf{0.370} \\
                                                     & 720                            & 0.376                          & \textbf{0.367}                         & 0.400                         & \textbf{0.380} & 0.366    & \textbf{0.360} & 0.391    & \textbf{0.372} & 0.409    & \textbf{0.390} & 0.387    & \textbf{0.377} \\
            \midrule
            \multirow{4}{*}{\myrotcell{ETTm2}}       & 96                             & 0.260                          & \textbf{0.244}                         & 0.278                         & \textbf{0.244} & 0.251    & \textbf{0.232} & 0.267    & \textbf{0.237} & 0.284    & \textbf{0.251} & 0.246    & \textbf{0.236} \\
                                                     & 192                            & 0.259                          & \textbf{0.243}                         & 0.284                         & \textbf{0.245} & 0.250    & \textbf{0.233} & 0.274    & \textbf{0.238} & 0.287    & \textbf{0.258} & 0.251    & \textbf{0.237} \\
                                                     & 336                            & 0.258                          & \textbf{0.241}                         & 0.278                         & \textbf{0.247} & 0.251    & \textbf{0.234} & 0.279    & \textbf{0.243} & 0.288    & \textbf{0.259} & 0.256    & \textbf{0.239} \\
                                                     & 720                            & 0.260                          & \textbf{0.236}                         & 0.281                         & \textbf{0.248} & 0.263    & \textbf{0.234} & 0.279    & \textbf{0.245} & 0.289    & \textbf{0.260} & 0.261    & \textbf{0.240} \\
            \midrule
            \multirow{4}{*}{\myrotcell{electricity}} & 96                             & 0.251                          & \textbf{0.241}                         & 0.263                         & \textbf{0.245} & 0.203    & \textbf{0.202} & 0.203    & \textbf{0.197} & 0.219    & \textbf{0.211} & 0.207    & \textbf{0.198} \\
                                                     & 192                            & 0.256                          & \textbf{0.241}                         & 0.265                         & \textbf{0.245} & 0.205    & \textbf{0.205} & 0.206    & \textbf{0.200} & 0.220    & \textbf{0.212} & 0.210    & \textbf{0.202} \\
                                                     & 336                            & 0.260                          & \textbf{0.242}                         & 0.267                         & \textbf{0.248} & 0.209    & \textbf{0.208} & 0.210    & \textbf{0.203} & 0.223    & \textbf{0.215} & 0.213    & \textbf{0.205} \\
                                                     & 720                            & 0.259                          & \textbf{0.242}                         & 0.272                         & \textbf{0.251} & 0.216    & \textbf{0.216} & 0.219    & \textbf{0.211} & 0.234    & \textbf{0.222} & 0.222    & \textbf{0.213} \\
            \midrule
            \multirow{4}{*}{\myrotcell{illness}}     & 24                             & 0.373                          & \textbf{0.310}                         & 0.387                         & \textbf{0.354} & 0.309    & \textbf{0.268} & 0.327    & \textbf{0.276} & 0.367    & \textbf{0.296} & 0.318    & \textbf{0.298} \\
                                                     & 36                             & 0.365                          & \textbf{0.291}                         & 0.398                         & \textbf{0.348} & 0.295    & \textbf{0.264} & 0.331    & \textbf{0.278} & 0.373    & \textbf{0.290} & 0.311    & \textbf{0.287} \\
                                                     & 48                             & 0.353                          & \textbf{0.272}                         & 0.397                         & \textbf{0.341} & 0.273    & \textbf{0.258} & 0.319    & \textbf{0.267} & 0.381    & \textbf{0.282} & 0.317    & \textbf{0.283} \\
                                                     & 60                             & 0.352                          & \textbf{0.262}                         & 0.401                         & \textbf{0.339} & 0.277    & \textbf{0.250} & 0.310    & \textbf{0.259} & 0.376    & \textbf{0.281} & 0.319    & \textbf{0.290} \\
            \midrule
            \multirow{4}{*}{\myrotcell{weather}}     & 96                             & 0.357                          & \textbf{0.327}                         & 0.365                         & \textbf{0.334} & 0.327    & \textbf{0.319} & 0.338    & \textbf{0.313} & 0.362    & \textbf{0.326} & 0.314    & \textbf{0.310} \\
                                                     & 192                            & 0.354                          & \textbf{0.339}                         & 0.367                         & \textbf{0.336} & 0.323    & \textbf{0.315} & 0.339    & \textbf{0.312} & 0.364    & \textbf{0.326} & 0.321    & \textbf{0.311} \\
                                                     & 336                            & 0.353                          & \textbf{0.340}                         & 0.363                         & \textbf{0.332} & 0.323    & \textbf{0.313} & 0.345    & \textbf{0.312} & 0.366    & \textbf{0.327} & 0.326    & \textbf{0.312} \\
                                                     & 720                            & 0.345                          & \textbf{0.333}                         & 0.362                         & \textbf{0.332} & 0.319    & \textbf{0.311} & 0.344    & \textbf{0.311} & 0.365    & \textbf{0.326} & 0.333    & \textbf{0.314} \\
            \bottomrule
        \end{tabular}
    }
\end{table*}

}

\end{document}